\RequirePackage{fix-cm}
\documentclass{article} 
\usepackage{iclr2027_conference,times}

\usepackage{amsmath,amsfonts,bm}

\def\Figref#1{Figure~\ref{#1}}

\def\eqref#1{equation~\ref{#1}}

\def\1{\bm{1}}

\DeclareMathAlphabet{\mathsfit}{\encodingdefault}{\sfdefault}{m}{sl}
\SetMathAlphabet{\mathsfit}{bold}{\encodingdefault}{\sfdefault}{bx}{n}

\newcommand{\E}{\mathbb{E}}

\newcommand{\R}{\mathbb{R}}

\usepackage{amssymb}
\usepackage{amsthm}
\usepackage{amsmath}
\usepackage{bm}
\theoremstyle{plain}
\newtheorem{theorem}{Theorem}[section]

\newtheorem{lemma}[theorem]{Lemma}

\theoremstyle{definition}
\newtheorem{definition}[theorem]{Definition}

\newtheoremstyle{myremark}
  {\topsep}
  {\topsep}
  {\normalfont}
  {}
  {\bfseries}
  {.}
  {.5em}
  {}
\theoremstyle{myremark}
\newtheorem{remark}{Remark}[section]

\newcommand{\indep}{\perp \!\!\! \perp}

\usepackage{hyperref}
\usepackage{url}
\usepackage{enumitem}
\usepackage{subcaption}
\usepackage{wrapfig}
\usepackage{bbm}
\usepackage{cleveref}
\usepackage{mathtools}
\usepackage{multirow}

\usepackage{tikz}
\usetikzlibrary{positioning, arrows.meta, calc, fit, shapes.geometric}
\usepackage{pgfplots}
\usepgfplotslibrary{groupplots}
\pgfplotsset{compat=1.18}
\usepackage{xcolor}

\definecolor{setfull}{RGB}{205,120,62}
\definecolor{setmid}{RGB}{114,92,171}
\definecolor{setinner}{RGB}{62,145,117}

\definecolor{xfill}{RGB}{232,240,254}   
\definecolor{ufill}{RGB}{255,243,224}   
\definecolor{xdraw}{RGB}{91,135,201}    
\definecolor{udraw}{RGB}{214,141,72}    
\definecolor{edgecol}{RGB}{120,130,145} 
\definecolor{Ifill}{RGB}{232,240,254}
\definecolor{Idraw}{RGB}{91,135,201}
\definecolor{Pfill}{RGB}{232,245,233}
\definecolor{Pdraw}{RGB}{102,187,106}
\definecolor{Bfill}{RGB}{243,229,245}
\definecolor{Bdraw}{RGB}{171,71,188}
\definecolor{Vfill}{RGB}{255,243,224}
\definecolor{Vdraw}{RGB}{214,141,72}
\definecolor{edgecol}{RGB}{120,130,145}

\title{
Neural Video Coding under Limited Delay: \\
Theoretical Optimality and System Design}

\author{
Likun Cai$^1$, \; Jun Chen$^2$ and Ashish Khisti$^1$ \\
$^1$University of Toronto \qquad $^2$ McMaster University \\
\texttt{likun.cai@mail.utoronto.ca \qquad chenjun@mcmaster.ca} \\
\texttt{akhisti@ece.utoronto.ca} \\
}

\iclrfinalcopy 

\begin{document}

\maketitle

\begin{abstract}

Traditional neural video compression (NVC) methods predominantly follow strictly causal predictive-coding paradigm, treating zero-frame delay as a fixed constraint rather than delay as an explicit design parameter. 
From an information-theoretic perspective, allowing limited frame delay enables a codec to exploit correlations with future frames within the delay window, thereby reducing uncertainty that is inaccessible to strictly causal prediction and potentially improving the rate-distortion tradeoff. 
In this work, we formalize this intuition through a rigorous rate–distortion analysis builds on the clique factorization of Markov networks, and propose \emph{Graph-Factored Sequential Coding} (GFSC), a principled framework for sequential video coding under limited delay. 
To our knowledge, we provide the first theoretical result showing that, for general \(K\)-order Markov sources with positive distributions, a $K$-frame delay is sufficient to attain joint-coding rate-distortion optimality over a nontrivial distortion region, a guarantee unavailable to zero-delay causal coding in general. 
In practice, we instantiate GFSC through lightweight neural modules, allowing existing causal predictive codecs to be augmented without modifying their core architectures. 
Experiments across multiple NVC models and datasets show that GFSC consistently improves rate-distortion performance over the corresponding causal counterparts. 
Under matched delay budgets, GFSC also achieves competitive performance compared with representative bidirectional codecs. 
\end{abstract}

\section{Introduction}

Video compression can be fundamentally viewed as a sequential lossy source coding problem~\citep{viswanathan2002sequential, yang2011rate, ma2007delayed}. 
Most existing neural video compression (NVC) models follow a strictly \emph{causal predictive coding} paradigm~\citep{agustsson2020scale, lu2019dvc, lu2020end, li2021deep, mentzer2022vct}, where each frame is encoded using only previously reconstructed frames and therefore cannot exploit future correlations. 
For temporally correlated video, future frames provide predictive information about the current frame and reduce its uncertainty, potentially improving the rate-distortion tradeoff.
In addition, allowing limited frame delay does not necessarily incur substantial additional latency. 
At $60$~fps, a single-frame lookahead introduces only $16.67$~ms of buffering latency, which is modest relative to the inference latency of modern NVC models and tolerable in many practical scenarios. 
Frame delay should therefore be treated as an explicit design parameter rather than fixed a priori to zero. 

The information-theoretic basis of limited delay coding has been studied through \emph{delayed sequential coding}~\citep{ma2007delayed}.
In particular, for \(K\)-order Gauss-Markov sources, a \(K\)-frame delay is sufficient to attain joint-coding rate-distortion optimality over a nonempty distortion region, a guarantee that zero-delay causal coding does not generally enjoy. 
However, it remains unclear whether this optimality extends beyond Gaussian sources, and the current theory does not provide a constructive architectural principle for translating a given delay budget into a practical coding system. 
Existing bidirectional codecs~\citep{wu2018video,djelouah2019neural,xu2024ibvc,yang2020learning,yilmaz2021end} exploit future frames in practice, but typically rely on fixed anchor-based structures designed for relatively large delay budgets, thereby inefficient in the limited-delay regime of interest.
Consequently, a unified framework that characterizes limited-delay optimality for general Markov sources while enabling flexible practical NVC remains largely underexplored. 

\begin{figure}[t]
\centering
\raisebox{1mm}{
\begin{subfigure}[c]{0.3275\linewidth}
\centering
\resizebox{\linewidth}{!}{%
\begin{tikzpicture}[
    font=\small,
    >=Latex,
    line cap=round,
    line join=round,
    enc/.style={
        draw=blue!75!black,
        fill=blue!8,
        rounded corners=2pt,
        thick,
        minimum width=1.85cm,
        minimum height=0.85cm,
        align=center
    },
    dec/.style={
        draw=teal!65!black,
        fill=teal!8,
        rounded corners=2pt,
        thick,
        minimum width=1.85cm,
        minimum height=0.85cm,
        align=center
    },
    mainarrow/.style={->,very thick,draw=black!75},
    history/.style={thick,dashed,draw=black!55},
    sidearrow/.style={history,->},
    lab/.style={inner sep=1pt,outer sep=0pt,text=black!85}
]

\foreach \i/\y in {1/0,2/-1.85,3/-3.70} {
    \node[enc] (f\i) at (0,\y)   {$f_{\i}$};
    \node[dec] (g\i) at (4.5,\y) {$g_{\i}$};

    \draw[mainarrow]
        ($(f\i.north)+(0,0.50)$) node[above] {$W$}
        -- (f\i.north);

    \draw[mainarrow]
        ($(g\i.north)+(0,0.50)$) node[above] {$W$}
        -- (g\i.north);

    \draw[mainarrow]
        (f\i.east)
        -- node[above,lab,pos=0.24] {$M_{\i}$}
        (g\i.west);

    \draw[mainarrow]
        (g\i.east) -- ++(0.55,0)
        node[right] {$\hat{X}_{\i}$};
}

\foreach \i/\j in {1/2,2/3} {
    \draw[mainarrow]
        ($(f\i.west)+(-0.55,0.18)$) node[left] {$X_{\i}$}
        -- ($(f\i.west)+(0,0.18)$);

    \draw[mainarrow]
        ($(f\i.west)+(-0.55,-0.18)$) node[left] {$X_{\j}$}
        -- ($(f\i.west)+(0,-0.18)$);
}

\draw[mainarrow]
    ($(f3.west)+(-0.55,0)$) node[left] {$X_3$}
    -- (f3.west);

\coordinate (tap1) at ($(f1.east)!0.58!(g1.west)$);
\coordinate (tap2) at ($(f2.east)!0.42!(g2.west)$);

\coordinate (level2)  at ($(f2.north)+(0,0.65)$);
\coordinate (level3A) at ($(f3.north)+(0,0.65)$);
\coordinate (level3B) at ($(f3.north)+(0,0.38)$);

\coordinate (m1to2) at (tap1 |- level2);
\coordinate (m1to3) at (tap1 |- level3A);
\coordinate (m2to3) at (tap2 |- level3B);

\coordinate (cross12) at (tap1 |- tap2);
\coordinate (cross23) at (tap2 |- m1to3);

\draw[history]
    (tap1) -- ($(cross12)+(0,0.09)$);
\draw[history]
    ($(cross12)+(0,-0.09)$) -- (m1to3);

\draw[history]
    (tap2) -- ($(cross23)+(0,0.09)$);
\draw[history]
    ($(cross23)+(0,-0.09)$) -- (m2to3);

\draw[sidearrow]
    (m1to2) -| ($(f2.north)+(0.32,0)$);
\draw[sidearrow]
    (m1to2) -| ($(g2.north)+(-0.32,0)$);

\draw[sidearrow]
    (m1to3) -| ($(f3.north)+(0.32,0)$);
\draw[sidearrow]
    (m1to3) -| ($(g3.north)+(-0.32,0)$);

\draw[sidearrow]
    (m2to3) -| ($(f3.north)+(0.62,0)$);
\draw[sidearrow]
    (m2to3) -| ($(g3.north)+(-0.62,0)$);

\foreach \p in {tap1,tap2,m1to2,m1to3,m2to3} {
    \fill[black!65] (\p) circle (1.3pt);
}

\end{tikzpicture}%
}
\end{subfigure}}%
\begin{subfigure}[c]{0.3175\linewidth}
\centering
\resizebox{\linewidth}{!}{
    \begin{tikzpicture}
    \begin{axis}[
        axis line style={line width=0.3pt},
        width=\linewidth,
        height=0.72\linewidth,
        xmin=0.030, xmax=0.095,
        scaled x ticks=false,
        xticklabel style={/pgf/number format/fixed,/pgf/number format/precision=2},
        xtick={0.03,0.05,0.07,0.09},
        ymin=33.75, ymax=37.75,
        xlabel={\scalebox{0.7}{Rate (bpp)}},
        ylabel={\scalebox{0.7}{PSNR (dB)}},
        xlabel style={
            font=\tiny,
        },
        ylabel style={font=\tiny},
        grid=both,
        major grid style={gray!30},
        minor grid style={gray!15},
        tick align=outside,
        tick pos=left,
        tick label style={
            font=\tiny, 
            inner sep=-2pt
        },
        major tick length=2pt,
        minor tick length=2pt,
        legend style={
            at={(0.5,1.00)},
            anchor=south,
            legend columns=2,
            draw=none,
            fill=none,
            font=\fontsize{5}{5.5}\selectfont,
            inner sep=0.5pt,
            /tikz/every even column/.append style={column sep=3pt},
            /tikz/row sep=-1pt
        },
        legend image post style={xscale=0.7},
        legend cell align={left},
        line width=0.8pt,
        mark size=1.0pt,
    ]
    
    \addplot[
        color=green!50!black,
        mark=o,
    ]
    coordinates {
        (0.02567,33.3877)
        (0.03864,34.6321)
        (0.06056,35.8432)
        (0.09874,36.9968)
    };
    \addlegendentry{Causal}
    
    \addplot[
        color=orange!50!white,
        mark=o,
    ]
    coordinates {
        (0.02713,34.1748)
        (0.04150,35.4637)
        (0.06621,36.6592)
        (0.11237,37.7571)
    };
    \addlegendentry{$K=1$}
    
    \addplot[
        color=orange!70!white,
        mark=o,
    ]
    coordinates {
        (0.02807,34.3799)
        (0.04228,35.6792)
        (0.06689,36.8741)
        (0.11286,37.9083)
    };
    \addlegendentry{$K=2$}
    
    \addplot[
        color=orange!90!white,
        mark=o,
    ]
    coordinates {
        (0.03108,34.6910)
        (0.04537,35.9891)
        (0.07377,37.2683)
        (0.12487,38.3358)
    };
    \addlegendentry{$K=3$}
    \end{axis}
    \end{tikzpicture}
}

\end{subfigure}%
\begin{subfigure}[c]{0.315\linewidth}
\centering
\begingroup
\definecolor{rlcausal}{RGB}{38,127,71}
\definecolor{rlgfsc}{RGB}{239,135,39}
\definecolor{rlbidir}{RGB}{65,126,188}

\def\PairBarWidth{10pt}
\def\PairHalfWidth{5pt}

\resizebox{\linewidth}{!}{%
\begin{tikzpicture}
\begin{groupplot}[
    group style={group size=1 by 2,vertical sep=0.42cm},
    scale only axis,
    width=0.84\linewidth,
    height=0.225\linewidth,
    ybar,
    /tikz/bar width=\PairBarWidth,
    /tikz/bar shift=0pt,
    xmin=-0.50,xmax=3.50,
    xtick={0,1,2,3},
    xticklabels={Causal,$K=1$,$K=2$,$K=3$},
    xticklabel style={font=\tiny},
    yticklabel style={font=\tiny},
    ylabel style={
        font=\tiny,
        at={(axis description cs:-0.18,0.5)},
        anchor=center
    },
    scaled y ticks=false,
    axis lines*=left,
    axis line style={line width=0.3pt},
    tick align=outside,
    major tick length=1.5pt,
    ymajorgrids,
    grid style={gray!20},
    nodes near coords={}, 
    legend style={
        at={(0.5,1.13)},anchor=south,
        legend columns=3,font=\tiny,
        draw=none,inner sep=1pt,
        /tikz/every even column/.append style={column sep=3pt}
    },
    legend cell align={left},
    legend image code/.code={
        \path[#1] (0,-0.06cm) rectangle (0.17cm,0.06cm);
    }
]

\nextgroupplot[
    ylabel={Latency (s/frame)},
    ymin=0.75, ymax=2.1,
    xticklabels=\empty
]
\addplot[fill=rlcausal,draw=none,/tikz/bar shift=0pt]
    coordinates {(0,0.96)};
\addlegendentry{Causal}
\addplot[
    fill=rlgfsc,draw=none,/tikz/bar shift=-\PairHalfWidth
]
    coordinates {(1,1.09) (2,1.24) (3,1.32)};
\addlegendentry{GFSC}
\addplot[
    fill=rlbidir,draw=none,/tikz/bar shift=\PairHalfWidth
]
    coordinates {(1,1.48) (2,2.03) (3,2.06)};
\addlegendentry{DCVC-B}

\nextgroupplot[
    ylabel={BD-rate (\%)},
    ymin=-35,ymax=10,
    ytick={-30,-20,-10,0,10}
]
\addplot[fill=rlcausal,draw=none,/tikz/bar shift=0pt]
    coordinates {(0,0.00)};
\addplot[
    fill=rlgfsc,draw=none,/tikz/bar shift=-\PairHalfWidth
]
    coordinates {(1,-20.57) (2,-25.29) (3,-28.22)};
\addplot[fill=rlbidir,draw=none,/tikz/bar shift=\PairHalfWidth]
    coordinates {(1,3.70) (2,-18.64) (3,-27.24)};
\draw[black!65,line width=0.4pt]
    (axis cs:-0.50,0) -- (axis cs:3.50,0);
\end{groupplot}
\end{tikzpicture}
}
\endgroup
\end{subfigure}
\caption{
\textit{Left}: GFSC-$1$ coding architecture for three frames.
\textit{Middle}: Rate-PSNR comparison between zero-delay DCVC and GFSC
with delays $K=1,2,3$ on UVG.
\textit{Right}: Latency (top; lower the better) and BD-rate (bottom; lower the better) of GFSC and DCVC-B
at matched delays, with zero-delay DCVC as the baseline.
BD-rate is measured relative to zero-delay DCVC.
}
\label{fig:lisc1_system_and_rd}
\vspace{-1em}
\end{figure}
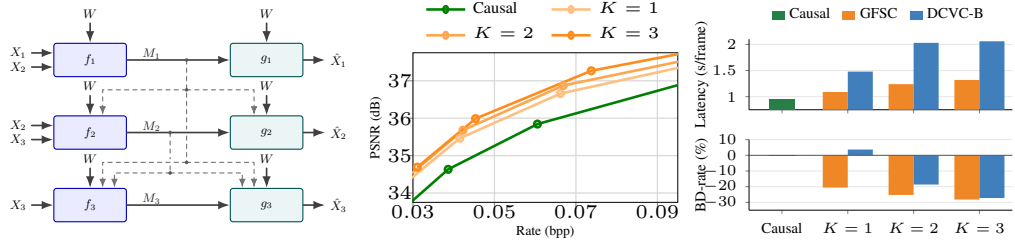

In this work, we propose \emph{Graph Factored Sequential Coding} (GFSC), a principled framework for sequential coding under limited delay.
To our knowledge, we are the first to prove that, for general $K$-order Markov sources with positive distributions, a $K$-frame delay is sufficient to attain joint-coding rate-distortion optimality over a nontrivial distortion region.
Our analysis builds on probabilistic graphical models, where $K$-order Markov dependencies factorize into local clique-wise interactions.
For $K$-order Gauss-Markov sources, the conditional independence structure further reduces to pairwise interactions with linear interpolation sufficient statistics.
\Figref{fig:lisc1_system_and_rd} (left) illustrates the resulting GFSC architecture under one-frame delay.
Although real videos are not assumed to follow an exact Markov model, their temporal dependencies are strongly local. 
As shown in \Figref{fig:infomat_and_curve}, the estimated inverse covariance of video features is approximately banded, with the strongest dependencies concentrated within a small temporal neighborhood.
This empirical locality further motivates exploiting short, configurable delays.

We show that GFSC can be implemented using lightweight neural modules that exploit delayed-frame information to reduce the coding uncertainty of the current frame and improve its rate-distortion tradeoff. 
Only minimal modifications are required to existing causal NVC architectures.
Experiments across multiple models, datasets, and delay settings show consistent rate-distortion improvements over causal baselines and representative B-frame codecs under matched delay budgets.
The main contributions of this work are summarized as follows:

\begin{enumerate}
    \item We propose GFSC, a flexible delayed sequential coding framework that uses clique-wise interactions architecture to systematically exploit a configurable frame-delay budget. 

    \item We establish rigorous rate-distortion guarantees for GFSC. For general $K$-order Markov sources, we prove that $K$-delay GFSC attains joint-coding rate-distortion optimality over a nontrivial distortion region. For $K$-order Gauss-Markov sources, the clique-wise construction further reduces to pairwise linear interpolation.

    \item We realize GFSC using lightweight neural modules that can be integrated into existing causal NVC models with minimal architectural changes. Experiments demonstrate consistent rate-distortion gains over causal baselines and competitive performance against representative B-frame codecs under matched delay budgets.
\end{enumerate}

\input{figs/fig-uvg_precision_correlation}

\section{Preliminaries}
\subsection{Joint coding}
Consider an $L$-frame sequence $\mathbf X \triangleq (X_1, \cdots, X_L)$ where $X_{\ell} \in \mathcal X_{\ell}$ and $\mathbf X \sim P_{\mathbf X}$. 
In joint coding (JC), all frames are available prior to compression. 
We adopt a \emph{one-shot} setup~\citep{li2018strong}, where a single source sequence $\mathbf x$ is compressed at one time. 
Given that encoder and decoder share the common randomness $W \in \mathcal{W}$ independent of source, a JC system is specified by
\begin{equation*}
    f: \mathcal{X}_1 \times \cdots \times \mathcal{X}_L \times \mathcal{W} \to \mathcal{M}, \qquad g: \mathcal{M} \times \mathcal{W} \to \hat{\mathcal{X}}_1 \times \cdots \times \hat{\mathcal{X}}_L,
\end{equation*}
where the message $M \in \mathcal{M}=\{0, 1\}^*$ is a binary string with expected length $\E[l(M)]$. 
The reconstruction $\hat{\mathbf X}$ follows $P_{\hat{\mathbf X} | \mathbf X}(\cdot | \mathbf x)$, i.e., $\hat{\mathbf X} = g(f(\mathbf x, W), W) \sim P_{\hat{\mathbf X} | \mathbf X}(\cdot | \mathbf x)$. 
The reconstruction fidelity for each $X_{\ell}$ is measured by a distortion function $d_{\ell}: \mathcal X_{\ell} \times \mathcal X_{\ell} \to [0, \infty)$, with $d_{\ell}(x, x) = 0$ and $d_{\ell}(x, w) > 0$ for $w \ne x$. 
According to \cite{li2018strong}, there exists a one-shot coding scheme such that minimum achievable $R^{\mathrm{JC},o }$ satisfies:
\begin{equation*}
    R^{\mathrm{JC},o} \le I(\mathbf X; \hat{\mathbf X}) + \mathcal O ( \log I(\mathbf X; \hat{\mathbf X})). 
\end{equation*}
where the mutual information $I(\mathbf X; \hat{\mathbf X})$ is the theoretical optimal solution in the asymptotic setting (infinite blocklength), i.e. JC rate-distortion function~\citep{csiszar2011information}.
\begin{theorem}[JC rate-distortion function]
\label{thm:jc_rd_func}
Define $\mathcal{P} \triangleq \{P_{\hat{\mathbf{X}} | \mathbf{X}} : \mathbb{E} [d_{\ell}( X_{\ell}, \hat{X}_{\ell})] \le D_{\ell}, 1 \le \ell \le L\}$, the single-letter JC rate-distortion function is
\begin{equation}
    R^{\mathrm{JC}}(\mathbf{D}) = \min_{P_{\hat{\mathbf{X}}|\mathbf{X}} \in \mathcal{P}} I(\mathbf{X}; \hat{\mathbf{X}}).
\end{equation}
\end{theorem}

\subsection{Delayed sequential coding}\label{subsec:delayed}
In delayed sequential coding (DSC), source frames are observed sequentially with a limited look-ahead of $K$ frames ($K<L$). 
Following~\cite{ma2007delayed}, there are three canonical setups. 
We focus on the noncausal-causal (NC-C) setting, which is most relevant to NVC, and defer the causal-noncausal (C-NC) and noncausal-noncausal (NC-NC) settings to Appendix~\ref{app-subsec:delayed}. 
An NC-C coding architecture is specified by:
\begin{equation*}
    f_{\ell}:
    \mathcal{X}_1 \times \cdots \times
    \mathcal{X}_{\min\{\ell+K,L\}}
    \times \mathcal{W}
    \to \mathcal{M}_{\ell},
    \qquad
    g_{\ell}:
    \mathcal{M}_1 \times \cdots \times \mathcal{M}_{\ell}
    \times \mathcal{W}
    \to \hat{\mathcal{X}}_{\ell},
\end{equation*}
where $M_{\ell} \in \mathcal{M}_{\ell}$ denotes $\ell$-th encoder message. 
Under $K$-delay relaxation, the $\ell$-th encoder has access up to $X_{\ell+K}$. 
The reconstruction $\hat X_\ell$ follows $P_{\hat X_\ell | X^{\ell+K}}(\cdot | x^{\ell+K})$ determined by $\{f_j, g_j\}_{j=1}^{\ell}$. 
A key result in~\cite{ma2007delayed} shows that, for $K$-order Gauss-Markov sources, a $K$-frame delay is sufficient for DSC to asymptotically attain the joint-coding rate-distortion function in the high-resolution regime.
However, this leaves two gaps.
First, the result is established for the Gauss-Markov setting and does not characterize whether the same optimality extends to general $K$-order Markov sources.
Second, DSC specifies the admissible information flow and its fundamental limits and does not provide an architectural principle for realizing these limits in practical coding systems.

In this work, we address both issues.
We first establish the joint-coding optimality of $K$-delay sequential coding for a broader class of $K$-order Markov sources through a probabilistic graphical model analysis.
We then introduce GFSC as a constructive realization of this structure. 

\section{Graph Factored Sequential Coding}
We introduce \emph{Graph Factored Sequential Coding} (GFSC), inspired by Markov networks. 
We adopt a one-shot setting and analyze the associated rate-distortion performance. 

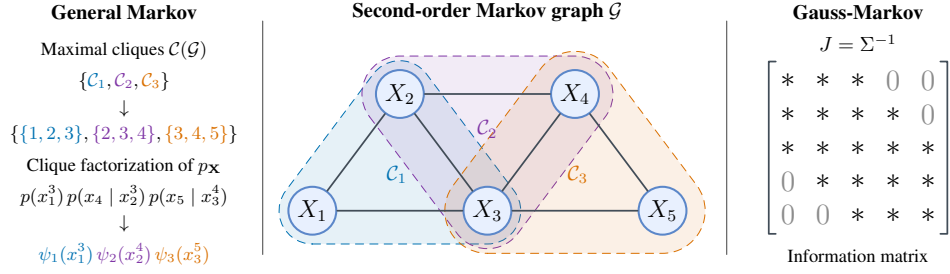
\begin{figure}[!t]
\centering
\begingroup
\definecolor{mkedge}{RGB}{66,77,88}
\providecolor{xdraw}{RGB}{56,108,139}
\providecolor{xfill}{RGB}{224,239,247}
\definecolor{mkcliqueone}{RGB}{0,114,178}
\definecolor{mkcliquetwo}{RGB}{128,55,166}
\definecolor{mkcliquethree}{RGB}{218,112,0}

\resizebox{.9\linewidth}{!}{
\begin{tikzpicture}[
    font=\small,
    >=Latex,
    line cap=round,
    line join=round,
    heading/.style={font=\normalsize\bfseries},
    section_label/.style={font=\small},
    xnode/.style={
        circle,
        draw=xdraw,
        line width=1pt,
        minimum size=0.82cm,
        inner sep=1pt,
        fill=xfill,
        font=\large,
        text=black
    },
    edge_style/.style={line width=0.85pt,draw=mkedge},
    clique_contour/.style={
        line width=0pt,
        dash pattern=on 4pt off 3pt,
        draw opacity=1,
        fill opacity=0.1
    },
    matrix_entry/.style={font=\Large,text=black},
    matrix_zero/.style={font=\Large,text=black!45},
    panel_note/.style={font=\small,text=mkedge}
]

\node[heading] at (2.625,2.90) {General Markov};
\node[section_label] at (2.625,2.25) {Maximal cliques $\mathcal C(\mathcal G)$};
\node at (2.625,1.25) {$
    \begin{gathered}
    \{\textcolor{mkcliqueone}{\mathcal C_1}, \textcolor{mkcliquetwo}{\mathcal C_2}, \textcolor{mkcliquethree}{\mathcal C_3}\} \\
    \downarrow \\
    \{ \textcolor{mkcliqueone}{\{1,2,3\}}, \textcolor{mkcliquetwo}{\{2,3,4\}}, \textcolor{mkcliquethree}{\{3,4,5\}} \}
    \end{gathered}
$};


\node[section_label] at (2.625,0.25) {Clique factorization of $p_{\mathbf X}$};
\node at (2.625,-0.75) {$
    \begin{gathered}
        p(x_1^3)\,p(x_4\mid x_2^3)\,p(x_5\mid x_3^4) \\
        \downarrow \\
        \textcolor{mkcliqueone}{\psi_1(x_1^3)} \,\textcolor{mkcliquetwo}{\psi_2(x_2^4)} \,\textcolor{mkcliquethree}{\psi_3(x_3^5)}
    \end{gathered}
$};

\draw[black!75,line width=0.4pt] (5.0,2.75) -- (5.0,-1.25);
\draw[black!75,line width=0.4pt] (12.95,2.75) -- (12.95,-1.25);

\begin{scope}[xshift=5.85cm,yshift=-0.0cm]
\path[clique_contour,draw=mkcliqueone,fill=mkcliqueone]
    (0,-1.07) -- (3,-1.07)
    arc[start angle=-90,end angle=36.86989765,radius=0.57cm]
    -- (1.956,1.842)
    arc[start angle=36.86989765,end angle=143.13010235,radius=0.57cm]
    -- (-0.456,-0.158)
    arc[start angle=143.13010235,end angle=270,radius=0.57cm]
    -- cycle;

\path[clique_contour,draw=mkcliquetwo,fill=mkcliquetwo]
    (0.98,1.11) -- (2.48,-0.89)
    arc[start angle=-143.13010235,end angle=-36.86989765,radius=0.65cm]
    -- (5.02,1.11)
    arc[start angle=-36.86989765,end angle=90,radius=0.65cm]
    -- (1.50,2.15)
    arc[start angle=90,end angle=216.86989765,radius=0.65cm]
    -- cycle;

\path[clique_contour,draw=mkcliquethree,fill=mkcliquethree]
    (3,-1.23) -- (6,-1.23)
    arc[start angle=-90,end angle=36.86989765,radius=0.73cm]
    -- (5.084,1.938)
    arc[start angle=36.86989765,end angle=143.13010235,radius=0.73cm]
    -- (2.416,-0.062)
    arc[start angle=143.13010235,end angle=270,radius=0.73cm]
    -- cycle;

\node[xnode] (x1) at (0,-0.5)   {$X_1$};
\node[xnode] (x2) at (1.5,1.5) {$X_2$};
\node[xnode] (x3) at (3,-0.5)   {$X_3$};
\node[xnode] (x4) at (4.5,1.5) {$X_4$};
\node[xnode] (x5) at (6,-0.5)   {$X_5$};

\foreach \i/\j in {1/2,1/3,2/3,2/4,3/4,3/5,4/5} {
    \draw[edge_style] (x\i) -- (x\j);
}

\node[text=mkcliqueone,font=\normalsize] at (1.45,0.12) {$\mathcal C_1$};
\node[text=mkcliquetwo,font=\normalsize] at (3.00,0.90) {$\mathcal C_2$};
\node[text=mkcliquethree,font=\normalsize] at (4.55,0.12) {$\mathcal C_3$};

\node[heading] at (3,2.90) {Second-order Markov graph $\mathcal G$};
\end{scope}

\node[heading] at (15.20,2.90) {Gauss-Markov};
\node[font=\normalsize] at (15.20,2.38) {$J=\Sigma^{-1}$};

\begin{scope}[xshift=14.0cm,yshift=0.58cm,x=0.60cm,y=0.72cm]
\draw[edge_style]
    (-0.35,2.05) -- (-0.55,2.05)
    -- (-0.55,-2.05) -- (-0.35,-2.05);
\draw[edge_style]
    (4.35,2.05) -- (4.55,2.05)
    -- (4.55,-2.05) -- (4.35,-2.05);

\foreach \row in {1,...,5} {
    \foreach \col in {1,...,5} {
        \pgfmathtruncatemacro{\outsideband}{abs(\row-\col)>2}
        \ifnum\outsideband=1
            \node[matrix_zero] at ({\col-1},{2.4-0.8*\row}) {$0$};
        \else
            \node[matrix_entry] at ({\col-1},{2.4-0.8*\row}) {$*$};
        \fi
    }
}
\end{scope}
\node[section_label] at (15.20,-1.27) {Information matrix};

\end{tikzpicture}%
}
\endgroup
\caption{
Second-order Markov structure illustration.
\textit{Left}: Maximal cliques and clique factorization.
\textit{Middle}: Second-order Markov network. 
\textit{Right}: The information matrix $J$ for Gauss-Markov satisfying $J_{ij}=0$ for $|i-j|>2$. 
``$*$'' denotes potentially nonzero entries.
}
\label{fig:gauss_markov_graph}
\end{figure}

\subsection{Markov Networks}
\label{subsec:graphical_models}
\begin{definition}[$K$-order Markov]\label{def:korder_markov}
A $K$-order Markov source is a random vector $\mathbf X$ satisfying Markov chains $X_{\ell} \to (X_{\ell+1}, \cdots, X_{\ell+K}) \to (X_{\ell+K+1},\cdots, X_L)$ for all $\ell$.
\label{def:markov}
\end{definition}

\begin{definition}[Maximal clique]
Let $\mathcal G=(\mathcal V,\mathcal E)$ be an undirected graph.
A nonempty subset $\mathcal C\subseteq\mathcal V$ is a \emph{clique} if
$\{i,j\}\in\mathcal E$ for every pair of distinct nodes $i,j\in\mathcal C$.
A clique is \emph{maximal} if it is not strictly contained in any other clique of $\mathcal G$.
\end{definition}

\textbf{Markov networks.}
A Markov network represents the conditional-independence structure of a joint distribution $P_{\mathbf X}$ through an undirected graph $\mathcal G=(\mathcal V,\mathcal E)$, where each node denotes a random variable~\citep{koller2009probabilistic}.
Let $\mathcal C(\mathcal G)$ be the set of maximal cliques of $\mathcal G$. 
For a strictly positive distribution, the Hammersley-Clifford theorem gives the clique factorization of probability density/mass function
$p_{\mathbf X}(\mathbf x) = \frac{1}{Z} \prod_{\mathcal C_j \in \mathcal{C}(\mathcal{G})} \psi_j (x_{\mathcal C_j})$, where $\psi_j$ is a positive clique factor and $Z$ a normalization constant.
For $K$-order Markov sources as in Definition~\ref{def:korder_markov}, the graph has a simple temporal structure.
Its maximal cliques are the consecutive $(K+1)$-variable windows, i.e. $\mathcal C_j \triangleq \{j,\ldots,j+K\}$. 
More importantly, the $K$-order Markov property itself directly yields a clique-wise factorization, without requiring the positivity condition:
\begin{align}
    p_{\mathbf X}(\mathbf x) = p(x^{K+1}) \prod_{j} p(x_{j+K}\mid x_j^{j+K-1}) \triangleq
    \prod_{j} \psi_j(x_{\mathcal C_j}),
    \label{eq:korder_clique_factorization}
\end{align}
where
$\psi_1(x_{\mathcal C_1}) \triangleq p(x^{K+1})$ and $\psi_j(x_{\mathcal C_j}) \triangleq p(x_{j+K} | x_j^{j+K-1})$ for $j\ge 2$.
Hence, the dependency structure of a $K$-order Markov source decomposes into local interactions over $(K+1)$-variable cliques. 
Figure~\ref{fig:gauss_markov_graph} (\textit{Left}) shows the clique factorization for a $2$-order Markov example. 
This observation motivates the construction of GFSC introduced in Section~\ref{subsec:system_model}.

\textbf{Gaussian graphical models (GGMs).}
GGMs specialize Markov networks to Gaussian sources and have been exploited in multi-terminal source coding~\citep{chen2017generalized,chen2020generalized,xie2020generalized}. 
For a Gaussian source $\mathbf X \sim \mathcal{N}(0,\Sigma)$, let $J \triangleq \Sigma^{-1}$ be its information matrix.
The induced GGM $\mathcal{G}(J)$ has edge set
$\mathcal{E} = \{ \{i,j\}: i \ne j, \; J_{ij} \neq 0 \}$.
For a $K$-order Gauss-Markov source, $J$ is $K$-banded, as illustrated in Figure~\ref{fig:gauss_markov_graph} (\textit{Right}).
Moreover, the Gaussian density admits the pairwise factorization
\begin{equation}
    p_{\mathbf X}(\mathbf x) \propto \prod_{\ell} \exp\left( -\frac{1}{2} J_{\ell\ell} x_{\ell}^2 \right) \prod_{ \{i,j\}} \exp\left( -J_{ij} x_i x_j \right).
    \label{eq:gaussian_pairwise_factorization}
\end{equation}
This additional structure leads to the edge-wise and node-wise auxiliary construction and, ultimately, the pairwise interpolation form derived in Section~\ref{subsec:gauss_rd_opt}.

\subsection{Architecture of Coding Scheme}\label{subsec:system_model}
Let $\mathbf X = (X_1, \cdots, X_L)$ be a sequentially observed source and $K$ the allowed delay relaxation ($K < L-1$). 
We define the encoding and decoding mappings of GFSC-$K$ architecture as follows. 

\textbf{Encoding / decoding.} At each stage $\ell$, the encoder observes at most $K$-delay source window $X_{\ell}^{\ell+K}$ and previous messages $M^{\ell-1}$. 
It then produces a new message $M_{\ell}$. 
The $\ell$-th decoder reconstructs $X_{\ell}$ using all messages available up to stage $\ell$. The encoding and decoding mappings are given by:
\begin{equation*}
    f_{\ell}: \mathcal{X}_{\ell}^{\min\{\ell+K, L\} } \times \mathcal{M}^{\ell-1} \times \mathcal{W} \to \mathcal{M}_{\ell}, \qquad g_{\ell}: \mathcal{M}^{\ell} \times \mathcal{W} \to \hat{\mathcal{X}}_{\ell}.
\end{equation*}
There are $L$ decoding stages for $L$ source frames respectively. 
Compared with the NC-C DSC-$K$ architecture, this formulation imposes a local-window structure. 
Figure~\ref{fig:lisc1_system_and_rd} (\textit{left}) and Figure~\ref{fig:ggm_gfsc_architecture} (\textit{Right}) illustrate the structure for $K = 1,2$. 
Now we characterize the operational rate-distortion region. 
\begin{definition}[Operational RD region]\label{def:vf_k_ope_rd}
A rate-distortion tuple $(\mathbf{R}, \mathbf{D})$ is achievable for one-shot GFSC-$K$ system, if there exist encoder and decoder as defined above such that
\begin{align*}
    R_{\ell} \ge \mathbb{E}[l(M_{\ell})], \qquad D_{\ell} \ge \mathbb{E} [d_{\ell}(X_{\ell}, \hat{X}_{\ell}) ],
\end{align*}
where $M_{\ell} \in \mathcal M_{\ell}$ denotes the $\ell$-th encoder message. 
The operational RD region $\mathcal{R}^o_K$ is the closure of the set of all achievable $(\mathbf{R}, \mathbf{D})$. 
For a given $\mathbf{D}$, the rate region $\mathcal{R}^o_K(\mathbf{D})$ consists of all $\mathbf{R}$ such that $(\mathbf{R}, \mathbf{D}) \in \mathcal{R}^o_K$, and the minimum sum-rate function is $R^{\mathrm{GFSC},o}_{K,\mathrm{sum}}(\mathbf{D}) = \min_{\mathbf R \in \mathcal R^o_K(\mathbf D)} \sum R_{\ell}$. 
\end{definition}

\subsection{Rate-distortion analysis}
The operational region $\mathcal{R}^o_{K}$ cannot be directly computed as it involves all possible one-shot encoders and decoders. 
Here we provide a tractable approximation in terms of mutual information, which corresponds to the information RD region. 
We introduce auxiliary variable $U_\ell$, which serves as a single-letter representation of the information conveyed by the $\ell$-th encoder message.

\begin{definition}[Information RD region]\label{def:gfsc_k_ird}
For a given source $\mathbf{X}$, the single-letter information RD region of a GFSC-$K$ system, denoted by $\mathcal{R}_K$, consists of all $(\mathbf{R}, \mathbf{D})$ satisfying
\begin{align}\label{eq:general_ird_constraints}
    R_{\ell} \ge I(X^{\min\{\ell+K,L\}}_{\ell}; U_{\ell} | U^{\ell-1}),  \qquad D_{\ell} \ge \mathbb{E} [d_{\ell}(X_{\ell}, \hat{X}_{\ell})],
\end{align}
for some auxiliary random variables $\mathbf U$ and $\hat{\mathbf X}$ satisfying $\hat{X}_{\ell} = g_{\ell}(U^{\ell})$ and the Markov conditions:
\begin{equation}\label{eq:korder_mc_def}
\begin{gathered}
    U_{\ell} \to (X^{ \min\{\ell+K,L\} }_{\ell}, U^{\ell - 1} ) \to \mathbf{X},
\end{gathered}
\end{equation}
where $g_{\ell}(\cdot)$ are deterministic functions. 
For a given $\mathbf{D}$, the information rate region $\mathcal{R}_{K}(\mathbf{D})$ is the closure of all $\mathbf{R}$ such that $(\mathbf{R}, \mathbf{D}) \in \mathcal{R}_K$, 
and the information minimum sum-rate function is $R^{\mathrm{GFSC}}_{K,\mathrm{sum}}(\mathbf D) = \min_{\mathbf R \in \mathcal R_K(\mathbf D)} \sum R_{\ell}$. 
\end{definition}

\begin{remark}[Role of the auxiliary variables]
The auxiliary variable $U_\ell$ serves as a single-letter representation of the information generated at the $\ell$-th GFSC stage.
For a $K$-order Markov source, the local window observed by this stage is naturally associated with the local clique $\mathcal C_{\ell}$ of the underlying Markov network.
Thus, $U_\ell$ can be viewed as a compressed representation of this clique, while $U^{\ell-1}$ carries information accumulated from preceding overlapping cliques.
The Markov condition in~\eqref{eq:korder_mc_def} ensures that this representation respects the sequential information flow of GFSC.
\end{remark}

\begin{theorem}[Rate-distortion characterization]\label{thm:ope_rd_region}
For a given distortion tuple $\mathbf{D}$ and any $\mathbf{R} \in \mathcal{R}_K(\mathbf{D})$, the operational rate region $\mathcal{R}^o_K(\mathbf{D})$ satisfies $\mathbf{R} + \log(\mathbf{R} + 1) + 5 \in \mathcal{R}^o_K(\mathbf{D}) \subseteq \mathcal{R}_K(\mathbf{D})$. 
Moreover, the operational sum-rate function satisfies
\begin{equation}
    R^{\mathrm{GFSC}}_{K,\mathrm{sum}}(\mathbf{D}) \le R^{\mathrm{GFSC},o}_{K,\mathrm{sum}}(\mathbf{D}) \le R^{\mathrm{GFSC}}_{K,\mathrm{sum}}(\mathbf{D}) + \mathcal{O}\!\left(\log R^{\mathrm{GFSC}}_{K,\mathrm{sum}}(\mathbf{D}) \right).
\end{equation}
\end{theorem}
\begin{proof}
See Appendix~\ref{app-subsec:general_ope_rd_region}.
\end{proof}
Theorem~\ref{thm:ope_rd_region} connects the information sum-rate function to the operational GFSC-$K$ architecture up to a logarithmic one-shot overhead.
We next identify a particular class of clique-wise auxiliary variables that satisfies the GFSC constraints together with stronger Markov conditions (See Appendix~\ref{app-subsec:general_jc_opt} for more details). 
For $K$-order Markov sources with positive distribution, this construction suffices to show that the GFSC information sum-rate function coincides with the JC rate-distortion function in Theorem~\ref{thm:jc_rd_func} over a nonempty distortion region.

\subsection{Joint-Coding Optimality for Markov Sources}
\label{subsec:jc_optimality}
 
\begin{figure}[!t]
\centering
\begingroup
\providecolor{xdraw}{RGB}{56,108,139}
\providecolor{xfill}{RGB}{224,239,247}
\providecolor{udraw}{RGB}{192,117,41}
\providecolor{ufill}{RGB}{255,237,212}
\providecolor{wdraw}{RGB}{52,131,113}
\providecolor{wfill}{RGB}{225,242,233}
\providecolor{edgecol}{RGB}{66,77,88}

\resizebox{.9\linewidth}{!}{
\begin{tikzpicture}[
    font=\small,
    >=Latex,
    line cap=round,
    line join=round,
    source_node/.style={
        circle,draw=xdraw,fill=xfill,line width=0.9pt,
        minimum size=0.60cm,inner sep=0.6pt,
        font=\fontsize{10}{11}\selectfont
    },
    clique_node/.style={
        source_node,draw=udraw,fill=ufill
    },
    refinement_node/.style={
        source_node,draw=wdraw,fill=wfill
    },
    edge_style/.style={
        line width=0.8pt,
    },
    augmentation_arrow/.style={
        -{Latex[length=2.8mm,width=2mm]},
        line width=2.5pt,
        draw=edgecol
    },
    panel_title/.style={
        font=\fontsize{11}{12}\selectfont\bfseries,
    },
    flow_symbol/.style={
        inner sep=1.5pt,outer sep=1pt,
        font=\fontsize{12}{13}\selectfont,text=black
    },
    encoder_block/.style={
        rounded corners=2.5pt,draw=xdraw,fill=xfill,
        line width=0.85pt,minimum width=1.08cm,
        minimum height=0.72cm,inner sep=2pt,
        font=\fontsize{12}{13}\selectfont
    },
    decoder_block/.style={
        encoder_block,draw=wdraw,fill=wfill
    },
    row_label/.style={
        anchor=west,
        font=\fontsize{9.5}{11}\selectfont,
    },
    definition_line/.style={
        anchor=south,inner sep=0pt,outer sep=0pt,
        font=\fontsize{10.5}{12}\selectfont
    },
    flow_arrow/.style={
        -{Latex[length=2.3mm,width=1.6mm]},
        line width=0.8pt
    },
    input_arrow/.style={
        flow_arrow,
        line width=0.72pt
    },
    sequential_arrow/.style={
        -{Latex[length=2.3mm,width=1.6mm]},
        draw=udraw,line width=0.9pt,
        dash pattern=on 3pt off 2.5pt
    }
]

\node[panel_title] at (3.30,5.05) {Markov network augmentation};

\begin{scope}[xshift=0cm,yshift=2.85cm,x=1.1cm,y=1cm]
    \node[source_node] (topx1) at (0,0)      {$X_1$};
    \node[source_node] (topx2) at (1.5,1.55) {$X_2$};
    \node[source_node] (topx3) at (3,0)      {$X_3$};
    \node[source_node] (topx4) at (4.5,1.55) {$X_4$};
    \node[source_node] (topx5) at (6,0)      {$X_5$};

    \foreach \i/\j in {1/2,1/3,2/3,2/4,3/4,3/5,4/5} {
        \draw[edge_style] (topx\i) -- (topx\j);
    }
\end{scope}

\draw[augmentation_arrow] (3.30,2.37) -- (3.30,1.84);

\begin{scope}[xshift=0cm,yshift=-0.15cm,x=1.1cm,y=1cm]
    \node[source_node] (midx1) at (0,0)      {$X_1$};
    \node[source_node] (midx2) at (1.5,1.80) {$X_2$};
    \node[source_node] (midx3) at (3,0)      {$X_3$};
    \node[source_node] (midx4) at (4.5,1.80) {$X_4$};
    \node[source_node] (midx5) at (6,0)      {$X_5$};

    \node[clique_node] (midz1) at (1.5,0.80) {$Z_1$};
    \node[clique_node] (midz2) at (3,1.20)   {$Z_2$};
    \node[clique_node] (midz3) at (4.5,0.80) {$Z_3$};

    \foreach \r/\j in {1/1,1/2,1/3,2/2,2/3,2/4,3/3,3/4,3/5} {
        \draw[edge_style] (midz\r) -- (midx\j);
    }
\end{scope}

\draw[augmentation_arrow] (3.30,-0.64) -- (3.30,-1.17);

\begin{scope}[xshift=0cm,yshift=-4.125cm,x=1.1cm,y=0.90cm]
    \node[source_node] (botx1) at (0,0)      {$X_1$};
    \node[source_node] (botx2) at (1.5,3.25) {$X_2$};
    \node[source_node] (botx3) at (3,0)      {$X_3$};
    \node[source_node] (botx4) at (4.5,3.25) {$X_4$};
    \node[source_node] (botx5) at (6,0)      {$X_5$};

    \node[refinement_node] (botw1) at (0.75,0.57) {$W_1$};
    \node[refinement_node] (botw2) at (1.5,2.20)  {$W_2$};
    \node[refinement_node] (botw3) at (3,0.99)    {$W_3$};
    \node[refinement_node] (botw4) at (4.5,2.20)  {$W_4$};
    \node[refinement_node] (botw5) at (5.25,0.57) {$W_5$};

    \node[clique_node] (botz1) at (1.5,1.12) {$Z_1$};
    \node[clique_node] (botz2) at (3,2.20)   {$Z_2$};
    \node[clique_node] (botz3) at (4.5,1.12) {$Z_3$};

    \foreach \r/\j in {1/1,1/2,1/3,2/2,2/3,2/4,3/3,3/4,3/5} {
        \draw[edge_style] (botz\r) -- (botw\j);
    }
    \foreach \j in {1,...,5} {
        \draw[edge_style] (botx\j) -- (botw\j);
    }
\end{scope}

\coordinate (panelDividerX) at (7.25,0);
\draw[black!75, line width=0.4pt]
    (7.25,5.20) -- (panelDividerX |- botx3.south);

\begin{scope}[xshift=7.75cm,yshift=0cm]

\node[panel_title] at (5.15,5.05) {GFSC with optimal auxiliaries};

\def\ySource{4.40}
\def\yEncoder{2.65}
\def\yAuxiliary{1.30}
\def\yMessage{0.10}
\def\yDecoder{-1.20}
\def\yReconstruction{-2.55}

\node[row_label] at (0,\ySource) {Sources};
\node[row_label] at (0,\yEncoder) {Encoders};
\node[row_label] at (0,\yAuxiliary) {Auxiliaries};
\node[row_label] at (0,\yMessage) {Messages};
\node[row_label] at (0,\yDecoder) {Decoders};
\node[row_label] at (0,\yReconstruction) {Reconstructions};

\foreach \j/\xc in {1/2.90,2/4.60,3/6.30,4/8.00,5/9.70} {
    \node[flow_symbol] (src\j) at (\xc,\ySource) {$X_{\j}$};
    \node[encoder_block] (enc\j) at (\xc,\yEncoder) {$f_{\j}$};
    \node[flow_symbol,text=udraw] (aux\j) at (\xc,\yAuxiliary) {$U_{\j}$};
    \node[flow_symbol] (msg\j) at (\xc,\yMessage) {$M_{\j}$};
    \node[decoder_block] (dec\j) at (\xc,\yDecoder) {$g_{\j}$};
    \node[flow_symbol] (rec\j) at (\xc,\yReconstruction) {$\hat X_{\j}$};
}

\foreach \j in {1,...,5} {
    \draw[input_arrow] (src\j.south) -- (enc\j.north);
}
\foreach \j/\k in {1/2,2/3,3/4,4/5} {
    \draw[input_arrow] (src\k.south) -- (enc\j.north);
}
\foreach \j/\k in {1/3,2/4,3/5} {
    \draw[input_arrow] (src\k.south) -- (enc\j.north);
}

\foreach \j in {1,...,5} {
    \draw[flow_arrow] (enc\j.south) -- (aux\j.north);
    \draw[flow_arrow] (aux\j.south) -- (msg\j.north);
    \draw[flow_arrow] (msg\j.south) -- (dec\j.north);
    \draw[flow_arrow] (dec\j.south) -- (rec\j.north);
}

\foreach \j/\k in {1/2,2/3,3/4,4/5} {
    \draw[sequential_arrow] (aux\j.east) -- (aux\k.west);
}

\draw[sequential_arrow] (3.25,-3.17) -- (4.0,-3.17);
\node[row_label] at (4.4,-3.17) {Sequential coding order};

\coordinate (rightFooterCenter) at (5.15,0);
\node[definition_line] (auxiliaryDefinitions)
    at (rightFooterCenter |- botx3.south) {$
        U_4\triangleq W_4\qquad\qquad U_5\triangleq W_5
$};
\node[definition_line] at ($(auxiliaryDefinitions.north)+(0,0.25)$) {$
    U_1 \triangleq (W_1,Z_1)\qquad
    U_2 \triangleq (W_2,Z_2)\qquad
    U_3 \triangleq (W_3,Z_3)
$};

\end{scope}
\end{tikzpicture}
}
\endgroup
\caption{
\textit{Left}: Second-order Markov network (top), its augmentation by clique-wise auxiliary variables $Z_j$ (middle), and further augmentation by node-wise auxiliary variables $W_{\ell}$ (bottom). 
\textit{Right}: GFSC architecture with $2$-delay and optimal auxiliary variables constructed in \textit{Left}.
}
\vspace{-1em}
\label{fig:ggm_gfsc_architecture}
\end{figure}
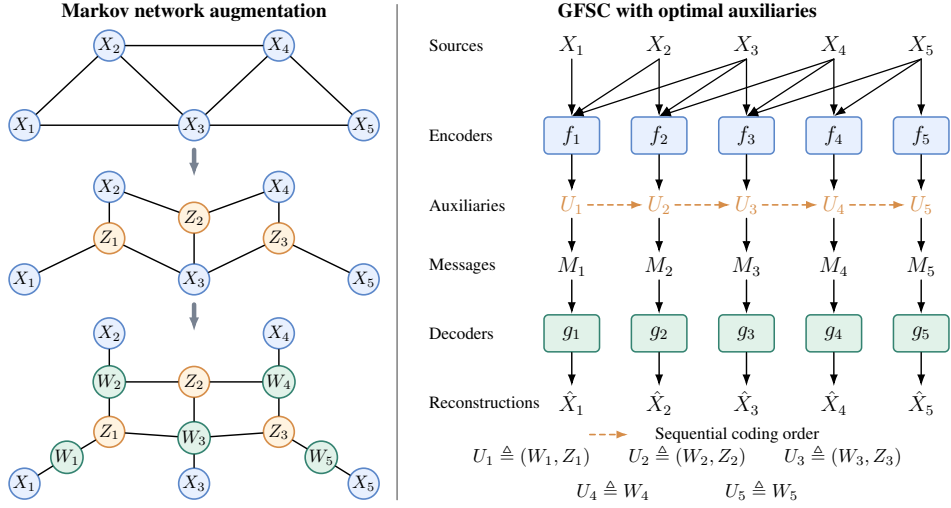

\subsubsection{General Markov sources}
\label{subsec:general_markov_analysis}

As discussed in Section~\ref{subsec:graphical_models}, a $K$-order Markov source factorizes over consecutive cliques $\mathcal C_j=\{j,\ldots,j+K\}$.
This graphical structure motivates a clique-wise auxiliary construction: each auxiliary variable is generated from the source variables within one local clique, while the collection of overlapping auxiliaries captures the global source dependence through local interactions.
For the optimality proof, we construct a particular class of such auxiliaries satisfying stronger Markov conditions than those required by GFSC.
These conditions preserve the relevant dependency structure while remaining admissible under the GFSC-$K$ information-flow constraints.

\begin{definition}[$K$-order Markov source with positive distribution]
\label{def:markov_pos_dist}
Let $\mathbf X$ be a $K$-order Markov source as in Definition~\ref{def:markov}. 
Moreover, each $X_j$ has finite alphabet $2 \le |\mathcal X_j| < \infty$ and the joint distribution is positive, i.e. $p_{\mathbf X}(\mathbf x)>0$ for all $\mathbf x\in\mathcal X_1\times\cdots\times\mathcal X_L$.
\end{definition}
\begin{theorem}[JC optimality]
\label{thm:general_main_thm}
Consider a Markov source $\mathbf X$ as in Definition~\ref{def:markov_pos_dist}. 
There exists a non-empty distortion region $\mathcal D_{\epsilon} \triangleq \{ \mathbf D: 0 < D_{\ell} \le \epsilon_{\ell}, 1 \le \ell \le L \ \}$, such that for all $\mathbf D \in \mathcal D_{\epsilon}$, a set of clique-wise auxiliaries $\mathbf U$ can be constructed for GFSC-$K$ and the following hold:
\begin{enumerate}
    \item \(\displaystyle 
        U_{\ell} \to (X^{\min\{\ell +K, L\}}_{\ell}, U^{\min\{\ell, L-K\}}_{\max\{1, \ell - K\} }) \to \mathbf X.
    \)
    \item \(\displaystyle
        R^{\mathrm{GFSC}}_{K,\mathrm{sum}}(\mathbf D) = R^{\mathrm{JC}}(\mathbf D)
    \), where $R^{\mathrm{JC}}(\mathbf D)$ is JC rate-distortion function in Theorem~\ref{thm:jc_rd_func}.
\end{enumerate}
\end{theorem}
\begin{proof}
    See Appendix~\ref{app-subsec:general_jc_opt}.
\end{proof}

\begin{remark}[Graphical interpretation]
To our knowledge, Theorem~\ref{thm:general_main_thm} is the first rigorous result showing that $K$-delay coding can attain JC rate-distortion optimality for general $K$-order Markov sources.
The key is the locality of the Markov network: clique-wise auxiliary variables compress local dependencies while preserving the conditional-independence structure required for optimal coding.
Figure~\ref{fig:ggm_gfsc_architecture} explicitly illustrates this construction, where clique-wise auxiliary variables $Z_{\ell}$ induce the desired conditional independencies and node-wise auxiliary variables $W_{\ell}$ further refine individual reconstructions.
This locality is also consistent with real video, whose estimated inverse covariance in Figure~\ref{fig:infomat_and_curve} is largely banded, with the strongest dependencies concentrated within about three off-diagonals.
In practice, we show that these local clique-wise interactions can be efficiently realized in NVC using learnable neural modules.
\end{remark}

\begin{figure}[!t]
\centering
\begingroup
\providecolor{xdraw}{RGB}{56,108,139}
\providecolor{xfill}{RGB}{224,239,247}
\providecolor{wdraw}{RGB}{52,131,113}
\providecolor{wfill}{RGB}{225,242,233}
\providecolor{udraw}{RGB}{192,117,41}
\providecolor{ufill}{RGB}{255,237,212}
\providecolor{edgecol}{RGB}{66,77,88}
\resizebox{.9\linewidth}{!}{
\begin{tikzpicture}[
    font=\small,
    >=Latex,
    line cap=round,
    line join=round,
    xnode/.style={
        circle, draw=xdraw, very thick,
        minimum size=0.5cm, fill=xfill,
        font=\large
    },
    unode/.style={
        circle, draw=udraw, very thick,
        minimum size=0.5cm, fill=ufill,
        font=\small
    },
    edge_style/.style={
        line width=0.8pt,
    },
    dashed_arrow/.style={thick, draw=edgecol, dashed, ->}
]

\begin{scope}[xshift=0cm, yshift=0cm, x=1.1cm, y=1.1cm]
    \node[xnode] (a1) at (0,0)   {$X_1$};
    \node[xnode] (a2) at (1.5,2) {$X_2$};
    \node[xnode] (a3) at (3,0)   {$X_3$};
    \node[xnode] (a4) at (4.5,2) {$X_4$};
    \node[xnode] (a5) at (6,0)   {$X_5$};

    \draw[edge_style] (a1) -- (a2);
    \draw[edge_style] (a1) -- (a3);
    \draw[edge_style] (a2) -- (a3);
    \draw[edge_style] (a2) -- (a4);
    \draw[edge_style] (a3) -- (a4);
    \draw[edge_style] (a3) -- (a5);
    \draw[edge_style] (a4) -- (a5);
\end{scope}

\draw[->, line width=2.0pt, draw=edgecol]
    (7.10,1.10) -- (7.70,1.10);

\begin{scope}[xshift=8.20cm, yshift=0cm, x=1.1cm, y=1.1cm]
    \node[xnode] (b1) at (0,0)   {$X_1$};
    \node[xnode] (b2) at (1.5,2) {$X_2$};
    \node[xnode] (b3) at (3,0)   {$X_3$};
    \node[xnode] (b4) at (4.5,2) {$X_4$};
    \node[xnode] (b5) at (6,0)   {$X_5$};

    \draw[edge_style] (b1) -- (b2)
        node[midway, unode] (u12) {$U_{1,2}$};
    \draw[edge_style] (b1) -- (b3)
        node[midway, unode] (u13) {$U_{1,3}$};
    \draw[edge_style] (b2) -- (b3)
        node[midway, unode] (u23) {$U_{2,3}$};
    \draw[edge_style] (b2) -- (b4)
        node[midway, unode] (u24) {$U_{2,4}$};
    \draw[edge_style] (b3) -- (b4)
        node[midway, unode] (u34) {$U_{3,4}$};
    \draw[edge_style] (b3) -- (b5)
        node[midway, unode] (u35) {$U_{3,5}$};
    \draw[edge_style] (b4) -- (b5)
        node[midway, unode] (u45) {$U_{4,5}$};

    \draw[dashed_arrow] (u12) -- (u13);
    \draw[dashed_arrow] (u13) -- (u23);
    \draw[dashed_arrow] (u23) -- (u24);
    \draw[dashed_arrow] (u24) -- (u34);
    \draw[dashed_arrow] (u34) -- (u35);
    \draw[dashed_arrow] (u35) -- (u45);
\end{scope}

\draw[draw=black!75, line width=0.4pt]
    (15.5,-0.80) -- (15.5,2.90);

\begin{scope}[xshift=15.6cm]
    \node[xnode] (localj) at (0.65,2.45) {$X_j$};
    \node[unode] (localu) at (3.45,2.45) {$U_{j,k}$};
    \node[xnode] (localk) at (6.25,2.45) {$X_k$};
    \draw[edge_style] (localj) -- (localu) -- (localk);

    \node[font=\normalsize, inner sep=0pt] (auxformula)
        at (3.45,1.05) {$
        \begin{aligned}
        U_{j,k} &= \omega_{j,k}X_j+ \bar\omega_{j,k}X_k + N_{j,k} \\
                &= \tilde{X}_{j,k} + N_{j,k}.
        \end{aligned}
        $};
    \draw[-{Latex[length=1.8mm,width=1.4mm]}, thick, draw=edgecol, shorten >=1pt, shorten <=1pt]
        (localu.south) -- (auxformula.north);

    \node[
        draw=wdraw!75,
        fill=wfill,
        line width=0.8pt,
        rounded corners=2pt,
        minimum width=3.00cm,
        minimum height=0.85cm,
        inner sep=3pt,
        align=center,
        font=\small
    ] (interpolation) at (3.45,-0.30) {Linear\\interpolation};

    \node[font=\large, inner sep=1pt] (inputj)
        at (0.65,-0.03) {$X_j$};
    \node[font=\large, inner sep=1pt] (inputk)
        at (0.65,-0.57) {$X_k$};
    \node[font=\large, inner sep=1pt] (interpolated)
        at (6.25,-0.30) {$\tilde{X}_{j,k}$};

    \draw[->, thick, draw=edgecol]
        (inputj.east) -- ($(interpolation.west)+(0,0.27)$);
    \draw[->, thick, draw=edgecol]
        (inputk.east) -- ($(interpolation.west)+(0,-0.27)$);
    \draw[->, thick, draw=edgecol]
        (interpolation.east) -- (interpolated.west);
    \draw[-{Latex[length=1.8mm,width=1.4mm]}, thick, draw=edgecol, shorten >=1pt, shorten <=1pt]
        (auxformula.south) -- (interpolation.north);
\end{scope}

\end{tikzpicture}
}
\endgroup
\caption{\textit{Left}: A second-order GGM and its augmented representation
within $\mathcal D^{\mathrm{G}}_{\epsilon}$; dashed arrows indicate sequential coding.
\textit{Right}: Edge-wise auxiliary construction by linear interpolation
and additive noise, where $\bar{\omega}_{j,k}=1-\omega_{j,k}$.}
\label{fig:ggm_and_linear_interp}
\end{figure}
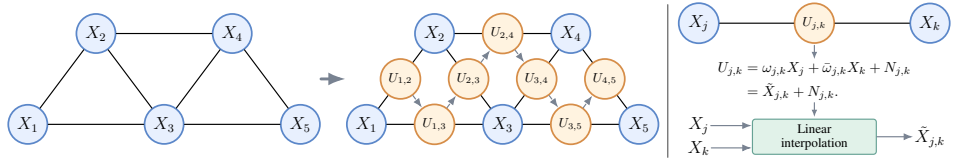
\subsection{Gauss-Markov sources under MSE}
\label{subsec:gauss_rd_opt}

The general result above relies on clique-wise interactions of the underlying Markov network.
Under the Gauss-Markov assumption and MSE distortion, the graphical structure further decomposes into edge-wise interactions as discussed in Section~\ref{subsec:graphical_models}, allowing the clique-wise auxiliaries to be replaced by pairwise Gaussian auxiliaries. 

\begin{theorem}[Gauss-Markov specialization]
\label{thm:jc_opt}
Consider a $K$-order Gauss-Markov source $\mathbf X$ under MSE distortion.
There exists a nonempty distortion region $\mathcal D^{\mathrm G}_{\epsilon}$ such that
\begin{equation}
    R^{\mathrm{GFSC-G}}_{K,\mathrm{sum}}(\mathbf D) =
    R^{\mathrm{JCG}}(\mathbf D), \qquad \forall\,\mathbf D\in\mathcal D^{\mathrm G}_{\epsilon}.
\end{equation}
Moreover, the optimal clique-wise construction can be realized using pairwise Gaussian auxiliaries associated with the edges of the underlying GGM, with linear interpolations as sufficient statistics.
\end{theorem}

\begin{proof}
See Appendix~\ref{app:jc_optimality}.
\end{proof}
\vspace{-.5em}

Gaussianity is not required for GFSC, but provides a particularly simple interpretation of its optimal auxiliary structure: $(K+1)$-variable clique interactions reduce to ordered pairwise interactions within the $K$-frame neighborhood.
Figure~\ref{fig:ggm_and_linear_interp} illustrates this decomposition for a second-order Gauss-Markov source.
Representative characterizations of $\mathcal D^{\mathrm G}_{\epsilon}$ for first-order Gauss-Markov example are provided in Appendix~\ref{app-sec:gauss_example}.

\section{Experimental Results}\label{sec:exp}
We conduct experiments on several NVC models based on causal predictive coding, including the DCVC~\citep{li2021deep}, 
video compression transformer (VCT)~\citep{mentzer2022vct}, and DHVC~\citep{lu2024deep}.  
We compare the rate-distortion performance of the GFSC extension and its causal counterparts. 
We further compare against both neural and conventional delayed coding baselines (bidirectional) under the same delay constraints. 
For quantitative comparison, we report BD-rate results on UVG~\citep{mercat2020uvg} and MCL-JCV~\citep{wang2016mcl}. 

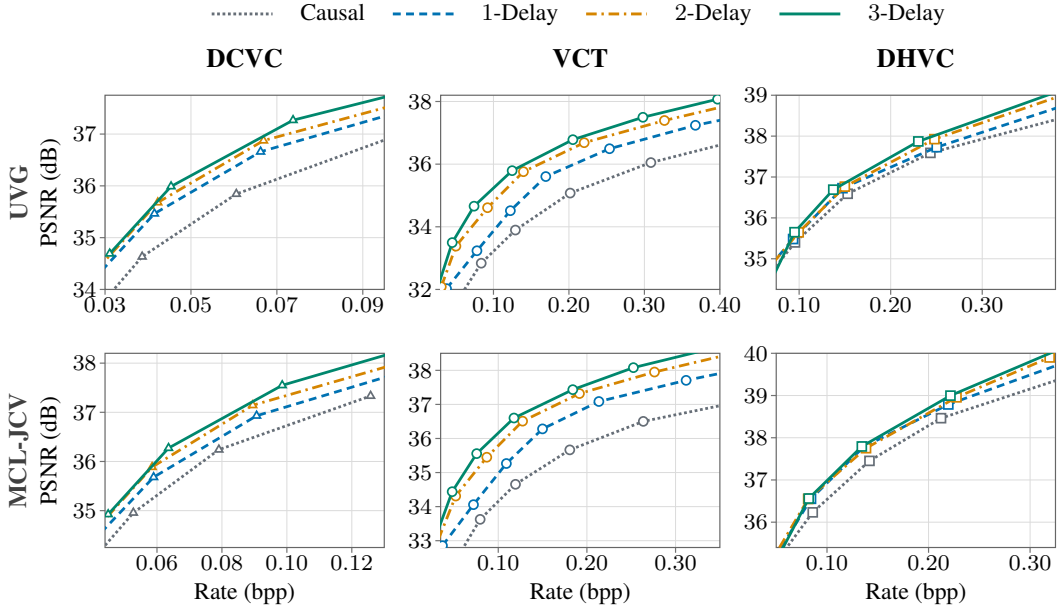
\begin{figure*}[t]
\centering
\begingroup
\definecolor{rdCausalColor}{RGB}{102,110,119}
\definecolor{rdDelayOneColor}{RGB}{0,114,178}
\definecolor{rdDelayTwoColor}{RGB}{213,133,0}
\definecolor{rdDelayThreeColor}{RGB}{0,135,109}

\pgfplotsset{
    compat=1.18,
    rd causal/.style={color=rdCausalColor,densely dotted},
    rd delay one/.style={color=rdDelayOneColor,densely dashed},
    rd delay two/.style={
        color=rdDelayTwoColor,
        dash pattern=on 4pt off 1.8pt on 0.9pt off 1.8pt
    },
    rd delay three/.style={color=rdDelayThreeColor,solid},
}

\resizebox{\linewidth}{!}{%
\begin{tikzpicture}
\begin{groupplot}[
    group style={
        group name=rdplots,
        group size=3 by 2,
        horizontal sep=0.75cm,
        vertical sep=0.85cm,
    },
    scale only axis,
    width=0.266\linewidth,
    height=0.185\linewidth,
    axis background/.style={fill=white},
    axis line style={black!60,line width=0.45pt},
    grid=major,
    major grid style={black!14,line width=0.3pt},
    tick align=outside,
    tick pos=left,
    major tick length=2pt,
    tick style={black!55,line width=0.45pt},
    tick label style={font=\fontsize{8.5}{9.5}\selectfont,inner sep=1pt},
    scaled x ticks=false,
    scaled y ticks=false,
    xticklabel style={
        /pgf/number format/fixed,
        /pgf/number format/precision=2,
        /pgf/number format/fixed zerofill,
    },
    yticklabel style={
        /pgf/number format/fixed,
        /pgf/number format/precision=0,
    },
    xlabel style={font=\fontsize{9}{10}\selectfont},
    ylabel style={font=\fontsize{9}{10}\selectfont},
    title style={font=\fontsize{10}{11}\selectfont\bfseries,yshift=1pt},
    every axis plot/.append style={
        line width=1.05pt,
        mark size=1.7pt,
        mark options={solid,fill=white,line width=0.65pt},
    },
    clip=true,
    clip marker paths=true,
]

\nextgroupplot[
    title={DCVC},
    xmin=0.03, xmax=0.095, ymin=34.0, ymax=37.75,
    xtick={0.03,0.05,0.07,0.09}, ytick={34,35,36,37},
    ylabel={PSNR (dB)},
    legend to name=rd-main-delay-legend,
    legend columns=4,
    legend style={
        draw=none,fill=none,
        font=\fontsize{9}{10}\selectfont,
        /tikz/column sep=0.28cm,
        legend cell align=left,
    },
    legend image post style={xscale=1.35},
]
\addplot[rd causal,mark=triangle*,forget plot] coordinates {
    (0.02567,33.3877)
    (0.03864,34.6321)
    (0.06056,35.8432)
    (0.09874,36.9968)
};
\addplot[rd delay one,mark=triangle*,forget plot] coordinates {
    (0.02713,34.1748)
    (0.04150,35.4637)
    (0.06621,36.6592)
    (0.11237,37.7571)
};
\addplot[rd delay two,mark=triangle*,forget plot] coordinates {
    (0.02807,34.3799)
    (0.04228,35.6792)
    (0.06689,36.8741)
    (0.11286,37.9083)
};
\addplot[rd delay three,mark=triangle*,forget plot] coordinates {
    (0.03108,34.6910)
    (0.04537,35.9891)
    (0.07377,37.2683)
    (0.12487,38.3358)
};
\addlegendimage{rd causal,mark=none,line width=1.1pt}
\addlegendentry{Causal}
\addlegendimage{rd delay one,mark=none,line width=1.1pt}
\addlegendentry{$1$-Delay}
\addlegendimage{rd delay two,mark=none,line width=1.1pt}
\addlegendentry{$2$-Delay}
\addlegendimage{rd delay three,mark=none,line width=1.1pt}
\addlegendentry{$3$-Delay}

\nextgroupplot[
    title={VCT},
    xmin=0.03, xmax=0.4, ymin=32, ymax=38.2,
    xtick={0.1,0.2,0.3,0.4}, ytick={32,34,36,38},
]
\addplot[rd causal,mark=*,forget plot] coordinates {
    (0.04997,31.55721)
    (0.08380,32.83787)
    (0.12906,33.89581)
    (0.20167,35.07653)
    (0.30855,36.05025)
    (0.44321,36.87307)
    (0.56906,37.42549)
};
\addplot[rd delay one,mark=*,forget plot] coordinates {
    (0.03780,32.02657)
    (0.07812,33.23532)
    (0.12240,34.51421)
    (0.16933,35.60394)
    (0.25381,36.49238)
    (0.36760,37.24016)
    (0.47178,37.75164)
};
\addplot[rd delay two,mark=*,forget plot] coordinates {
    (0.03089,32.08654)
    (0.05025,33.38226)
    (0.09215,34.60941)
    (0.13970,35.75925)
    (0.22019,36.68484)
    (0.32655,37.39328)
    (0.42379,37.94642)
};
\addplot[rd delay three,mark=*,forget plot] coordinates {
    (0.02761,32.17103)
    (0.04535,33.49548)
    (0.07430,34.65797)
    (0.12476,35.79216)
    (0.20522,36.78151)
    (0.29802,37.49588)
    (0.39694,38.07374)
};

\nextgroupplot[
    title={DHVC},
    xmin=0.075, xmax=0.38, ymin=34.25, ymax=39.0,
    xtick={0.1,0.2,0.3}, ytick={35,36,37,38,39},
]
\addplot[rd causal,mark=square*,forget plot] coordinates {
    (0.05993,34.40)
    (0.09547,35.40)
    (0.15302,36.59)
    (0.24340,37.59)
    (0.40850,38.57)
};
\addplot[rd delay one,mark=square*,forget plot] coordinates {
    (0.05877,34.57654)
    (0.09344,35.48281)
    (0.14734,36.72710)
    (0.25027,37.72904)
    (0.39144,38.76438)
};
\addplot[rd delay two,mark=square*,forget plot] coordinates {
    (0.05587,34.47966)
    (0.09953,35.64613)
    (0.14994,36.76464)
    (0.24776,37.92631)
    (0.38073,38.94710)
};
\addplot[rd delay three,mark=square*,forget plot] coordinates {
    (0.06481,34.24572)
    (0.095,35.65)
    (0.13729,36.69049)
    (0.23018,37.87000)
    (0.38016,39.07409)
};

\nextgroupplot[
    xmin=0.044, xmax=0.130, ymin=34.25, ymax=38.2,
    xtick={0.06,0.08,0.10,0.12}, ytick={34,35,36,37,38},
    xlabel={Rate (bpp)}, ylabel={PSNR (dB)},
]
\addplot[rd causal,mark=triangle*,forget plot] coordinates {
    (0.03530,33.6282)
    (0.05272,34.9562)
    (0.07906,36.2403)
    (0.12579,37.3329)
};
\addplot[rd delay one,mark=triangle*,forget plot] coordinates {
    (0.03995,34.3566)
    (0.05896,35.6799)
    (0.09068,36.9297)
    (0.14295,37.9653)
};
\addplot[rd delay two,mark=triangle*,forget plot] coordinates {
    (0.04006,34.5374)
    (0.05850,35.8857)
    (0.08949,37.1432)
    (0.14199,38.1420)
};
\addplot[rd delay three,mark=triangle*,forget plot] coordinates {
    (0.04497,34.9252)
    (0.06356,36.2726)
    (0.09857,37.5493)
    (0.15096,38.5614)
};

\nextgroupplot[
    xmin=0.035, xmax=0.35, ymin=32.8, ymax=38.5,
    xtick={0.1,0.2,0.3}, ytick={33,34,35,36,37,38},
    xlabel={Rate (bpp)},
]
\addplot[rd causal,mark=*,forget plot] coordinates {
    (0.04936,32.30293)
    (0.07960,33.62650)
    (0.11945,34.65480)
    (0.18070,35.66566)
    (0.26361,36.50201)
    (0.38521,37.14819)
    (0.49728,37.67031)
};
\addplot[rd delay one,mark=*,forget plot] coordinates {
    (0.03707,32.85395)
    (0.07205,34.05769)
    (0.10917,35.26500)
    (0.14994,36.28185)
    (0.21353,37.08527)
    (0.31191,37.70641)
    (0.40297,38.18966)
};
\addplot[rd delay two,mark=*,forget plot] coordinates {
    (0.03226,33.02106)
    (0.05200,34.31004)
    (0.08708,35.44975)
    (0.12746,36.50837)
    (0.19160,37.31885)
    (0.27623,37.95020)
    (0.35670,38.44654)
};
\addplot[rd delay three,mark=*,forget plot] coordinates {
    (0.02951,33.11501)
    (0.04802,34.43818)
    (0.07567,35.55590)
    (0.11771,36.60079)
    (0.18408,37.43351)
    (0.25245,38.07912)
    (0.33351,38.59466)
};

\nextgroupplot[
    xmin=0.05, xmax=0.325, ymin=35.4, ymax=40.0,
    xtick={0.1,0.2,0.3}, ytick={36,37,38,39,40},
    xlabel={Rate (bpp)},
]
\addplot[rd causal,mark=square*,forget plot] coordinates {
    (0.05541,35.23831)
    (0.08596,36.23277)
    (0.14182,37.45190)
    (0.21228,38.46342)
    (0.34421,39.50359)
};
\addplot[rd delay one,mark=square*,forget plot] coordinates {
    (0.05492,35.39925)
    (0.08426,36.55334)
    (0.13609,37.78316)
    (0.21946,38.78806)
    (0.32743,39.72442)
};
\addplot[rd delay two,mark=square*,forget plot] coordinates {
    (0.05314,35.39700)
    (0.08143,36.55041)
    (0.13844,37.75236)
    (0.22753,38.95721)
    (0.31926,39.90146)
};
\addplot[rd delay three,mark=square*,forget plot] coordinates {
    (0.05221,35.18501)
    (0.08200,36.56)
    (0.13428,37.79318)
    (0.22158,38.99765)
    (0.32709,40.07794)
};

\end{groupplot}

\coordinate (rdTopCenter) at
    ($(rdplots c1r1.north west)!0.5!(rdplots c3r1.north east)$);
\node[anchor=south,inner sep=0pt]
    at ($(rdTopCenter)+(0,0.80cm)$)
    {\pgfplotslegendfromname{rd-main-delay-legend}};

\node[rotate=90,font=\fontsize{9.5}{10.5}\selectfont\bfseries,text=black!75]
    at ($(rdplots c1r1.west)+(-1.15cm,0)$) {UVG};
\node[rotate=90,font=\fontsize{9.5}{10.5}\selectfont\bfseries,text=black!75]
    at ($(rdplots c1r2.west)+(-1.15cm,0)$) {MCL-JCV};

\end{tikzpicture}%
}
\endgroup
\caption{
Rate-PSNR comparison between causal and GFSC-enhanced models on
UVG and MCL-JCV dataset. Columns show DCVC, VCT, and DHVC, respectively.
}
\label{fig:rd_baseline}
\vspace{-.5em}
\end{figure*}

\textbf{Clique-wise interaction.}
For a $K$-delay setting, GFSC operates on the clique-wise latent
$(\mathbf y_\ell,\ldots,\mathbf y_{\ell+K})$, where $\mathbf y_{\ell} = f_{\ell}(\mathbf x_{\ell})$ denotes the latent representation of frame $\mathbf x_{\ell}$. 
We implement this interaction using a window-based spatial-temporal attention module~\cite{liu2022video}, which jointly models the current and lookahead latent representations within each local spatial window. 
This enables configurable $K$-delay fusion without modifying the underlying analysis transforms and naturally handles boundary cases with fewer than $K$ available future frames.
Architectural details are provided in Appendix~\ref{app-subsec:net_arch}.

\subsection{Rate-distortion analysis}\label{subsec:main_res}
\textbf{Causal counterparts.} We evaluate GFSC on three causal NVC models. 
Figure~\ref{fig:rd_baseline} shows the rate-PSNR curves for the baselines and their GFSC-enhanced variants with delays $K=1, 2, 3$.
Across all models, incorporating the delay window interaction consistently improves rate-distortion performance, with gains increasing as $K$ grows. 
Notably, even a single-frame delay yields significant improvement.  
Table~\ref{tab:bd_rate} further quantifies the BD-rate improvements. 
Across all datasets and models, GFSC achieves consistent bitrate reductions relative to causal baselines, demonstrating the effectiveness of limited delay for improving rate-distortion performance. 
See Appendix~\ref{app-subsec:add} for results on 4K video dataset.


\begin{wraptable}{r}{0.55\textwidth}
\centering
\caption{BD-rate (\%) of GFSC-enhanced models relative to causal
baselines. 
}
\label{tab:bd_rate}

\setlength{\tabcolsep}{3pt}
\renewcommand{\arraystretch}{1.05}
\resizebox{\linewidth}{!}{%
\begin{tabular}{|l|ccc|ccc|}
\hline
 & \multicolumn{3}{c|}{UVG ($1080$p)}
 & \multicolumn{3}{c|}{MCL-JCV ($1080$p)} \\
\cline{2-7}
Delay & DCVC & VCT & DHVC & DCVC & VCT & DHVC \\
\hline
$K=1$ & -20.57 & -28.38 & -7.39
      & -10.90 & -32.35 & -13.35 \\
$K=2$ & -25.29 & -48.00 & -10.17
      & -17.32 & -50.14 & -14.65 \\
$K=3$ & -28.22 & -55.78 & -11.91
      & -20.30 & -56.70 & -16.07 \\
\hline
\end{tabular}%
}
\vspace{-0.8em}
\end{wraptable}

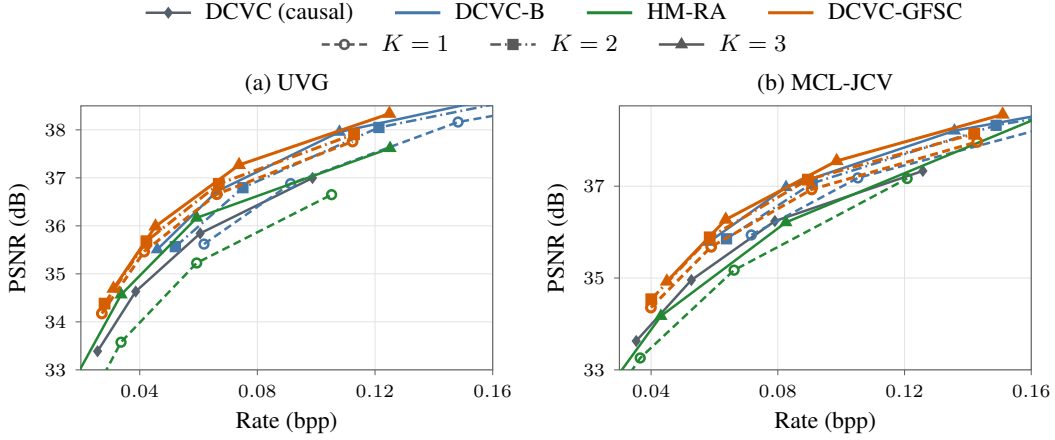
\begin{figure}[!t]
\centering
\begingroup

\definecolor{rdCausal}{HTML}{56606B}
\definecolor{rdB}{HTML}{4477AA}
\definecolor{rdHM}{HTML}{228833}
\definecolor{rdGFSC}{HTML}{D55E00}

\pgfplotsset{
    delayedRDbase/.style={
        line width=0.95pt,
        mark size=1.65pt,
        mark options={solid, line width=0.65pt},
    },
    delayedRDone/.style={
        mark=o,
        dash pattern=on 3pt off 1.7pt,
        mark options={solid, fill=white},
    },
    delayedRDtwo/.style={
        mark=square*,
        dash pattern=on 4pt off 1.5pt on 0.7pt off 1.5pt,
        mark options={solid},
    },
    delayedRDthree/.style={
        mark=triangle*, solid,
        mark size=1.95pt,
        mark options={solid},
    },
    delayedRDlegend/.style={
        draw=none,
        fill=none,
        font=\footnotesize,
        inner sep=0pt,
        column sep=3pt,
        /tikz/every even column/.append style={column sep=13pt},
        cells={anchor=west},
    },
}

\begin{tikzpicture}
\begin{groupplot}[
    group style={
        group name=delayedRD,
        group size=2 by 1,
        horizontal sep=0.12\linewidth,
    },
    scale only axis,
    width=0.39\linewidth,
    height=0.25\linewidth,
    xmin=0.02, xmax=0.16,
    xtick={0.04,0.08,0.12,0.16},
    scaled x ticks=false,
    xticklabel style={
        /pgf/number format/.cd,
        fixed, fixed zerofill, precision=2,
    },
    xlabel={Rate (bpp)},
    ylabel={PSNR (dB)},
    xlabel style={font=\footnotesize, yshift=1pt},
    ylabel style={font=\footnotesize, yshift=-1pt},
    tick label style={font=\scriptsize},
    axis line style={draw=black!60, line width=0.45pt},
    tick style={draw=black!60, line width=0.4pt},
    tick align=outside,
    tick pos=left,
    major tick length=2pt,
    grid=major,
    major grid style={draw=black!11, line width=0.3pt},
    axis background/.style={fill=white},
    clip=true,
    clip mode=individual,
    every axis plot/.append style={delayedRDbase},
]
\nextgroupplot[
    ymin=33.0, ymax=38.5,
    ytick={33,34,35,36,37,38},
    legend to name=delayedRDModels,
    legend columns=4,
    legend style={delayedRDlegend},
]
\addplot[color=rdCausal, solid, mark=diamond*, mark size=1.8pt, forget plot] coordinates {
    (0.02567,33.3877)
    (0.03864,34.6321)
    (0.06056,35.8432)
    (0.09874,36.9968)
};

\addplot[color=rdB, delayedRDone, forget plot] coordinates {
    (0.06186,35.6201)
    (0.09131,36.8795)
    (0.14823,38.1628)
    (0.25009,39.2655)
};

\addplot[color=rdB, delayedRDtwo, forget plot] coordinates {
    (0.05212,35.5669)
    (0.07509,36.7946)
    (0.12132,38.0494)
    (0.20672,39.1036)
};

\addplot[color=rdB, delayedRDthree, forget plot] coordinates {
    (0.04597,35.5080)
    (0.06616,36.7149)
    (0.10790,37.9564)
    (0.18708,39.0051)
};

\addplot[color=rdHM, delayedRDone, forget plot] coordinates {
    (0.01031,29.9935)
    (0.01801,31.7560)
    (0.03365,33.5764)
    (0.05937,35.2262)
    (0.10523,36.6494)
};

\addplot[color=rdHM, delayedRDthree, forget plot] coordinates {
    (0.01076,31.3171)
    (0.01931,32.9630)
    (0.03369,34.5674)
    (0.05938,36.1669)
    (0.12511,37.6169)
};

\addplot[color=rdGFSC, delayedRDone, line width=1.15pt, forget plot] coordinates {
    (0.02713,34.1748)
    (0.04150,35.4637)
    (0.06621,36.6592)
    (0.11237,37.7571)
};

\addplot[color=rdGFSC, delayedRDtwo, line width=1.15pt, forget plot] coordinates {
    (0.02807,34.3799)
    (0.04228,35.6792)
    (0.06689,36.8741)
    (0.11286,37.9083)
};

\addplot[color=rdGFSC, delayedRDthree, line width=1.15pt, forget plot] coordinates {
    (0.03108,34.6910)
    (0.04537,35.9891)
    (0.07377,37.2683)
    (0.12487,38.3358)
};

\addlegendimage{color=rdCausal, line width=0.95pt, mark=diamond*, mark size=1.8pt}
\addlegendentry{DCVC (causal)}
\addlegendimage{color=rdB, line width=1.15pt, no markers}
\addlegendentry{DCVC-B}
\addlegendimage{color=rdHM, line width=1.15pt, no markers}
\addlegendentry{HM-RA}
\addlegendimage{color=rdGFSC, line width=1.15pt, no markers}
\addlegendentry{DCVC-GFSC}

\nextgroupplot[
    xmin=0.03, 
    ymin=33.0, ymax=38.75,
    ytick={31,33,35,37,39},
    legend to name=delayedRDDelays,
    legend columns=3,
    legend style={delayedRDlegend},
]
\addplot[color=rdCausal, solid, mark=diamond*, mark size=1.8pt, forget plot] coordinates {
    (0.03530,33.6282)
    (0.05272,34.9562)
    (0.07906,36.2403)
    (0.12579,37.3329)
};

\addplot[color=rdB, delayedRDone, forget plot] coordinates {
    (0.07162,35.9374)
    (0.10527,37.1812)
    (0.17556,38.4734)
    (0.27881,39.5668)
};

\addplot[color=rdB, delayedRDtwo, forget plot] coordinates {
    (0.06380,35.8541)
    (0.09085,37.0630)
    (0.14892,38.3259)
    (0.23716,39.3769)
};

\addplot[color=rdB, delayedRDthree, forget plot] coordinates {
    (0.05827,35.7855)
    (0.08259,36.9776)
    (0.13574,38.2085)
    (0.21816,39.2444)
};

\addplot[color=rdHM, delayedRDone, forget plot] coordinates {
    (0.01108,29.7893)
    (0.01934,31.4772)
    (0.03656,33.2585)
    (0.06619,35.1712)
    (0.12082,37.1676)
};

\addplot[color=rdHM, delayedRDthree, forget plot] coordinates {
    (0.01255,30.6989)
    (0.02334,32.2926)
    (0.04310,34.1709)
    (0.08254,36.2116)
    (0.16167,38.4734)
};

\addplot[color=rdGFSC, delayedRDone, line width=1.15pt, forget plot] coordinates {
    (0.03995,34.3566)
    (0.05896,35.6799)
    (0.09068,36.9297)
    (0.14295,37.9653)
};

\addplot[color=rdGFSC, delayedRDtwo, line width=1.15pt, forget plot] coordinates {
    (0.04006,34.5374)
    (0.05850,35.8857)
    (0.08949,37.1432)
    (0.14199,38.1420)
};

\addplot[color=rdGFSC, delayedRDthree, line width=1.15pt, forget plot] coordinates {
    (0.04497,34.9252)
    (0.06356,36.2726)
    (0.09857,37.5493)
    (0.15096,38.5614)
};

\addlegendimage{color=black!65, delayedRDbase, delayedRDone}
\addlegendentry{$K=1$}
\addlegendimage{color=black!65, delayedRDbase, delayedRDtwo}
\addlegendentry{$K=2$}
\addlegendimage{color=black!65, delayedRDbase, delayedRDthree}
\addlegendentry{$K=3$}

\end{groupplot}

\node[anchor=south, inner sep=0pt, font=\small] at
    ($(delayedRD c1r1.north west)!0.5!(delayedRD c1r1.north east)+(0,0.14cm)$)
    {(a) UVG};
\node[anchor=south, inner sep=0pt, font=\small] at
    ($(delayedRD c2r1.north west)!0.5!(delayedRD c2r1.north east)+(0,0.14cm)$)
    {(b) MCL-JCV};

\coordinate (legendcenter) at
    ($(delayedRD c1r1.north west)!0.5!(delayedRD c2r1.north east)$);
\node[anchor=south, inner sep=0pt] at ($(legendcenter)+(0,1.02cm)$)
    {\pgfplotslegendfromname{delayedRDModels}};
\node[anchor=south, inner sep=0pt] at ($(legendcenter)+(0,0.60cm)$)
    {\pgfplotslegendfromname{delayedRDDelays}};
\end{tikzpicture}
\endgroup
\vspace{-.5em}
\caption{Rate-PSNR comparison under delay $K\leq 3$ on (a) UVG and (b) MCL-JCV. Colors identify model families. Line styles and markers distinguish delay settings.}
\label{fig:delayed_rd_curves}
\vspace{-.5em}
\end{figure}

\textbf{Delayed baselines.}
In Fig.~\ref{fig:delayed_rd_curves}, we compare GFSC with delayed coding baselines, including DCVC-B~\citep{sheng2025bi} and HM (\textit{random access}), under delay budgets $K\le 3$.
For DCVC-B, we evaluate $K=1,2,3$, corresponding to \verb|I-B-I|, \verb|I-B-B-I|, and \verb|I-B-B-B-I|, respectively.
For HM, due to its supported GOP structures, we evaluate $K=1,3$ using \verb|I-B-I| and \verb|I-B-B-B-I|.
We also include DCVC-L, a widened DCVC baseline with a parameter count matched to DCVC-GFSC, to isolate the gain from delayed modeling and that due to increased model capacity.
GFSC consistently achieves better rate-distortion performance than both the delayed and capacity-matched baselines. 
The corresponding BD-rate are reported in Table~\ref{tab:complexity}. 

\textbf{Complexity analysis.}
Table~\ref{tab:complexity} compares DCVC-GFSC with causal DCVC and DCVC-B.
GFSC introduces only modest additional latency over the causal baseline as the delay budget increases, while consistently improving rate-distortion performance.
Under matched delay constraints, GFSC remains substantially faster than DCVC-B and also achieves better BD-rate performance.
These results show that GFSC provides a more efficient use of limited delay, offering stronger compression gains with considerably lower latency than bidirectional coding.

\section{Related work}
\vspace{-1em}
\textbf{Neural video compression.}
Recent NVC methods have achieved rate-distortion performance competitive with conventional video codecs. 
Most existing approaches follow a strictly causal predictive coding paradigm, in which each frame is encoded using only previously reconstructed frames. 
Early methods adopt motion compensation followed by residual coding~\citep{lu2019dvc, agustsson2020scale, lin2020m}, whereas more recent approaches directly model temporal dependencies in the latent space through learned conditional coding~\citep{li2021deep, sheng2022temporal, li2023neural, li2024neural, mentzer2022vct}. 
Although these methods support zero-frame delay coding, they cannot exploit correlations with future frames. 
Bidirectional codecs address this limitation through interpolation-based prediction~\citep{wu2018video,djelouah2019neural,xu2024ibvc} or bidirectional conditional modeling~\citep{yang2020learning,yilmaz2021end,sheng2025bi}. 
However, they typically rely on fixed anchor-frame structures and are designed for relatively large frame delays, making them less flexible under small delay budgets. 
More recently, \citet{xiang2025real} incorporated $1$-frame delay into a unified intra/inter coding framework. 
Their method, however, is specialized to the $1$-delay setting and does not provide either a general construction for arbitrary delay budgets or an information-theoretic justification of the resulting rate-distortion gains. 

\begin{table}[t]
\centering
\caption{BD-rate (\%) and computational complexity under matched delay budgets. Latency, memory, FLOPs, and model parameters are reported for DCVC, DCVC-GFSC, and DCVC-B.}
\label{tab:complexity}
\resizebox{\linewidth}{!}{
\begin{tabular}{|l|c|c|c|c|c|c|c|c|}
\hline
\multirow{2}{*}{Method} & Delay & BD-rate & Enc. Latency & Dec. Latency & Memory & FLOPs & Params \\
 & $K$ & (\%) & (s/frame) & (s/frame) & (GB) & (G/frame) & (M) \\
\hline
DCVC 
& 0 
& 0.0 
& 0.52
& 0.44
& 5.54 
& 4430.53 
& 34.53 \\
\hline
DCVC-L
& 0 
& -8.57 
& 0.53
& 0.44
& 5.54
& 4447.51 
& 39.21 \\
\hline
\multirow{3}{*}{DCVC-B}
& 1
& 3.70
& 0.84
& 0.64
& 10.95
& 7153.37
& 55.38 \\

& 2
& -18.64
& 1.14
& 0.89
& 10.95
& 8805.79
& 55.38 \\

& 3
& -27.24
& 1.16
& 0.90
& 10.95
& 9632.01
& 55.38 \\
\hline
\multirow{3}{*}{DCVC-GFSC}
& 1 
& -20.57 
& 0.65
& 0.44
& 5.52
& 5916.28 
& 39.21 \\

& 2 
& -25.29
& 0.80
& 0.44
& 5.52
& 7451.81
& 39.21 \\

& 3
& -28.22
& 0.88
& 0.44
& 5.52
& 8987.58 
& 39.21 \\
\hline
\end{tabular}}
\vspace{-1em}
\end{table}

\textbf{Delayed sequential coding.}
The information-theoretic foundations of delayed sequential coding have been studied in \citet{ma2007delayed,ma2007value}, showing that limited delay can improve the rate-distortion performance of sequential coding and even a single-frame delay holds potential for matching the JC optimality in specific cases, e.g. first-order Gauss-Markov sources. 
However, these results do not establish whether such optimality extends to general $K$-order Markov sources, nor do they provide a constructive coding architecture for practical video compression.

\vspace{-0.75em}
\section{Conclusion}
\vspace{-0.75em}
We introduced GFSC, a general framework that treats frame delay as an explicit coding resource rather than fixing sequential video coding to the causal setting.
By exploiting the clique structure of $K$-order Markov sources, we established the first rigorous result showing that a $K$-frame delay can attain joint-coding rate-distortion optimality for a broad class of general Markov sources.
For Gauss-Markov sources, this construction further reduces to pairwise interactions with linear interpolation sufficient statistics, revealing a simple structure behind the optimal auxiliaries.
GFSC can be implemented with lightweight neural modules and integrated into existing NVC models with minimal architectural changes.
Experiments consistently demonstrate improved rate-distortion performance over causal coding and bidirectional codecs under matched delay budgets.
Together, these results identify limited delay as a fundamental and practically useful design dimension for neural video compression, bridging information-theoretic optimality and implementable coding architectures.
Limitations and future directions are discussed in Appendix~\ref{app-sec:limitations}.

\newpage
\bibliography{iclr2027_conference}
\bibliographystyle{iclr2027_conference}

\newpage
\appendix
\section{Comparison of coding structures}

\subsection{Delayed sequential coding}\label{app-subsec:delayed}
\input{figs/app-fig-dsc_setup}

For completeness, we briefly review the three canonical setups of delayed sequential coding introduced in~\cite{ma2007delayed}. 
These include: (i) noncausal encoder and causal decoder (NC-C), (ii) causal encoder and noncausal decoder (C-NC), and (iii) noncausal encoder and noncausal decoder (NC-NC). 
The NC-C setting is defined in Section~\ref{subsec:delayed}. 
Here we provide the definitions of the remaining two setups.

\textbf{C-NC (causal encoder, noncausal decoder).}
In the C-NC setting, the encoder operates causally, while the decoder is allowed a delay of $K$ frames. 
The encoding and decoding mappings are given by
\begin{gather*}
    f_{\ell}: \mathcal{X}_1 \times \cdots \times \mathcal{X}_{\ell} \times \mathcal{W} \to \mathcal{M}_{\ell}, \\
    g_{\ell}: \mathcal{M}_1 \times \cdots \times \mathcal{M}_{\min\{\ell+K, L\}} \times \mathcal{W} \to \hat{\mathcal{X}}_{\ell}.
\end{gather*}
That is, the $\ell$-th encoder has access only to past and current frames $(X_1, \cdots, X_{\ell})$, while the decoder reconstructs $\hat{X}_{\ell}$ using messages up to time $\ell + K$.

\textbf{NC-NC (noncausal encoder, noncausal decoder).}
In NC-NC setting, both encoders and decoders are allowed limited lookahead. 
Let $K_1$ and $K_2$ denote the delays at the encoder and decoder, respectively, such that $K_1 + K_2 = K$. 
The encoding and decoding mappings are defined as
\begin{gather*}
    f_{\ell}: \mathcal{X}_1 \times \cdots \times \mathcal{X}_{\min\{\ell + K_1, L\}} \times \mathcal{W} \to \mathcal{M}_{\ell}, \\
    g_{\ell}: \mathcal{M}_1 \times \cdots \times \mathcal{M}_{\min\{\ell + K_2, L\}} \times \mathcal{W} \to \hat{\mathcal{X}}_{\ell}.
\end{gather*}
Thus, the encoder can access future frames up to $K_1$ steps ahead, while the decoder reconstructs each frame with an additional delay of $K_2$.

These three setups characterize different ways of distributing the delay budget between the encoder and decoder, and together provide a unified framework for analyzing delayed sequential coding systems. 
We illustrate the corresponding system architectures for the three delayed setup in Figure~\ref{app-fig:delayed_setups}. 
From a theoretical perspective, Corollary~6.1 in~\cite{ma2007delayed} shows that all three setups achieve the same sum-rate performance for a given total delay $K$.

\subsection{NVC paradigm comparison}
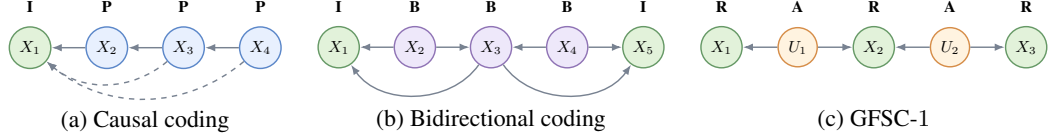
\begin{figure}[!ht]
    \centering

    \definecolor{iframefill}{RGB}{226,242,218}   
    \definecolor{iframedraw}{RGB}{102,153,102}

    \definecolor{pfill}{RGB}{232,240,254}        
    \definecolor{pdraw}{RGB}{91,135,201}

    \definecolor{bfill}{RGB}{238,232,250}        
    \definecolor{bdraw}{RGB}{143,115,181}

    \definecolor{vfill}{RGB}{255,243,224}        
    \definecolor{vdraw}{RGB}{214,141,72}

    \definecolor{edgecol}{RGB}{120,130,145}      

    \begin{subfigure}[t]{0.275\columnwidth}
        \centering
        \resizebox{\linewidth}{!}{
        \begin{tikzpicture}[
            baseline=(current bounding box.north),
            inode/.style={
                circle, draw=iframedraw, thick,
                minimum size=0.62cm, fill=iframefill,
                font=\small
            },
            pnode/.style={
                circle, draw=pdraw, thick,
                minimum size=0.62cm, fill=pfill,
                font=\small
            },
            edge_style/.style={
                thick, draw=edgecol, -{Latex[length=2mm]}
            },
            skip_style/.style={
                thick, draw=edgecol, dashed, -{Latex[length=2mm]}
            }
        ]
            \path[use as bounding box] (-0.5,-1.6) rectangle (5.0,1.2);

            \node[inode] (x1) at (0,0) {$X_1$};
            \node[pnode] (x2) at (1.5,0) {$X_2$};
            \node[pnode] (x3) at (3.0,0) {$X_3$};
            \node[pnode] (x4) at (4.5,0) {$X_4$};

            \node[font=\small] at (0,0.8) {\textbf{I}};
            \node[font=\small] at (1.5,0.8) {\textbf{P}};
            \node[font=\small] at (3.0,0.8) {\textbf{P}};
            \node[font=\small] at (4.5,0.8) {\textbf{P}};

            \draw[edge_style] (x4) -- (x3);
            \draw[edge_style] (x3) -- (x2);
            \draw[edge_style] (x2) -- (x1);

            \draw[skip_style] (x3) to[out=220,in=320] (x1);
            \draw[skip_style] (x4) to[out=220,in=320] (x1);
        \end{tikzpicture}
        }
        \vspace{-1.5em}
        \caption{Causal coding}
        \label{app-fig:p_code}
    \end{subfigure}
    \hfill
    \begin{subfigure}[t]{0.345\columnwidth}
        \centering
        \resizebox{\linewidth}{!}{
        \begin{tikzpicture}[
            baseline=(current bounding box.north),
            inode/.style={
                circle, draw=iframedraw, thick,
                minimum size=0.62cm, fill=iframefill,
                font=\small
            },
            bnode/.style={
                circle, draw=bdraw, thick,
                minimum size=0.62cm, fill=bfill,
                font=\small
            },
            edge_style/.style={
                thick, draw=edgecol, -{Latex[length=2mm]}
            }
        ]
            \path[use as bounding box] (-0.5,-1.6) rectangle (6.5,1.2);

            \node[inode] (x1) at (0,0) {$X_1$};
            \node[bnode] (x2) at (1.5,0) {$X_2$};
            \node[bnode] (x3) at (3.0,0) {$X_3$};
            \node[bnode] (x4) at (4.5,0) {$X_4$};
            \node[inode] (x5) at (6.0,0) {$X_5$};

            \node[font=\small] at (0,0.8) {\textbf{I}};
            \node[font=\small] at (1.5,0.8) {\textbf{B}};
            \node[font=\small] at (3.0,0.8) {\textbf{B}};
            \node[font=\small] at (4.5,0.8) {\textbf{B}};
            \node[font=\small] at (6.0,0.8) {\textbf{I}};

            \draw[edge_style] (x2) -- (x3);
            \draw[edge_style] (x2) -- (x1);
            \draw[edge_style] (x4) -- (x3);
            \draw[edge_style] (x4) -- (x5);

            \draw[edge_style] (x3) to[out=235,in=305] (x1);
            \draw[edge_style] (x3) to[out=305,in=235] (x5);
        \end{tikzpicture}
        }
        \vspace{-1.5em}
        \caption{Bidirectional coding}
        \label{app-fig:b_code}
    \end{subfigure}
    \hfill
    \begin{subfigure}[t]{0.345\columnwidth}
        \centering
        \resizebox{\linewidth}{!}{
        \begin{tikzpicture}[
            baseline=(current bounding box.north),
            rnode/.style={
                circle, draw=iframedraw, thick,
                minimum size=0.62cm, fill=iframefill,
                font=\small
            },
            vnode/.style={
                circle, draw=vdraw, thick,
                minimum size=0.62cm, fill=vfill,
                font=\small
            },
            edge_style/.style={
                thick, draw=edgecol, -{Latex[length=2mm]}
            }
        ]
            \path[use as bounding box] (-0.5,-1.6) rectangle (6.5,1.2);

            \node[rnode] (x1) at (0,0) {$X_1$};
            \node[vnode] (u12) at (1.5,0) {$U_1$};
            \node[rnode] (x2) at (3.0,0) {$X_2$};
            \node[vnode] (u13) at (4.5,0) {$U_2$};
            \node[rnode] (x3) at (6.0,0) {$X_3$};

            \node[font=\small] at (0,0.8) {\textbf{R}};
            \node[font=\small] at (1.5,0.8) {\textbf{A}};
            \node[font=\small] at (3.0,0.8) {\textbf{R}};
            \node[font=\small] at (4.5,0.8) {\textbf{A}};
            \node[font=\small] at (6.0,0.8) {\textbf{R}};

            \draw[edge_style] (u12) -- (x1);
            \draw[edge_style] (u12) -- (x2);
            \draw[edge_style] (u13) -- (x2);
            \draw[edge_style] (u13) -- (x3);
        \end{tikzpicture}
        }
        \vspace{-1.5em}
        \caption{GFSC-$1$}
        \label{app-fig:lisc}
    \end{subfigure}
    \caption{
    \textit{Left}: graphical illustration for causal predictive coding. 
    \textit{Middle}: graphical illustration for bidirectional coding. 
    \textit{Right}: graphical illustration for proposed GFSC-$1$. }
    \label{app-fig:p_b_v}
\end{figure}

We briefly review common practical video coding paradigms and compare them with the proposed GFSC architecture. 
Figure~\ref{app-fig:p_b_v} provides a systematic comparison of these coding paradigms.

\textbf{P-frame coding.}
P-frame coding follows a causal predictive coding paradigm, where each frame is encoded conditioned only on previously reconstructed frames. 
This design enables low-latency processing and efficient implementation, and forms the foundation of most existing neural video compression methods. 
However, the strictly causal structure prevents the encoder from exploiting future frames, which limits the ability to capture longer-range temporal dependencies.

\textbf{B-frame coding.}
B-frame coding extends the predictive paradigm by incorporating both past and future frames through bidirectional prediction. 
In practice, this is typically implemented via interpolation or bidirectional motion modeling between anchor frames. 
While such designs can improve compression efficiency, they usually rely on fixed temporal structures and are better suited to larger-delay settings. 
As a result, they are less flexible in scenarios with limited delay budgets.

\textbf{GFSC.}
The proposed GFSC framework generalizes these paradigms by explicitly modeling a limited delay budget. 
Instead of relying on fixed bidirectional structures, GFSC constructs clique-wise dependencies within a delay window, enabling a flexible tradeoff between delay and compression efficiency. 
This formulation bridges the gap between strictly causal coding and joint coding, while remaining compatible with modern NVC architectures.

\section{Markov source models under general distortion}

\subsection{Operational rate-distortion analysis} \label{app-subsec:general_ope_rd_region}
Recall the information RD region for GFSC in Definition~\ref{def:gfsc_k_ird}. 
We first derive the information sum-rate function.

\begin{theorem}[Information minimum sum-rate]\label{app-thm:general_min_sum_rate}
For a given distortion tuple $\mathbf{D}$, the minimum information sum-rate of GFSC-$K$ as in Definition~\ref{def:gfsc_k_ird} is
\begin{equation}\label{app-eq:general_sum_rate}
    R^{\mathrm{GFSC}}_{K,\mathrm{sum}}(\mathbf{D}) = \min I(\mathbf{X}; \mathbf U),
\end{equation}
and the minimum is taken over all auxiliary variables $\mathbf U$ satisfying Markov chains in \eqref{eq:korder_mc_def} and the distortion constraints in \eqref{eq:general_ird_constraints}. 
\end{theorem}
\begin{proof}
By Definition~\ref{def:gfsc_k_ird}, for a given distortion tuple $\mathbf{D}$, the minimum sum-rate is given by
\begin{equation*}
    R^{\mathrm{GFSC}}_{K,\mathrm{sum}}(\mathbf{D}) = \min \sum^L_{\ell = 1} I(X^{\min\{\ell + K, L\} }_{\ell}; U_{\ell} | U^{\ell - 1}),
\end{equation*}
where the minimization is over all feasible auxiliary variables satisfying the constraints in \eqref{eq:korder_mc_def} and the distortion constraints. 
For notational convenience, we denote each stage rate as $I(X^{\ell + K }_{\ell}; U_{\ell} | U^{\ell - 1})$. 
Then by the Markov chain in \eqref{eq:korder_mc_def}, we apparently have:
\begin{align*}
    I(\mathbf X; U_{\ell} | U^{\ell - 1}) &= I(X^{\ell + K}_{\ell}; U_{\ell} | U^{\ell - 1}) + I(\mathbf X \setminus X^{\ell + K}_{\ell}; U_{\ell} | X^{\ell + K}_{\ell}, U^{\ell - 1}) \\
    &= I(X^{\ell + K}_{\ell}; U_{\ell} | U^{\ell - 1}).
\end{align*}
Substituting this into the sum-rate expression, we have
\begin{align}
    R^{\mathrm{GFSC}}_{K,\mathrm{sum}}(\mathbf{D})
    &= \min \sum^L_{\ell = 1} I(X^{\min\{\ell + K, L\} }_{\ell} ; U_{\ell} | U^{\ell - 1}) \\
    &= \min \sum^L_{\ell = 1} I(\mathbf{X}; U_{\ell} | U^{\ell - 1}) \label{app-eq:general_sum_rate_1} \\
    &= \min I(\mathbf{X}; \mathbf U), \label{app-eq:general_sum_rate_2}
\end{align}
where \eqref{app-eq:general_sum_rate_1} follows from the Markov chain in \eqref{eq:korder_mc_def}, and \eqref{app-eq:general_sum_rate_2} follows from the chain rule. 
This completes the proof.
\end{proof}

Next, we show the proof for Theorem~\ref{thm:ope_rd_region} to characterize the operational RD region. 

\begin{theorem}[Rate-distortion characterization]\label{app-thm:general_ope_rd_region}
For a given distortion tuple $\mathbf{D}$, the following holds:
\begin{enumerate}[leftmargin=17.5pt]
    \item For any $\mathbf{R} \in \mathcal{R}_K(\mathbf{D})$, the operational rate region $\mathcal{R}^o_K(\mathbf{D})$ satisfies
    \begin{equation}\label{app-eq:general_ope_rd_1}
        \mathbf{R} + \log(\mathbf{R} + 1) + 5 \in \mathcal{R}^o_K(\mathbf{D}),
    \end{equation}
    and 
    \begin{equation}\label{app-eq:general_ope_rd_2}
        \mathcal{R}^o_K(\mathbf{D}) \subseteq \mathcal{R}_K(\mathbf{D}).
    \end{equation}
    \item The operational sum-rate satisfies
    \begin{equation}
        R^{\mathrm{GFSC}}_{K,\mathrm{sum}}(\mathbf{D}) \le R^{\mathrm{GFSC},o}_{K,\mathrm{sum}}(\mathbf{D}) \le 
        R^{\mathrm{GFSC}}_{K,\mathrm{sum}}(\mathbf{D}) + \mathcal{O}\!\left(\log R^{\mathrm{GFSC}}_{K,\mathrm{sum}}(\mathbf{D}) \right).
    \end{equation}
    where $R^{\mathrm{GFSC}}_{K,\mathrm{sum}}(\mathbf{D})$ is characterized in Theorem~\ref{app-thm:general_min_sum_rate}. 
\end{enumerate}
\end{theorem}

\begin{proof}
\textbf{Part 1. One-shot achievability} \\
By Strong Functional Representation Lemma (SFRL) in \cite{li2018strong}, for jointly distributed random variables $X,Y \sim P_{XY}$, there exists $Z$ independent of $X$, and function $f$ such that $Y = f(X, Z)$. $Z$ is not necessarily unique. 
SFRL further states that there exists a $Z$ such that:
\begin{equation*}
    H(Y|Z) \le I(X; Y) + \log(I(X; Y) + 1) + 4.
\end{equation*}
Notice that this can be extended to conditional distribution case. Given $P_{XY|W}$, there exists $Z$ independent of $(X, W)$ such that $Y = f(X, W, Z)$ and:
\begin{equation*}
    H(Y|W, Z) \le I(X; Y | W) + \log(I(X; Y | W) + 1) + 4.
\end{equation*}
\vspace{0.5em}
\underline{\textit{Proof of \eqref{app-eq:general_ope_rd_1}}:} $\mathbf{R} + \log(\mathbf{R} + 1) + 5 \in \mathcal{R}^o_K(\mathbf{D})$. \\
For a given $\mathbf{D}$ and $\mathbf{R} \in \mathcal{R}_{K}(\mathbf{D})$, let $\mathbf{U}$ be jointly distributed with $\mathbf{X}$, where the Markov chains in \eqref{eq:korder_mc_def}, rate and distortion constraints in \eqref{eq:general_ird_constraints} are all satisfied. 
Denote the joint distribution of such $(\mathbf{X}, \mathbf{U}, \hat{\mathbf{X}})$ by $P_{\mathbf{X} \mathbf{U} \hat{\mathbf{X}}}$ and according to the Markov chains in \eqref{eq:korder_mc_def}, we have:
\begin{equation}\label{app-eq:general_joint_dist}
    P_{\mathbf{X} \mathbf{U} \hat{\mathbf{X}}} = P_{\mathbf X}\cdot \prod^L_{\ell = 1} P_{U_{\ell} | X^{ \min\{ \ell + K, L \} }_{\ell}, U^{\ell - 1}} \cdot \prod^L_{\ell = 1} \mathbf{1}\{ \hat{X}_{\ell} = g_{\ell}(U^{\ell}) \}.
\end{equation}
Now, fix the marginal distribution $P_{\mathbf{U}}$ in \eqref{app-eq:general_joint_dist} and use it to construct the codebook, i.e., 
\begin{equation}\label{app-eq:general_codebook_distribution}
    P_{\mathbf{U}} = P_{U_L |U^{L-1}} \cdots P_{U_{\ell} | U^{\ell - 1}} \cdots P_{U_1}.
\end{equation} 
By the SFRL, we have:
\begin{enumerate}
    \item (\textit{First stage}) There exists a random variable $Z_1$ independent of $X^{K+1}$ and a deterministic function $q_{1}$ such that $U_1 = q_1(X^{K+1}, Z_1)$ and:
    \begin{equation*}
        H(U_1| Z_1) \le I(X^{K+1}; U_1) + \log(I(X^{K+1}; U_1) + 1) + 4.
    \end{equation*}
    Therefore, the first stage encoder applies the deterministic function $q_1$ to get $U_1$ whose distribution needs to be preserved according to \eqref{app-eq:general_codebook_distribution}.
    \item (\textit{Intermediate stages}) By the conditional SFRL, there exists a random variable $Z_{\ell}$ independent of $(U^{\ell - 1}, X^{ \min\{\ell + K, L\} }_{\ell})$ and a deterministic function $q_{\ell}$ such that $U_{\ell} = q_{\ell}(U^{\ell - 1}, X^{ \min\{\ell + K, L\} }_{\ell}, Z_{\ell})$ and:
    \begin{equation*}
        H(U_{\ell} | U^{\ell - 1}, Z_{\ell}) \le I( X^{ \min\{\ell + K, L\} }_{\ell} ; U_{\ell} | U^{\ell - 1} ) + \log\left( I( X^{ \min\{\ell + K, L\} }_{\ell} ; U_{\ell} | U^{\ell - 1} ) + 1 \right) + 4
    \end{equation*}
    At the intermediate $\ell$-th stage (including the last stage), the $\ell$-th encoder has access to $U^{\ell - 1}$ from earlier stages. 
    So upon observing the delayed source window $X^{ \min\{\ell + K, L\} }_{\ell}$, it applies $q_{\ell}$ to get $U_{\ell}$ whose conditional distribution needs to be preserved according to \eqref{app-eq:general_codebook_distribution}.
\end{enumerate}
Now the encoding and decoding schemes are as follows:
\begin{enumerate}
    \item (\textit{First stage encoder}) With $Z_1$ available at encoder and decoder, we can design prefix-free codes indexed by $Z_1$ to losslessly represent $U_1$ with the expected codeword length $\bar{R}_1$
    \begin{equation*}
        \bar{R}_1 \le H(U_1 | Z_1) + 1 \le I( X^{K+1}; U_1) + \log( I(X^{K+1}; U_1) + 1) + 5
    \end{equation*}
    \item (\textit{Intermediate stage encoders}) With $Z_{\ell}$ available at encoder and decoder, we can design prefix-free codes indexed by $(Z_{\ell}, U^{\ell - 1})$ to losslessly represent $U_{\ell}$ with the expected codeword length $\bar{R}_{\ell}$ 
    \begin{equation*}
    \begin{aligned}
        \bar{R}_{\ell} & \le H(U_{\ell} | U^{\ell - 1}, Z_{\ell}) + 1 \\
        & \le I( X^{ \min\{\ell + K, L\} }_{\ell} ; U_{\ell} | U^{\ell - 1}) + \log(I( X^{ \min\{\ell + K, L\} }_{\ell} ; U_{\ell} | U^{\ell - 1}) + 1) + 5
    \end{aligned}
    \end{equation*}
    \item (\textit{Decoders}) Upon receiving all $U^{\ell}$, the $\ell$-th decoder generate the reconstruction via $\hat{X}_{\ell} = g_{\ell} (U^{\ell})$.
\end{enumerate}
Then we have $\mathbf{R} + \log(\mathbf{R} + 1) + 5 \in \mathcal{R}^o_{K}(\mathbf{D})$. 

\vspace{0.5em}
\underline{\textit{Proof of \eqref{app-eq:general_ope_rd_2}}:}  $\mathcal{R}^o_{K}(\mathbf{D}) \subseteq \mathcal{R}_{K}(\mathbf{D})$. \\
For any $\mathbf{D}$ and $\mathbf{R} \in \mathcal{R}^o_K(\mathbf{D})$, shared randomness $W\in\mathcal{W}$, encoding and decoding functions
\begin{equation*}
\begin{aligned}
    f_{\ell}: \mathcal{X}_{\ell}^{\min\{\ell+K, L\} } \times \mathcal{M}^{\ell-1} \times \mathcal{W} \to \mathcal{M}_{\ell}, \qquad 
    g_{\ell}: \mathcal{M}^{\ell} \times \mathcal{W} \to \hat{\mathcal{X}}_{\ell},
\end{aligned}
\end{equation*}
such that:
\begin{align}
    & R_{\ell} \ge \mathbb{E}[l(M_{\ell})], \quad \ell = 1, \cdots, L, \label{app-eq:general_oneshot_constraint_1} \\
    & D_{\ell} \ge \mathbb{E}[d_{\ell}(X_{\ell}, \hat{X}_{\ell})], \quad \ell = 1, \cdots, L, \label{app-eq:general_oneshot_constraint_2}
\end{align}
we need to lower bound the expected length of messages $l(M_{\ell})$. Define:
\begin{equation}
\begin{aligned}
    U_1 \triangleq (M_1, W), \qquad U_{\ell} \triangleq (M^{\ell}, W), \; \ell = 1, \cdots, L. \label{app-eq:general_oneshot_def_ujk}
\end{aligned}
\end{equation}
Then for the first stage:
\begin{equation}\label{app-eq:general_oneshot_constraint_3}
    R_{1} \ge \mathbb{E}[l(M_1)] \ge H(M_1 | W) = I(X^{K+1}; U_1). 
\end{equation}
For the intermediate stages:
\begin{equation}\label{app-eq:general_oneshot_constraint_4}
\begin{aligned}
    R_{\ell} \ge \mathbb{E}[l(M_{\ell})] &\ge H(M_{\ell} | M^{\ell - 1}, W) \\
    &= I( X^{ \min\{\ell + K, L\} }_{\ell} ; M_{\ell} | M^{\ell - 1}, W) \\
    &= I( X^{ \min\{\ell + K, L\} }_{\ell} ; U_{\ell} | U^{\ell - 1}).
\end{aligned}
\end{equation}
Notice that \eqref{app-eq:general_oneshot_def_ujk} implies the following Markov chains:
\begin{equation}\label{app-eq:general_oneshot_mc}
    U_{\ell} \to ( X^{ \min\{\ell + K, L\} }_{\ell}, U^{\ell - 1}) \to \mathbf{X}, \quad \ell = 1, \cdots, L. 
\end{equation}
On the other hand, by the definition of decoding functions and \eqref{app-eq:general_oneshot_def_ujk}, we have that:
\begin{equation}\label{app-eq:general_oneshot_decode_func}
    \hat{X}_{\ell} = g_{\ell} (M^{\ell}, W) \triangleq \eta_{\ell}(U^{\ell})
\end{equation}
where $\eta_{\ell}$ are deterministic functions. 

Now consider the fact that the set of constraints in \cref{app-eq:general_oneshot_constraint_1,app-eq:general_oneshot_constraint_2,app-eq:general_oneshot_constraint_3,app-eq:general_oneshot_constraint_4}, 
Markov chains in \cref{app-eq:general_oneshot_mc} and deterministic functions in \cref{app-eq:general_oneshot_decode_func} constitute an information rate-distortion region, denote by $\tilde{\mathcal{R}}_K$, which is the set of $(\mathbf{R}, \mathbf{D})$ such that:
\begin{gather*}
    R_{\ell} \ge I( X^{ \min\{\ell + K, L\} }_{\ell} ; U_{\ell} | U^{\ell - 1}), \quad \ell = 1, \cdots, L, \\
    D_{\ell} \ge \mathbb{E}[ d_{\ell}(X_{\ell}, \hat{X}_{\ell})], \quad \ell = 1, \cdots, L,
\end{gather*}
for all auxiliary random variables $\mathbf U$ and $\hat{\mathbf{X}}$ satisfying:
\begin{equation*}
\begin{gathered}
    \hat{X}_{\ell} = \eta_{\ell}(U^{\ell}), \qquad U_{\ell} \to ( X^{ \min\{\ell + K, L\} }_{\ell}, U^{\ell - 1}) \to \mathbf{X},
\end{gathered}
\end{equation*}
where $\eta_{\ell}$ are some deterministic functions. 
By the definition of $\mathcal R_K$ in Definition~\ref{def:gfsc_k_ird}, apparently we have $\tilde{\mathcal{R}}_K =  \mathcal{R}_K$. Therefore, $\mathcal R^o_K(\mathbf D) \subseteq \tilde{\mathcal R}_K(\mathbf D) = \mathcal R_K(\mathbf D) $. This completes the proof of Part~1.

\textbf{Part 2. Sum-rate bounds} \\
As we have already shown in Theorem~\ref{app-thm:general_min_sum_rate} that the information minimum sum-rate function of GFSC-$K$ system is:
\begin{equation*}
    R^{\mathrm{GFSC}}_{K,\mathrm{sum}}(\mathbf{D}) = \min I(\mathbf{X}; \mathbf U),
\end{equation*}
subject to constraints in \cref{eq:korder_mc_def,eq:general_ird_constraints}. 
We also have $\mathcal{R}^o_K(\mathbf{D}) \subseteq \mathcal{R}_{K}(\mathbf{D})$ in the proof of \eqref{app-eq:general_ope_rd_2}. 
Thus, we have the following lower bound:
\begin{align}
    R^{\mathrm{GFSC},o}_{K,\mathrm{sum}}(\mathbf{D}) = \min_{\mathbf{R}\in\mathcal{R}^o_K(\mathbf{D})} \sum^L_{\ell = 1} R_{\ell}\ge \min_{\mathbf{R}\in\mathcal{R}_K(\mathbf{D})} \sum_{\ell = 1} R_{\ell} = R^{\mathrm{GFSC}}_{K,\mathrm{sum}}(\mathbf{D}).
\end{align}
For the sum-rate upper bound, based on the proof of \eqref{app-eq:general_ope_rd_1}, we have:
\begin{align}
    R^{\mathrm{GFSC},o}_{K,\mathrm{sum}}(\mathbf{D}) &=
    \min_{\mathbf{R}\in\mathcal{R}^o_K(\mathbf{D})} \sum^L_{\ell = 1} R_{\ell} \\
    &\le \min_{\mathbf{R}\in\mathcal{R}_K(\mathbf{D})} \sum^L_{\ell = 1}
    \left( R_{\ell} + \log(R_{\ell}+1) + 5 \right) \label{app-eq:general_sr_ub_1} \\
    &=   \min_{\mathbf{R}\in\mathcal{R}_K(\mathbf{D})} \sum^L_{\ell = 1} R_{\ell} + \sum^L_{\ell = 1} \log(R_{\ell} + 1) + 5L \\
    &\le \min_{\mathbf{R}\in\mathcal{R}_K(\mathbf{D})} I(\mathbf{X} ; \mathbf U) + L \cdot \log\!\left( \frac{\sum^L_{\ell = 1} (R_{\ell}+1) }{L} \right) + 5L \label{app-eq:general_sr_ub_2} \\
    &\le R^{\mathrm{GFSC}}_{K,\mathrm{sum}}(\mathbf{D}) + L \cdot \log\!\left( \frac{ R^{\mathrm{GFSC}}_{K,\mathrm{sum}}(\mathbf{D}) + L }{ L } \right) + 5L \\
    &= R^{\mathrm{GFSC}}_{K,\mathrm{sum}}(\mathbf{D}) +
    \mathcal{O} \!\left( \log R^{\mathrm{GFSC}}_{K,\mathrm{sum}}(\mathbf{D}) \right),
\end{align}
where \eqref{app-eq:general_sr_ub_1} follows \eqref{app-eq:general_ope_rd_1}, \eqref{app-eq:general_sr_ub_2} follows the Jensen's inequality for logarithms. 
This completes the proof of Part~2. 
\end{proof}

\subsection{Joint-coding optimality for general Markov sources}
\label{app-subsec:general_jc_opt}

We provide several definitions for a general distortion function. 
\begin{definition}[Distortion function]\label{app-def:distortion}
Define the general distortion function form as:
\begin{equation*}
d: \mathcal X \times \mathcal X \to [0, \infty),    
\end{equation*}
with $d(x, x) = 0$ and $d(x, w) > 0$ for $w, x \in \mathcal X$, $w \ne x$. 
\end{definition}
\begin{definition}\label{app-def:min_distortion}
    Define $d_{\min} \triangleq \min_{x \ne w} d(x, w) > 0$ for all $x, w \in \mathcal X$.
\end{definition}
\begin{definition}\label{app-def:max_distortion}
    Define $d_{\max} \triangleq \max_{x \ne w} d(x, w) < \infty$ for all $x, w \in \mathcal X$. 
\end{definition}

For a general $K$-order Markov source as in Definition~\ref{def:markov_pos_dist}, its probability mass function $p_{\mathbf X}$ can be naturally factorized as $p_{\mathbf X}(\mathbf x) = \prod^{L-K}_{j=1} \psi_j(x_{\mathcal C_j})$ according to \cref{eq:korder_clique_factorization}, where $\mathcal C_j \triangleq \{j, \cdots, j + K\}$. 
With the positive distribution assumption, $\psi_j(x_{\mathcal C_j})$ is strictly positive factor. 
We define $\psi_j$ as the corresponding vectorized factor. 
\begin{definition}
    For $p_{\mathbf X}$, define the minimum probability mass $p_{\min} \triangleq \min_{\mathbf x\in \mathcal X_1 \times \cdots \times \mathcal X_L} p_{\mathbf X}(\mathbf x) > 0$. 
\end{definition}
\begin{definition}
    Define the clique minimum mass $ c_{j} \triangleq \min_{x_{\mathcal C_{j}}} \psi_{j} (x_{\mathcal C_{j}}) > 0$ for $1 \le j \le L-K$. 
\end{definition}
\begin{definition}
    Define the index set $\mathcal I_{\ell} \triangleq \{ \max\{1, \ell - K\}, \cdots, \min\{\ell, L-K\} \}$ for $1 \le \ell \le L$. 
\end{definition}
Now we provide the proof for Theorem~\ref{thm:general_main_thm}.

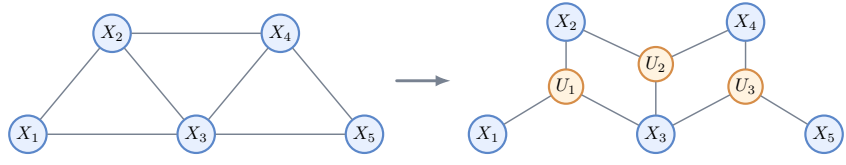
\begin{figure}[!t]
\centering
\resizebox{.8\linewidth}{!}{
\begin{tikzpicture}[
    font=\small,
    >=Latex,
    line cap=round,
    line join=round,
    xnode/.style={
        circle,
        draw=xdraw,
        very thick,
        minimum size=0.72cm,
        inner sep=1pt,
        fill=xfill,
        font=\normalsize
    },
    unode/.style={
        circle,
        draw=udraw,
        very thick,
        minimum size=0.66cm,
        inner sep=1pt,
        fill=ufill,
        font=\normalsize
    },
    edge_style/.style={
        thick,
        draw=edgecol
    },
    transform_arrow/.style={
        -{Latex[length=3.0mm,width=2.2mm]},
        line width=1.6pt,
        draw=edgecol
    }
]

\begin{scope}[x=1.08cm,y=1.08cm]
    \node[xnode] (lx1) at (0.0,0.0) {$X_1$};
    \node[xnode] (lx2) at (1.5,1.8) {$X_2$};
    \node[xnode] (lx3) at (3.0,0.0) {$X_3$};
    \node[xnode] (lx4) at (4.5,1.8) {$X_4$};
    \node[xnode] (lx5) at (6.0,0.0) {$X_5$};

    \draw[edge_style] (lx1) -- (lx2);
    \draw[edge_style] (lx2) -- (lx3);
    \draw[edge_style] (lx3) -- (lx4);
    \draw[edge_style] (lx4) -- (lx5);

    \draw[edge_style] (lx1) -- (lx3);
    \draw[edge_style] (lx2) -- (lx4);
    \draw[edge_style] (lx3) -- (lx5);
\end{scope}

\draw[transform_arrow] (7.10,1.03) -- (8.15,1.03);

\begin{scope}[xshift=8.85cm,x=1.08cm,y=1.08cm]
    \node[xnode] (rx1) at (0.0,0.0) {$X_1$};
    \node[xnode] (rx2) at (1.4,2.0) {$X_2$};
    \node[xnode] (rx3) at (3.0,0.0) {$X_3$};
    \node[xnode] (rx4) at (4.6,2.0) {$X_4$};
    \node[xnode] (rx5) at (6.0,0.0) {$X_5$};

    \node[unode] (u1) at (1.4,0.85) {$U_1$};
    \node[unode] (u2) at (3.0,1.25) {$U_2$};
    \node[unode] (u3) at (4.6,0.85) {$U_3$};

    \draw[edge_style] (u1) -- (rx1);
    \draw[edge_style] (u1) -- (rx2);
    \draw[edge_style] (u1) -- (rx3);

    \draw[edge_style] (u2) -- (rx2);
    \draw[edge_style] (u2) -- (rx3);
    \draw[edge_style] (u2) -- (rx4);

    \draw[edge_style] (u3) -- (rx3);
    \draw[edge_style] (u3) -- (rx4);
    \draw[edge_style] (u3) -- (rx5);
\end{scope}

\end{tikzpicture}
}
\caption{
Clique-wise augmentation of a second-order Markov network, where $U_j$ is associated with the maximal clique
$(X_j,X_{j+1},X_{j+2})$.
}
\label{fig:second_order_clique_graph}
\end{figure}

\begin{lemma}[Nontrivial clique-wise auxiliary variables]
\label{lem:clique_auxiliary}
Let $\mathbf X$ be a $K$-order Markov source as in Definition~\ref{def:markov_pos_dist}. 
Define a positive constant $\delta_0$ as follow:
\begin{equation}\label{app-eq:aux_delta_bound}
    \delta_0 \triangleq \min_{1 \le j \le L-K} \left\{ 1 - \left( 1 + \frac{ c_{j} }{ \|\psi_{j}\|_2 } \right)^{-\frac{1}{K+1}} \right\}.
\end{equation}
Then for any $0 < \delta < \delta_0$, there exist clique-wise auxiliary random variables
$(U^{(\delta)})^{L-K}$, jointly distributed with $\mathbf X$, such that:
\begin{enumerate}
    \item Joint distribution of $(\mathbf X, (U^{(\delta)})^{L-K})$ is also positive. 
    \item The following Markov chains hold:
    \begin{gather}
        U^{(\delta)}_{j} \to X^{j + K}_{j} \to (\mathbf X, \; U^{(\delta)}_{-j}), \quad j = 1,\ldots,L-K, \label{app-eq:general_u_markov_1}\\
        X_{\ell} \to U^{(\delta)}_{ \mathcal I_{\ell} } \to X_{-\ell}, \quad \ell = 1,\ldots,L. \label{app-eq:general_u_markov_2}
    \end{gather}
    \item Define $\Delta_{\ell}(\delta) \triangleq \min_{g_{\ell}} \mathbb E[d_{\ell} (X_{\ell}, g_{\ell}( U^{(\delta)}_{\mathcal I_{\ell}} ) )]$ over arbitrary deterministic mapping $g_{\ell}$. 
    Then $\Delta_{\ell}(\delta)$ is nontrivial, i.e. $0< \Delta_{\ell}(\delta) \le d_{\ell,\max}$.
\end{enumerate}
\end{lemma}
\begin{proof} 
We provide the proof in two parts.
We first show that $\Delta_{\ell}(\delta) < d_{\ell,\max}$ is generally true based on the definition of distortion in Definition~\ref{app-def:distortion}. 
Then we provide a way to explicitly construct auxiliary variables $(U^{(\delta)})^{L-K}$, which satisfy the Markov conditions in \cref{app-eq:general_u_markov_1,app-eq:general_u_markov_2} and induce strictly positive distortion for each source variable, i.e. $\Delta_{\ell}(\delta) > 0$, for all $0 < \delta < \delta_0$. 

\textbf{Part I. $\Delta_{\ell}(\delta) \le d_{\ell,\max}$.} \\
For arbitrary decoder input $U_{\mathcal I_{\ell}}$, we define $V_{\ell} \triangleq U_{\mathcal I_{\ell}}$ with $\mathcal V_{\ell} \triangleq \mathcal U_{\max\{1, \ell-K\}} \times \cdots \times \mathcal U_{{\min\{\ell, L-K\}}}$. 
Let $\hat x_{\ell} \triangleq g_{\ell}(v_{\ell})$ denote the decoder reconstruction given observation $v_{\ell} \in \mathcal V_{\ell}$. 
By the definition of distortion $d_{\ell}: \mathcal X_{\ell} \times \mathcal X_{\ell} \to [0, \infty) $, we also have $ \hat x_{\ell} \in \mathcal X_{\ell}$. 
Let $p_{X_{\ell}V_{\ell}}$ be the joint distribution of $X_{\ell}$ and $V_{\ell}$, then we can bound the induced distortion: 
\begin{align}
    \Delta_{\ell} &= \min_{g_{\ell}} \E[d_{\ell}(X_{\ell}, g_{\ell}(V_{\ell}))] \\
    &\le \E[d_{\ell}(X_{\ell}, g_{\ell}(V_{\ell}))] \label{app-eq:distortion_ineq_1} \\
    &  = \sum_{x_{\ell}, v_{\ell}} p_{X_{\ell}V_{\ell}}(x_{\ell}, v_{\ell}) d_{\ell}(x_{\ell}, \hat x_{\ell}) \\
    &\le \sum_{x_{\ell}, v_{\ell}} p_{X_{\ell}V_{\ell}}(x_{\ell}, v_{\ell}) d_{\ell, \max} \label{app-eq:distortion_ineq_2} \\
    &= d_{\ell, \max} < \infty, 
\end{align}
where \eqref{app-eq:distortion_ineq_1} is from the definition of $\Delta_{\ell}$, \eqref{app-eq:distortion_ineq_2} is from Definition~\ref{app-def:max_distortion}. 
So $\Delta_{\ell} \le d_{\ell,\max}$ is generally true for arbitrary auxiliary variables $V_{\ell}$. 

\textbf{Part II. $\Delta_{\ell}(\delta) > 0$.} \\
For $\Delta_{\ell}(\delta) > 0$, we proceed in two steps. 
First, for the general distortion function in Definition~\ref{app-def:distortion}, we derive the necessary and sufficient condition to attain positive distortion.
We then construct a set of auxiliary random variables $\{ U_{j} \}^{L-K}_{j = 1}$ and show that they satisfy both the required Markov conditions in \cref{app-eq:general_u_markov_1,app-eq:general_u_markov_2} and the derived condition for positive distortion. 

\underline{\textit{Positive distortion analysis}}. For a given observation $v_{\ell} \in \mathcal V_{\ell}$ at decoder side, the goal of $j$-stage decoding is to minimize expected distortion:
\begin{equation}
    \E[d_{\ell}(X_{\ell}, g_{\ell}(V_{\ell}))] = \sum_{v_{\ell}} p_{V_{\ell}}(v_{\ell}) \sum_{x_{\ell}} p_{X_{\ell} | V_{\ell}}(x_{\ell} | v_{\ell}) d_{\ell}(x_{\ell}, \hat x_{\ell}).
\end{equation} 
Here, we define the conditional risk $R_{v_{\ell}}(\hat x_{\ell}) \triangleq \sum_{x_{\ell}} p_{X_{\ell} | V_{\ell}}(x_{\ell} | v_{\ell}) d_{\ell}(x_{\ell}, \hat x_{\ell})$. 
Apparently we have $R_{v_{\ell}}(\hat x_{\ell}) \ge \min_{\hat x_{\ell} \in \mathcal X_{\ell}} R_{v_{\ell}}(\hat x_{\ell})$, thereby
\begin{equation}
    \E[d_{\ell}(X_{\ell}, g_{\ell}(V_{\ell}))] \ge \sum_{v_{\ell}} p_{V_{\ell}}(v_{\ell}) \min_{\hat x_{\ell} \in \mathcal X_{\ell}} R_{v_{\ell}}(\hat x_{\ell}).
\end{equation}  
Then the optimal decoding function $g^*_{\ell}$ satisfies:
\begin{equation}
    g^*_{\ell}(v_{\ell}) \in \arg\min_{\hat{x}_{\ell} \in \mathcal X_{\ell}} \sum_{x_{\ell} \in \mathcal X_{\ell}} p_{X_{\ell} | V_{\ell}}(x_{\ell} | v_{\ell}) d_{\ell}(x_{\ell}, \hat{x}_{\ell}).
\end{equation}
In addition, as in Definition~\ref{app-def:distortion}, we have $d_{\ell}(x_{\ell}, x_{\ell}) =0 $ and $d_{\ell}(x_{\ell}, \hat x_{\ell}) > 0$ for any $x_{\ell} \ne \hat x_{\ell}$.
Then for the minimum conditional risk $\min_{\hat x_{\ell}} R_{v_{\ell}}(\hat x_{\ell})$, we have:
\begin{align}
    \min_{\hat x_{\ell}} R_{v_{\ell}}(\hat x_{\ell}) &= \min_{\hat x_{\ell}}\sum_{x_{\ell}} p_{X_{\ell} | V_{\ell}}(x_{\ell} | v_{\ell}) d_{\ell}(x_{\ell}, \hat x_{\ell}) \\
    &= \min_{\hat x_{\ell}} \sum_{x_{\ell} \ne \hat x_{\ell}} p_{X_{\ell} | V_{\ell}}(x_{\ell} | v_{\ell}) d_{\ell}(x_{\ell}, \hat x_{\ell}) \\
    &\ge \min_{\hat x_{\ell}} \sum_{x_{\ell} \ne \hat x_{\ell}} p_{X_{\ell} | V_{\ell}}(x_{\ell} | v_{\ell}) d_{j,\min} \\
    &= d_{j,\min} \left( 1 - \max_{\hat x_{\ell}} p_{X_{\ell} | V_{\ell}}(\hat x_{\ell} | v_{\ell}) \right).
\end{align}
The minimum conditional risk is zero if and only if $p_{X_{\ell}|V_{\ell}}(\cdot | v_{\ell})$ is Dirac delta function, i.e.,
\begin{equation}
    p_{X_{\ell}|V_{\ell}}( x_{\ell} | v_{\ell} ) = \delta(x_{\ell} - \hat{x}_{\ell}) = \begin{cases}
        1, & x_{\ell} = \hat x_{\ell}, \\
        0, & x_{\ell} \ne \hat x_{\ell}.
    \end{cases}
\end{equation}
Therefore, the induced distortion $\Delta_{\ell}$ is zero if and only if $p_{X_{\ell}|V_{\ell}}( \cdot | v_{\ell} )$ is delta function for all $v_{\ell}$ with $p_{V_{\ell}}(v_{\ell}) > 0$. 
This is also equivalent to $H(X_{\ell} | V_{\ell}) = 0$. 
It is obvious that both distortion $\epsilon_{\ell}$ and $p_{X_{\ell} | V_{\ell}}$ entries are non-negative. 
Therefore, it suffice to induce positive distortion $\Delta_{\ell}(\delta) > 0$ if we can construct auxiliary variables $V_{\ell}$ with positive distribution and positive backward test channel $p_{X_{\ell} | V_{\ell}}$ for all $\ell$. 
In the following part, we provide an explicit construction. 

\underline{\textit{Construction of auxiliary variables}}. In this part, we show that for $K$-order Markov source $\mathbf X$ as defined in Definition~\ref{def:markov_pos_dist}, we can always construct appropriate auxiliary variables, which satisfy all the conditions. 
For $\mathbf X \sim p_{\mathbf X}$, $p_{\mathbf X}$ naturally adopts the following clique factorization:
\begin{equation}
    p_{\mathbf X}(\mathbf x) = \prod_{j = 1}^{L-K} \psi_{j}(x_{\mathcal C_{j}})
\end{equation}
where $\mathcal C_{j} \triangleq \{j, \cdots, j + K\}$, and $\psi_{j}(x_{\mathcal C_{j}}) > 0$ is a positive clique factor since $\mathbf X$ has positive joint distribution. 
In addition, $\psi_{j}(x_{\mathcal C_{j}})$ can be further represented by a vectorized clique factor $\psi_{j} \in \R^{N_{j}}_+$ with $N_{j} \triangleq \prod^{j+K}_{r = j} |\mathcal X_r|$. 

For $j$-th clique, we define the following auxiliary variable $U^{(\delta)}_j$ with $(K + 1)$ components:
\begin{equation}\label{eq:u_aux_def}
    U^{(\delta)}_{j} \triangleq (U^{(\delta)}_{j,r} : r \in \mathcal C_{j}) \quad \text{with} \quad U^{(\delta)}_{j,r} \in \mathcal X_r,
\end{equation}
which is also $U^{(\delta)}_{j} \triangleq (U^{(\delta)}_{j,j}, \cdots, U^{(\delta)}_{j,j+K})$. 
Now suppose a positive constant $0 < \delta < 1$, for each $U^{(\delta)}_{j,r}$, we consider a transition matrix $Q^{(\delta)}_{j, r} \in \R^{|\mathcal X_r| \times |\mathcal X_r|} $ from $U^{(\delta)}_{j,r}$ to $X_r$ with each entry as:
\begin{equation}
\begin{aligned}
    Q^{(\delta)}_{j, r}(x_r, u^{(\delta)}_{j,r}) = q_{X_r | U^{(\delta)}_{j,r}}(x_r | u^{(\delta)}_{j,r}) = (1 - \delta) \mathbf 1\{x_r = u^{(\delta)}_{j,r}\} + \delta \cdot \nu_r(x_r),
\end{aligned}
\end{equation}
where $\nu_r$ denotes arbitrary positive distribution over $\mathcal X_r$. 
Here, we choose $\nu_r(x_r)$ to be the uniform distribution, therefore $\nu_r(x_r) = \frac{1}{|\mathcal X_r|}$ for all $x_r \in \mathcal X_r$. 
Then $Q^{(\delta)}_{j,r}$ can be further written as:
\begin{equation}
    Q^{(\delta)}_{j,r} = (1 - \delta) \cdot I + \frac{\delta}{|\mathcal X_r|} \cdot \mathbf 1 \mathbf 1^{\top}
    = \underbrace{ \begin{bmatrix}
        1 - \delta + \frac{\delta}{|\mathcal X_r|} & \cdots & \frac{\delta}{|\mathcal X_r|} \\
        \vdots & \ddots & \vdots \\
        \frac{\delta}{|\mathcal X_r|} & \cdots & 1 - \delta + \frac{\delta}{|\mathcal X_r|}
    \end{bmatrix}
    }_{ u^{(\delta)}_{j,r} \in \mathcal X_r}
    \left.\vphantom{\begin{bmatrix}
        1 \\ \vdots \\ 1
    \end{bmatrix}}
    \right\}
    x_r \in \mathcal X_r.
\end{equation}
It can be verified that $Q^{(\delta)}_{j,r}$ has eigenvalues:
\begin{equation*}
    \lambda(Q^{(\delta)}_{j,r}) = \{ 1, 1 -\delta \}.
\end{equation*} 
Therefore as long as $0 < \delta < 1$, $Q^{(\delta)}_{j, r}$ is invertible and has all positive entries, which makes it a valid transition matrix. 
We also have the following form of the inverse of $Q^{(\delta)}_{j, r}$
\begin{equation}
    \left( Q^{(\delta)}_{j, r} \right)^{-1} = \frac{1}{1 - \delta}\left( I - \frac{\delta}{|\mathcal X_r|} \mathbf 1\mathbf 1^{\top} \right). 
\end{equation}
For the maximal clique $\mathcal C_{j}$, we further define the clique-wise transition matrix $Q^{(\delta)}_{j} \in \R^{ N_{j} \times N_{j}}_+$ as:
\begin{equation}
    Q^{(\delta)}_{j} \triangleq \bigotimes_{r \in \mathcal C_{j}} Q^{(\delta)}_{j, r},
\end{equation}
where $\bigotimes$ denotes the matrix Kronecker product. 
Each entry $Q^{(\delta)}_{j}(x_{\mathcal C_{j}}, u^{(\delta)}_{j})$ represents the backward test channel from $u^{(\delta)}_{j}$ to $x_{\mathcal C_j}$, i.e. $q_{X_{\mathcal C_{j}} | U^{(\delta)}_{j}} (x_{\mathcal C_{j}} | u^{(\delta)}_{j})$. 
For example, consider the $2$-order Markov source $(X_1, \cdots, X_5)$ in \Figref{fig:second_order_clique_graph}, we define the following constructions:
\begin{equation}
\begin{aligned}
    U^{(\delta)}_1 \triangleq (U^{(\delta)}_{1,1}, U^{(\delta)}_{1,2}, U^{(\delta)}_{1,3}) \xrightarrow{Q^{(\delta)}_1 \triangleq Q^{(\delta)}_{1,1} \otimes Q^{(\delta)}_{1,2} \otimes Q^{(\delta)}_{1,3}} (X_1, X_2, X_3) \\
    U^{(\delta)}_2 \triangleq (U^{(\delta)}_{2,2}, U^{(\delta)}_{2,3}, U^{(\delta)}_{2,4}) \xrightarrow{Q^{(\delta)}_2 \triangleq Q^{(\delta)}_{2,2} \otimes Q^{(\delta)}_{2,3} \otimes Q^{(\delta)}_{2,4}} (X_2, X_3, X_4) \\
    U^{(\delta)}_3 \triangleq (U^{(\delta)}_{3,3}, U^{(\delta)}_{3,4}, U^{(\delta)}_{3,5}) \xrightarrow{Q^{(\delta)}_3 \triangleq Q^{(\delta)}_{3,3} \otimes Q^{(\delta)}_{3,4} \otimes Q^{(\delta)}_{3,5}} (X_3, X_4, X_5)
\end{aligned}
\end{equation}
For $0 < \delta < 1$, $Q^{(\delta)}_{j}$ is also invertible since the Kronecker product of invertible matrices is also invertible. 
It can be verified that $Q^{(\delta)}_{j}$ has the following set of eigenvalues:
\begin{equation}
    \lambda(Q^{(\delta)}_{j}) = \{ 1, 1 - \delta, \cdots, (1 - \delta)^{K+1} \}.
\end{equation}
Accordingly, the eigenvalues for its inverse $(Q^{(\delta)}_{j})^{-1}$ can be written as $(1 - \delta)^{-k}$, $k = 0, \cdots, K+1$. 
Recall the positive vectorized clique factor $\psi_{j}$ for clique $\mathcal C_{j}$, given $Q^{(\delta)}_j$ constructed above, we need to further find a positive vectorized factor $\phi^{(\delta)}_j$ satisfying:
\begin{equation}
    \psi_j = Q^{(\delta)}_j \phi^{(\delta)}_j.
\end{equation}
Since $Q^{(\delta)}_j$ is invertible, it is equivalent to find a valid positive factor such that:
\begin{equation}\label{eq:u_factor}
    \phi^{(\delta)}_{j} \triangleq (Q^{(\delta)}_{j})^{-1} \psi_{j} \in \R^{N_{j}}.
\end{equation}
Now we will show that $Q^{(\delta)}_{j} \to I$ as $\delta \to 0$, thereby $\phi^{(\delta)}_{j} \to \psi_{j}$. 
Let $\|\cdot\|_2$ denotes the induced spectral norm of a matrix. 
It is known that the spectral norm of a positive semi-definite matrix is also its maximum eigenvalue. 
Notice that $(Q^{(\delta)}_j )^{-1} - I$ is a positive semi-definite matrix, then
\begin{equation}
    \left\|(Q^{(\delta)}_{j})^{-1} - I \right\|_2 = \lambda_{\max}\left( (Q^{(\delta)}_{j})^{-1} - I \right) = (1 - \delta)^{-(K+1)} - 1.
\end{equation}
By the definition of $\phi^{(\delta)}_{j}$ in \eqref{eq:u_factor}, we further have:
\begin{align}
    \| \phi^{(\delta)}_{j} - \psi_{j} \|_{\infty} &\le \| \phi^{(\delta)}_{j} - \psi_{j} \|_2 \\
    &= \left\| (Q^{(\delta)}_{j})^{-1} \psi_{j} - \psi_{j} \right\|_2 \\
    &= \left\| \left( (Q^{(\delta)}_{j})^{-1} - I \right) \psi_{j} \right\|_2 \\
    &\le \left\| (Q^{(\delta)}_{j})^{-1} - I \right\|_2 \| \psi_{j} \|_2 \\
    &= \left( (1 - \delta)^{-(K+1)} - 1 \right) \| \psi_{j} \|_2 \triangleq F(\delta).
\end{align}
We know that $\psi_{j}$ is strictly positive factor due to positive joint distribution of source $\mathbf X$.
By the definition of $U^{(\delta)}_j$ in \eqref{eq:u_aux_def}, we have $\mathcal U_j = \mathcal X_{\mathcal C_j}$.
Therefore, vectorized factors $\phi^{(\delta)}_{j}$ and $\psi_{j}$ have the same coordinate ordering. 
Then we can have:
\begin{align}
    \phi^{(\delta)}_{j}(u_{j}) &\ge \psi_{j}(u_{j}) - | \phi^{(\delta)}_{j}(u_{j}) - \psi_{j}(u_{j}) | \\
    &\ge c_{j} - \| \phi^{(\delta)}_{j} - \psi_{j} \|_{\infty} \\
    &\ge c_{j} - F(\delta).
\end{align}
Now we can set $\delta$ as following:
\begin{equation}
    0 < \delta < 1 - \left( 1 + \frac{ c_{j} }{ \|\psi_{j}\|_2 } \right)^{-\frac{1}{K+1}}.
\end{equation}
Then it can be verified that $F(\delta) < c_{j}$, thereby $\phi^{(\delta)}_{j} (u_{j}) > 0$ for every $u_{j}$. 
As we have $L-K$ maximal cliques for the $K$-order Markov source $\mathbf X$, we can define the following positive constant:
\begin{equation}
    \delta_0 \triangleq \min_{1 \le j \le L-K} \left\{ 1 - \left( 1 + \frac{ c_{j} }{ \|\psi_{j}\|_2 } \right)^{-\frac{1}{K+1}} \right\}
\end{equation}
Thus for any $0 < \delta < \delta_0$ and all $j$, $\phi^{(\delta)}_{j}$ is a valid positive vectorized factor. 
By \eqref{eq:u_factor}, we also have 
$\psi_{j}(x_{\mathcal C_{j}}) = \sum_{u^{(\delta)}_{j}} \phi^{(\delta)}_{j}(u^{(\delta)}_{j}) \prod_{r \in \mathcal C_{j}} Q^{(\delta)}_{j, r}(x_r, u^{(\delta)}_{j,r})$. 
Now we define the following augmented joint distribution:
\begin{equation}\label{eq:aug_joint_dist}
    q\!\left(\mathbf x, u^{(\delta)}_1, \cdots, u^{(\delta)}_{L-K} \right) \triangleq \prod^{L-K}_{j=1} \left( \phi^{(\delta)}_{j}(u^{(\delta)}_{j}) \prod_{r \in \mathcal C_{j}} Q^{(\delta)}_{j, r}(x_r, u^{(\delta)}_{j,r}) \right).
\end{equation}
First, we already know that $Q^{(\delta)}_{j, r}$ has all positive entries and $\phi^{(\delta)}_{j}$ is a positive vectorized factor for $0 < \delta < \delta_0$. 
Then $q\!\left(\mathbf x, (u^{(\delta)})^{L-K} \right)$ is also positive for $0 < \delta < \delta_0$. 
Second, we have:
\begin{equation}
\begin{aligned}
    \sum_{\mathbf x} \sum_{(u^{(\delta)})^{L-K}} q\!\left( \mathbf x, (u^{(\delta)})^{L-K} \right) &= \sum_{\mathbf x} \sum_{u^{(\delta)}_1} \cdots \sum_{u^{(\delta)}_{L-K}} \prod^{L-K}_{j = 1} \left( \phi^{(\delta)}_{j}(u^{(\delta)}_{j}) \prod_{r \in \mathcal C_{j}} Q^{(\delta)}_{j, r}(x_r, u^{(\delta)}_{j,r}) \right) \\
    &= \sum_{\mathbf x} \prod^{L-K}_{j = 1} \sum_{u^{(\delta)}_{j}} \left( \phi^{(\delta)}_{j}(u^{(\delta)}_{j}) \prod_{r \in \mathcal C_{j}} Q^{(\delta)}_{j, r}(x_r, u^{(\delta)}_{j,r}) \right) \\
    &= \sum_{\mathbf x} \prod^{L-K}_{j = 1} \psi_{j}(x_{\mathcal C_{j}}) \\
    &= \sum_{\mathbf x} p_{\mathbf X}(\mathbf x) = 1.
\end{aligned}
\end{equation}
Therefore, this is a valid positive joint distribution for $(\mathbf X, (U^{(\delta)})^{L-K})$. Next we show that the above constructed auxiliary variables and the corresponding joint distribution satisfy all the Markov constraints and the positive induced distortion. 

\underline{\textit{Markov conditions in \eqref{app-eq:general_u_markov_1}}}. Now we verify the constructed joint distribution satisfy Markov conditions in \eqref{app-eq:general_u_markov_1}. 
First, Markov conditions in \eqref{app-eq:general_u_markov_1} is equivalent to:
\begin{equation}
\begin{aligned}
    p(u^{L-K} | \mathbf x) = \prod^{L-K}_{j=1} p(u_{j} | x_{\mathcal C_{j}})
\end{aligned}
\end{equation}
Based on the joint distribution defined in \eqref{eq:aug_joint_dist}, we can write:
\begin{equation}
\begin{aligned}
    q\!\left((u^{(\delta)})^{L-K} | \mathbf x \right) &= \frac{ q\!\left(\mathbf x, (u^{(\delta)})^{L-K} \right) }{p_{\mathbf X}(\mathbf x)} \\
    &= \frac{ \prod^{L-K}_{j=1} \left( \phi^{(\delta)}_{j}(u^{(\delta)}_{j}) \prod_{r \in \mathcal C_{j}} Q^{(\delta)}_{j, r}(x_r, u^{(\delta)}_{j,r}) \right) }{ \prod^{L-K}_{j = 1} \psi_{j}(x_{\mathcal C_{j}}) } \\
    &= \prod^{L-K}_{j=1} \frac{ \phi^{(\delta)}_{j}(u^{(\delta)}_{j}) \prod_{r \in \mathcal C_{j}} Q^{(\delta)}_{j, r}(x_r, u^{(\delta)}_{j,r}) }{ \psi_{j}(x_{\mathcal C_{j}}) }
\end{aligned}
\end{equation}
Then it can be verified that $q(u^{L-K} | \mathbf x) = \prod^{L-K}_{j=1} q(u_{j} | x^{j+K}_{j})$ is true, thereby the Markov conditions in \eqref{app-eq:general_u_markov_1} are satisfied. 

\underline{\textit{Markov conditions in \eqref{app-eq:general_u_markov_2}}}. Now we verify the constructed joint distribution satisfy Markov conditions in \eqref{app-eq:general_u_markov_2}, which is equivalent to:
\begin{equation}
\begin{aligned}
    p(\mathbf x | u^{L-K}) &= p(x_1 | u_1) p(x_2 | u^2) \cdots p(x_{\ell} | u^{\ell}_{\ell-K}) \cdots p(x_L | u_{L-K}) \\
    &= \prod^L_{\ell = 1} p(x_{\ell} | u^{ \min\{\ell, L-K\} }_{ \max\{1, \ell - K\} }) = \prod^L_{\ell = 1} p(x_{\ell} | u_{\mathcal I_{\ell}}).
\end{aligned}
\end{equation}
Now for the constructed joint distribution in \eqref{eq:aug_joint_dist}, we can reorder the probability mass function as:
\begin{equation}
\begin{aligned}
    q\! \left(\mathbf x, (u^{(\delta)})^{L-K} \right) &= \left[ \prod^{L-K}_{j=1} \phi^{(\delta)}_{j}(u^{(\delta)}_{j}) \right] \left[ \prod^{L-K}_{j=1} \prod_{r \in \mathcal C_{j}} Q^{(\delta)}_{j,r}(x_r, u^{(\delta)}_{j,r}) \right] \\
    &= \left[ \prod^{L-K}_{j=1} \phi^{(\delta)}_{j}(u^{(\delta)}_{j}) \right] \left[ \prod^{L}_{\ell=1} \prod_{m \in \mathcal I_{\ell}} Q^{(\delta)}_{m, \ell}(x_{\ell}, u^{(\delta)}_{m,\ell}) \right] \\
    &= \left[ \prod^{L-K}_{j=1} \phi^{(\delta)}_{j}(u^{(\delta)}_{j}) \right] \left[ \prod^{L}_{\ell = 1} F_{\ell}(x_{\ell}, u^{(\delta)}_{\mathcal I_{\ell}}) \right]
\end{aligned}
\end{equation}
It can be verified that we have the following conditional distribution form:
\begin{equation}\label{app-eq:clique_back_test_ch}
    q\!\left(x_{j} | x_{-j}, (u^{(\delta)})^{L-K} \right) = \frac{ F_{j}(x_{j}, u^{(\delta)}_{\mathcal I_{j}}) }{ \sum_{ y \in \mathcal X_{j} } F_{j} (y, u^{(\delta)}_{\mathcal I_{j}}) },
\end{equation}
so the factorized form of this conditional distribution is only dependent on $u^{(\delta)}_{\mathcal I_{j}}$. Therefore, $X_{j} \indep ( X_{-j}, \mathbf U \setminus U_{\mathcal I_{j}} ) | U_{\mathcal I_{j}}$, which is equivalent to the Markov conditions in \eqref{app-eq:general_u_markov_2}.

\underline{\textit{Positive induced distortion}}. We have already shown that $\Delta_{\ell}(\delta) > 0$ if and only if $p_{X_{\ell} | V_{\ell}}$ is not point mass. 
For arbitrary $0 < \delta < \delta_0$, we can construct auxiliary variables $U^{(\delta)}_j \triangleq (U^{(\delta)}_{j,j}, \cdots, U^{(\delta)}_{j,j+K})$ and the corresponding joint distribution $q\!\left(\mathbf x, (u^{(\delta)})^{L-K} \right)$ with all positive entries. 
Therefore the induced distortion for each $X_{\ell}$ is also positive, i.e. $\Delta_{\ell}(\delta) > 0$ for $0 < \delta < \delta_0$. 
Specifically, for $X_{\ell}$ and a set of auxiliary variables $(U^{(\delta)})^{L-K}$ constructed above, we have the induced distortion:
\begin{align}
    \Delta_{\ell}(\delta) &\triangleq \min_{g_{\ell}} \E[d_{\ell}(X_{\ell}, g_{\ell}( U^{(\delta)}_{\mathcal I_{\ell}} ))] \\
    &= \sum_{u^{(\delta)}_{\mathcal I_{\ell}}} q(u^{(\delta)}_{\mathcal I_{\ell}}) \min_{\hat x_{\ell} \in \mathcal X_{\ell}} \sum_{x_{\ell}} Q^{(\delta)}(x_{\ell}, u_{\mathcal I_{\ell}}) d_{\ell}(x_{\ell}, \hat x_{\ell}) \\
    &\ge \sum_{u^{(\delta)}_{\mathcal I_{\ell}}} q( u^{(\delta)}_{\mathcal I_{\ell}} ) d_{\ell,\min} \min_{\hat x_{\ell}} \sum_{x_{\ell} \ne \hat x_{\ell}} Q^{(\delta)}(x_{\ell}, u_{\mathcal I_{\ell}}) > 0
\end{align}
where $\min_{\hat x_{\ell}} \sum_{x_{\ell} \ne \hat x_{\ell}} Q^{(\delta)}(x_{\ell}, u_{\mathcal I_{\ell}})$ is a positive function of $\hat x_{\ell}$. 
Moreover, for arbitrary $\delta$ such that $0 < \delta < \delta_0$, we can define the following distortion region:
\begin{equation}
    \mathcal D^{\mathrm{AUX}} \triangleq \{ \mathbf D: 0 < D_{\ell} < \Delta_{\ell}(\delta), 1 \le \ell \le L \}.
\end{equation}
Within this distortion region $\mathcal D^{\mathrm{AUX}}$, there always exist appropriate auxiliary variables $(U^{(\delta)})^{L-K}$ constructed as above. This completes the proof. 
\end{proof}

Lemma~\ref{lem:clique_auxiliary} shows that there always exist a class of clique-wise auxiliary variables for a $K$-order Markov source with positive distribution, such that the induced distortion for each source variable is also positive. 
Next we prove two Lemmas to characterize the RD optimal backward test channel for a positive distributed source. 
We first define the Neumann series which will be used in the proof. 
\begin{definition}[Neumann series]
\label{app-def:neumann_series}
For a matrix $A \in \mathbb{R}^{n \times n}$, the series $\sum_{k=0}^{\infty} A^k$, with $A^0 = I_n$, is called the Neumann series associated with $A$.
If $\|A\|_1 < 1$, where $\|\cdot\|_1$ denotes the induced matrix $1$-norm, this series converges, and $I_n - A$ is invertible with
\begin{equation}
    (I_n-A)^{-1} = \sum_{k=0}^{\infty} A^k.
\end{equation}
Moreover, we have
\begin{equation}
    \left\|(I_n - A)^{-1}-I_n \right\|_1 \le \sum_{k=1}^{\infty} \|A\|_1^k
    = \frac{\|A\|_1}{1-\|A\|_1}.
\end{equation}
\end{definition}

\begin{lemma}\label{lem:test_channel_converg}
Consider random variable $X \in \mathcal X$ with alphabet size $2 \le |\mathcal X| < \infty$ and positive distribution. Let $W\in \mathcal X$ denote arbitrary reconstruction of $X$ with $D \triangleq \E[d(X, W)]$ and $T$ the backward test channel from $W$ to $X$. 
Then for any joint distribution of $(X, W)$ satisfying $0 < D < p_{\min} d_{\min}$, we have:
\begin{enumerate}
\item $p_{W}$ is positive distribution, i.e. $p_W(w) > 0$ for all $w \in \mathcal X$.
\item $T$ is arbitrarily close to but not exact identity matrix:
\begin{align}\label{app-eq:tm_to_I_bound}
    \frac{2D}{d_{\max}} \le \|T - I\|_1 \le \frac{2D}{p_{\min} d_{\min} - D}.
\end{align}
\item If additionally $D < \frac{p_{\min} d_{\min} }{3}$, then $T$ is invertible and
\begin{equation}\label{app-eq:inv_tm_to_I_bound}
    \|T^{-1} - I\|_1 \le \frac{2D}{ p_{\min} d_{\min} - 3D }.
\end{equation}
\end{enumerate}
\end{lemma}
\begin{proof}
\textbf{Part 1. Positive reconstruction distribution.} \\
Consider arbitrary joint distribution $p_{XW}$ with $D = \mathbb E[d(X, W)]$.
We have:
\begin{align}
    D &= \mathbb E[d(X, W)] \\
    &= \sum_w \sum_{x \ne w} p_{XW}(x, w) d(x, w) \\
    &\ge d_{\min} \sum_w \sum_{x \ne w} p_{XW}(x, w) \\
    &= d_{\min} \cdot\mathrm{Pr}\{X \ne W\}.
\end{align}
Therefore $\mathrm{Pr}\{X \ne W\} \le \frac{D}{d_{\min}}$. 
By the coupling inequality for total variation distance, for any $w \in \mathcal X$ we have:
\begin{align}
    \mathrm{Pr}\{X \ne W\} & \ge d_{TV}(p_{X}, p_{W}) \\
    &= \sup_{\mathcal A \in \mathcal X} |p_{X}(\mathcal A) - p_{W}(\mathcal A)| \\
    &\ge p_{X}(w) - p_{W}(w) \\
    &\ge p_{\min} - p_W(w). 
\end{align}
Then for all $0 < D < d_{\min}p_{\min}$, we have $\mathrm{Pr}\{X \ne W\} \le \frac{D}{d_{\min}} < p_{\min}$, thereby: 
\begin{equation}
    p_W(w) \ge p_{\min} - \mathrm{Pr}\{X \ne W\} > 0.
\end{equation}
This completes the proof of Part 1. 

\textbf{Part 2. Transition matrix.} \\
For all $w\in \mathcal X$, we have:
\begin{align}
    \sum_x | T(x, w) - I(x, w) | &= 1 - T(w, w) + \sum_{x \ne w} T(x, w)  \\
    &= 2 \mathrm{Pr} \{ X \ne W | W = w \} \\
    &= 2 \frac{\mathrm{Pr}\{X \ne W, W = w\} }{p_{W}(w)} \\
    &\le 2 \frac{\mathrm{Pr} \{X \ne W\} }{ p_{W}(w) } \\
    &\le 2 \frac{\mathrm{Pr} \{X \ne W\} }{ p_{\min} - \mathrm{Pr} \{X \ne W\} } \\
    &\le \frac{ 2D }{ p_{\min}d_{\min} - D }.
\end{align}
Therefore by the definition of induced matrix $1$-norm, we have $\|T - I\|_1 \le \frac{ 2D }{ p_{\min}d_{\min} - D }$. This proves the upper bound in \eqref{app-eq:tm_to_I_bound}. 
In addition, we also have:
\begin{align}
    D &= \sum_{x \ne w} p_{XW}(x, w) d(x, w) \\
    &\le d_{\max} \sum_{x\ne w} p_{XW}(x, w) \\
    &= d_{\max} \sum_w p_W(w) \sum_{x\ne w}p_{X|W}(x | w) \\
    &= d_{\max} \sum_w p_W(w) \left( 1 - T(w, w) \right) \\
    &\le d_{\max} \max_w \left(1 - T(w, w)\right) \\
    &= \frac{d_{\max}}{2} \| T - I \|_1.
\end{align}
Therefore, we have $\|T - I\|_1 \ge \frac{2D}{d_{\max}} > 0$, which proves the lower bound in \eqref{app-eq:tm_to_I_bound}. 
This completes the proof of \eqref{app-eq:tm_to_I_bound}. 

\textbf{Part 3. Existence of inverse matrix.} \\
Now we define $E \triangleq I - T$, then:
\begin{equation*}
    \|E\|_1 = \left\|I - T \right\|_1 \le \frac{ 2D }{ p_{\min}d_{\min} - D }.
\end{equation*} 
If $D < \frac{p_{\min} d_{\min}}{3}$, then $\|E\|_1 < 1$. 
Therefore, consider $\sum^{\infty}_{k=0} E^k $ as the Neumann series, it converges and $T^{-1} = \sum^{\infty}_{k=0} E^k$ exists. 
By the definition of Neumann series in Definition~\ref{app-def:neumann_series}, we also have:
\begin{equation}
    T^{-1} - I = E^0 + \sum^{\infty}_{k=1} E^k - I = \sum^{\infty}_{k=1} E^k. 
\end{equation}
Then:
\begin{equation}
\begin{aligned}
    \| T^{-1} - I \|_1 &= \left\| \sum^{\infty}_{k=1} E^k \right\|_1 \\
    & \le \sum^{\infty}_{k=1} \|E\|^k_1 \\
    & \le \sum^{\infty}_{k=1} \left( \frac{2D}{p_{\min} d_{\min} - D} \right)^k \\
    & = \frac{2 D}{p_{\min} d_{\min} - 3 D}.
\end{aligned}
\end{equation}
Then we have \eqref{app-eq:inv_tm_to_I_bound}. This completes the proof. 
\end{proof}

For arbitrary random vector, recall the characterization of joint-coding rate-distortion function in Theorem~\ref{thm:jc_rd_func}, i.e.,
\begin{equation*}
    R^{\mathrm{JC}}(\mathbf D) \triangleq \min_{\mathcal P} I(\mathbf X; \hat{\mathbf X}). 
\end{equation*}
where $\mathcal P \triangleq \{ P_{ \hat{\mathbf X} | \mathbf X}: \E[d_{\ell}(X_{\ell}, \hat{X_{\ell}})] \le D_{\ell}, \; \ell = 1,\cdots, L \}$. 
For the per-letter distortion $D_{\ell}$, we have the general form of Shannon lower bound:
\begin{align}
    I(\mathbf X; \hat{\mathbf X}) &= H(\mathbf X) - H(\mathbf X | \hat{\mathbf X}) \\
    &\ge H(\mathbf X) - \sum^L_{\ell=1}H(X_{\ell} | \hat{X}_{\ell}) \\
    &= H(\mathbf X) - \sum^L_{\ell = 1} H(X_{\ell}) + \sum^L_{\ell = 1} I(X_{\ell}; \hat{X}_{\ell}) \\
    &\ge H(\mathbf X) - \sum^L_{\ell = 1} H(X_{\ell}) + \sum^L_{\ell = 1} R_{\ell}(D_{\ell}), \label{app-eq:general_slb}
\end{align}
where $R_{\ell}(D_{\ell})$ denotes the marginal rate-distortion function of source variable $X_{\ell}$. 
Next we prove that for source with positive distribution, the Shannon lower bound can be achieved within a non-empty distortion region. 

\begin{lemma}[SLB achievability]\label{app-lem:general_slb}
Consider arbitrary source $\mathbf X$ with finite component alphabets $2 \le |\mathcal X_{\ell}| < \infty$ and positive joint distribution. 
For per-letter distortion function in Definition~\ref{app-def:distortion}, there exists a non-empty distortion region $\mathcal D^{\mathrm{SLB}} = \{ \mathbf D: 0 < D_{\ell} \le \varepsilon_{\ell}, \ell = 1, \cdots, L \}$, such that for all $\mathbf D \in \mathcal D^{\mathrm{SLB}}$, the Shannon lower bound can be attained. 
\end{lemma}
\begin{proof}
The main idea of the proof is to construct a valid joint reconstruction distribution from the marginal rate-distortion optimal backward test channels, and show that the SLB can be achieved. 
First, we define the following positive constants:
\begin{gather*}
    p^*_{\min} \triangleq \min_{\mathbf x} p_{\mathbf X}(\mathbf x)  > 0, \qquad 
    p_{\ell,\min} \triangleq \min_{x_{\ell}} p_{X_{\ell}}(x_{\ell}), \qquad 
    d_{\ell,\min} \triangleq \min_{x_{\ell} \ne w_{\ell}} d_{\ell}(x_{\ell}, w_{\ell}), \\
    \tau \triangleq \frac{p^*_{\min}}{ 2 + p^*_{\min} }, \qquad 
    \varepsilon_{\ell} \triangleq \frac{p_{\ell,\min} d_{\ell,\min} \tau}{2L + \tau}.
\end{gather*}
Since $\tau < \frac{1}{2}$, we have $\varepsilon_{\ell} < \frac{p_{\ell,\min} d_{\ell,\min}}{3}$. 
Then we start the proof. 

\textbf{Step 1. Marginal RD problem.} For each $\ell$, select any $W_{\ell}$ that achieves the marginal RD function of $X_{\ell}$. 
Such $W_{\ell}$ exists because $X_{\ell}$ has finite alphabet. 
Denote the vectorized reconstruction probability of $p_{W_{\ell}}$ as $r_{\ell}$ and the backward transition matrix as:
\begin{equation}
    T_{\ell}(x_{\ell}, w_{\ell}) = p_{X_{\ell} | W_{\ell}}(x_{\ell} | w_{\ell}).
\end{equation}
By Lemma~\ref{lem:test_channel_converg}, for all $0 < D_{\ell} \le \varepsilon_{\ell} < \frac{p_{\ell, \min} d_{\ell, \min}}{3}$, we have:
\begin{equation}
    \|T_{\ell} - I\|_1 \le \frac{2 D_{\ell}}{p_{\ell,\min} d_{\ell,\min} - D_{\ell}} \le \frac{\tau}{L},
\end{equation}
Apparently, $T^{-1}_{\ell}$ exists and we further have:
\begin{equation}
    \| T^{-1}_{\ell} - I \|_1 \le \frac{2 D_{\ell}}{ p_{\ell,\min} d_{\ell,\min} - 3D_{\ell} } \le \frac{\tau}{L - \tau}.
\end{equation}

\textbf{Step 2. Joint backward test channel.} In this part, we first provide an explicit way to construct a valid joint backward test channel based on marginal test channels. 
Then we show that this constructed joint backward test channel achieves Shannon lower bound within a non-empty distortion region. 
We first define the product backward transition matrix form:
\begin{equation}
    S \triangleq \bigotimes^L_{\ell = 1} T_{\ell}.  
\end{equation}
Then for $\varepsilon_{\ell}$ defined above and all $\mathbf D \in \mathcal D^{\mathrm{SLB}}$, $S^{-1}$ exists. 
Let $p$ be the vectorized source distribution $p_{\mathbf X}$. 
Then we need to show that there exists a valid probability vector $r$ satisfying:
\begin{equation}
    r = S^{-1} p.
\end{equation}
Since each $T_{\ell}$ is column transition matrix, we have $\|T_{\ell}\|_1 = 1$ for all $\ell$. 
Let $N \triangleq \prod^L_{\ell = 1} |\mathcal X_{\ell}|$. 
Using the telescoping expansion:
\begin{equation}
    \bigotimes ^L_{\ell = 1}T_{\ell} - \bigotimes^L_{\ell = 1} I_{|\mathcal X_{\ell}|} = \sum^L_{\ell = 1} \left( \bigotimes^{\ell - 1}_{j = 1} T_j \right) \otimes (T_{\ell} - I_{|\mathcal X_{\ell}|}) \otimes \left( \bigotimes^{L}_{k=\ell + 1} I_{|\mathcal X_k|} \right),
\end{equation}
we have the following inequality:
\begin{align}
    \|S - I_N\|_1 &= \left\| \bigotimes ^L_{\ell = 1}T_1 - \bigotimes^L_{\ell = 1} I_{|\mathcal X_{\ell}|} \right\| \\
    &\le \sum^L_{\ell = 1} \left( \prod^{\ell-1}_{j=1} \|T_j\|_1 \right) \left\|T_{\ell} - I_{|\mathcal X_{\ell}|} \right\|_1 \left( \prod^L_{k=\ell + 1} \|I_{|\mathcal X_k|}\|_1 \right) \\
    &= \sum^L_{\ell = 1} \left\|T_{\ell} - I_{|\mathcal X_{\ell}|} \right\|_1 \le \tau < 1.
\end{align}
By again applying the Neumann series, we have:
\begin{equation}
    \|S^{-1} - I_N\| \le \frac{\tau}{1 - \tau} = \frac{p^*_{\min}}{2}.
\end{equation}
Then for the probability vectors, we have:
\begin{align}
    \max_{\mathbf w} |r(\mathbf w) - p_{\mathbf X}(\mathbf w)| &= \| r - p \|_{\infty} \\
    &\le \| r - p \|_1 \\
    &= \| (S^{-1} - I) p \|_1 \\
    &\le \| S^{-1} - I \|_1 \|p\|_1 \\
    &\le \frac{p^*_{\min}}{2}. 
\end{align}
Therefore, for the reconstruction probability vector $r$, we have:
\begin{enumerate}
    \item $r(\mathbf w) \ge p_{\mathbf X}(\mathbf w) - \frac{p^*_{\min}}{2} \ge \frac{p^*_{\min}}{2} > 0$. 
    \item $\mathbf 1^{\top} r = \mathbf 1^{\top} S^{-1} p = \mathbf 1^{\top} SS^{-1}p = \mathbf 1^{\top} p = 1$. 
\end{enumerate}
Thus $r$ is a valid positive probability distribution vector. 
Furthermore, by marginalizing the constructed joint distribution $S$, it can be verified that $p_{X_{\ell}} = T_{\ell} r_{\ell}$. 
Therefore, the marginal RD optimal backward test channels are preserved. 

\textbf{Step 3. Shannon lower bound achievability.} For any feasible reconstruction $\hat{\mathbf X}$, the general form of Shannon lower bound (SLB) is shown in \cref{app-eq:general_slb}. 
For the constructed auxiliary random variables $\mathbf W$, the product backward test channel and marginal RD optimality give:
\begin{equation}
    I(\mathbf X; \mathbf W) = H(\mathbf X) - \sum_{\ell} H(X_{\ell}) + \sum_{\ell} R_{\ell}(D_{\ell}).
\end{equation}
Therefore, the Shannon lower bound is attained by those constructed auxiliary variables $\mathbf W$ and the product backward test channel. 
Therefore, for all $0 < D_{\ell} < \varepsilon_{\ell}$, there always exist auxiliary variables $\mathbf W$ such that SLB is attained. 
This completes the proof. 
\end{proof}

By Lemma~\ref{lem:clique_auxiliary}, for a $K$-order Markov source $\mathbf X$ as in Definition~\ref{def:markov_pos_dist}, we can construct nontrivial auxiliary random variables $ U^{L-K}$ satisfying the Markov constraints in \cref{app-eq:general_u_markov_1,app-eq:general_u_markov_2}. 
Those constructed auxiliary variables also satisfy the Markov constraints in Definition~\ref{def:gfsc_k_ird}, thereby feasible for the GFSC architecture. 
For example, for first-order Markov, we can construct:
\begin{equation}
    X_1 \to U_1 \to X_2 \to U_2 \to \cdots \to U_{L-1} \to X_L.
\end{equation}
Apparently $X_{\ell} \to (U_{\ell-1}, U_{\ell}) \to X_{-\ell}$ holds. 
Next, to prove the JC rate-distortion optimality of GFSC architecture for general $K$-order Markov source, we introduce auxiliary random variables $\mathbf W$ satisfying the following properties:
\begin{enumerate}
    \item Define \( V_{\ell} \triangleq U_{\mathcal I_{\ell}} \) and \(X_{\ell} \to W_{\ell} \to V_{\ell} \) holds.
    \item The backward test channel from \(W_{\ell}\) to \(X_{\ell}\) is rate-distortion optimal. 
\end{enumerate}
Next, we show that such auxiliary variables can always be constructed within a nonempty distortion region for source with positive distribution. 

\begin{definition}
Let \(Q_{\ell} \in \R^{|\mathcal X_{\ell}| \times |\mathcal V_{\ell}|}\) be the transition matrix constructed in Lemma~\ref{lem:clique_auxiliary}, representing the backward test channel maps \(P_{V_{\ell}}\) to \(P_{X_{\ell}}\), i.e.,
$P_{X_{\ell}} = Q_{\ell} P_{V_{\ell}}$ with $Q_{\ell}(x, v) = p_{X_{\ell} | V_{\ell}}(x | v)$. 
\end{definition}
\begin{definition}
Let \(T_{\ell} \in \R^{|\mathcal X_{\ell}| \times |\mathcal W_{\ell}|} \) be the transition matrix constructed in Lemma~\ref{app-lem:general_slb}, representing the RD optimal backward test channel for \(X_{\ell}\), i.e., 
$P_{X_{\ell}} = T_{\ell} P_{W_{\ell}}$ with $T_{\ell}(x, w) = p_{X_{\ell} | W_{\ell}}(x | w)$.
\end{definition}
\begin{definition}
    For $Q_{\ell}$, define the positive term $q^{(\delta)}_{\ell,\min} \triangleq \min_{x, v} Q_{\ell}(x, v)$.
\end{definition}
Before we start, we define the following positive constant:
\begin{equation}
    \eta_{\ell}(\delta) \triangleq \frac{p_{\ell,\min} d_{\ell,\min} q^{(\delta)}_{\ell, \min} }{ 2 + 3 q^{(\delta)}_{\ell, \min} }. 
\end{equation}

\begin{lemma}[RD-optimal refinement of clique auxiliary variables]\label{app-lem:rd_optimal_refine}
Fix the clique-wise auxiliary variables constructed by \eqref{eq:aug_joint_dist} with $0 < \delta < \delta_0$. Assume the finite-alphabet and distortion conditions established above. 
Let $\varepsilon_{\ell} > 0$ be the distortion thresholds proved in Lemma~\ref{app-lem:general_slb}. 
Define $\epsilon_{\ell}(\delta) \triangleq \min\{\varepsilon_{\ell}, \; \eta_{\ell}(\delta) \}$ and the corresponding nonempty distortion region:
\begin{equation}\label{app-eq:opt_d_region}
    \mathcal D_{\epsilon} \triangleq \{ \mathbf D: 0 < D_{\ell} \le \epsilon_{\ell}(\delta), \; \ell = 1, \cdots, L \}.
\end{equation}
Then for any $\mathbf D \in \mathcal D_{\epsilon}$, there exist auxiliary variables $\mathbf W \in \mathcal X_1 \times \cdots \times \mathcal X_L$ such that:
\begin{enumerate}
    \item The Markov conditions hold:
    \begin{equation}\label{app-eq:opt_markov_condition}
        X_{\ell} \to W_{\ell} \to U^{(\delta)}_{\mathcal I_{\ell}} \to X_{-\ell}, \quad \ell = 1, \cdots, L.
    \end{equation}
    \item The reconstruction $\hat X_{\ell} \triangleq W_{\ell}$ and the corresponding backward test channel $p_{X_{\ell} | W_{\ell}}$ are jointly rate-distortion optimal. 
\end{enumerate}
\end{lemma}
\begin{proof}

\textbf{Step 1. Construction.} \\
For a given $\delta$ such that $0 < \delta < \delta_0$, by the definition of $\eta_{\ell}(\delta)$ and $\Delta_{\ell}(\delta)$, we have:
\begin{align}
    \Delta_{\ell}(\delta) &\triangleq \min_{g_{\ell}} \E[d_{\ell}(X_{\ell}, g_{\ell}( U^{(\delta)}_{\mathcal I_{\ell}} ))] \\
    &= \sum_{u^{(\delta)}_{\mathcal I_{\ell}}} q(u^{(\delta)}_{\mathcal I_{\ell}}) \min_{\hat x_{\ell} \in \mathcal X_{\ell}} \sum_{x_{\ell}} Q^{(\delta)}(x_{\ell}, u_{\mathcal I_{\ell}}) d_{\ell}(x_{\ell}, \hat x_{\ell}) \\
    &\ge \sum_{u^{(\delta)}_{\mathcal I_{\ell}}} q(u^{(\delta)}_{\mathcal I_{\ell}}) d_{\ell,\min} \min_{\hat x_{\ell}} \sum_{x_{\ell} \ne \hat x_{\ell}} Q^{(\delta)}(x_{\ell}, u_{\mathcal I_{\ell}}) \\
    &\ge \sum_{u^{(\delta)}_{\mathcal I_{\ell}}} q(u^{(\delta)}_{\mathcal I_{\ell}}) d_{\ell,\min} \min_{\hat x_{\ell}} \sum_{x_{\ell} \ne \hat x_{\ell}} q^{(\delta)}_{\ell,\min} \\
    &= (|\mathcal X_{\ell}| - 1) d_{\ell,\min} q^{(\delta)}_{\ell,\min} \ge \eta_{\ell}(\delta).
\end{align}
Then for $\mathcal D_{\epsilon}$ defined in \eqref{app-eq:opt_d_region}, we have:
\begin{equation}
    \mathcal D_{\epsilon} \subseteq \mathcal D^{\mathrm{SLB}}, \quad \mathcal D_{\epsilon} \subseteq \mathcal D^{\mathrm{AUX}}.
\end{equation}
Therefore, for all $\mathbf D \in \mathcal D_{\epsilon}$:
\begin{enumerate}
\item Lemma~\ref{app-lem:general_slb} provides a backward test channel construction from $\mathbf W$ to $\mathbf X$:
\begin{equation*}
    p_{\mathbf X | \mathbf W}(\mathbf x| \mathbf w) = \prod^L_{\ell = 1} T_{\ell} (x_{\ell}, w_{\ell}),
\end{equation*}
such that the Shannon lower bound can be attained. Thereby it is jointly rate-distortion optimal. $T_{\ell}$ denotes the marginal rate-distortion optimal test channel from $W_{\ell}$ to $X_{\ell}$.
\item Lemma~\ref{lem:clique_auxiliary} provides a backward test channel construction from $(U^{(\delta)})^{L-K}$ to $\mathbf X$ such that the induced distortion $\Delta_{\ell}(\delta)$ is positive and the following Markov conditions hold:
\begin{align*}
    & U^{(\delta)}_{j} \to X^{j + K}_{j} \to (\mathbf X, \; U^{(\delta)}_{-j}), && j=1,\ldots,L-K, \\
        & X_{\ell} \to U^{(\delta)}_{ \mathcal I_{\ell} } \to X_{-\ell}, && \ell = 1,\ldots,L.
\end{align*}
Denote the backward test channel from $V_{\ell} \triangleq U_{\mathcal I_{\ell}}$ to $X_{\ell}$ as $Q_{\ell}$. 
\end{enumerate}
Then it suffices to show that such constructed auxiliary variables $W_{\ell}$ and $V_{\ell}$ also satisfy the Markov conditions in \eqref{app-eq:opt_markov_condition}. 

\textbf{Step 2. Markov conditions.} \\
For constructed $T_{\ell} \in \R^{|\mathcal X_{\ell}| \times |\mathcal W_{\ell}|}$ and $Q_{\ell} \in \R^{ |\mathcal X_{\ell}| \times |\mathcal V_{\ell}| }$, the Markov condition in \eqref{app-eq:opt_markov_condition} boils down to finding a valid transition matrix \(A_{\ell} \in \R^{|\mathcal W_{\ell}| \times |\mathcal V_{\ell}|}\) such that
\begin{equation}
    Q_{\ell} = T_{\ell} A_{\ell}.
\end{equation}
As we already shown in Lemma~\ref{lem:test_channel_converg}, \(T_{\ell}\) is close to but not exactly an identity matrix and \(T^{-1}_{\ell}\) exists. 
This is equivalent to showing that \(A_{\ell} = T^{-1}_{\ell} Q_{\ell}\) is a valid transition matrix. \\
\underline{\textit{Non-negativity}}. \\
We first show that all the entries of \(T^{-1}_{\ell} Q_{\ell}\) are non-negative. 
First, by Lemma~\ref{lem:test_channel_converg}, we have
\begin{align}
    \| A_{\ell} - Q_{\ell} \|_1 &= \left\| (T^{-1}_{\ell} - I) Q_{\ell} \right\|_1 \\
    &\le \left\| T^{-1}_{\ell} - I \right\|_1 \|Q_{\ell}\|_1 \\
    &\le \frac{2D_{\ell}}{p_{\ell,\min}d_{\ell,\min} - 3D_{\ell}} \\
    &\le \frac{2\eta_{\ell}(\delta)}{ p_{\ell,\min}d_{\ell,\min} - 3 \eta_{\ell}(\delta) } \\
    &= q^{(\delta)}_{\ell,\min}. 
\end{align}
Then we have:
\begin{align}
    A_{\ell}(w_{\ell}, v_{\ell}) &= Q_{\ell}(w_{\ell}, v_{\ell}) - \left( Q_{\ell}(w_{\ell}, v_{\ell}) - A_{\ell}(w_{\ell}, v_{\ell}) \right) \\
    &\ge Q_{\ell}(w_{\ell}, v_{\ell}) - \| A_{\ell} - Q_{\ell} \|_1 \\
    &\ge q^{(\delta)}_{\ell,\min} - q^{(\delta)}_{\ell,\min}  = 0. \label{app-eq:valid_A_1}
\end{align}
Therefore all entries of $A_{\ell}$ are non-negative by construction. 

\underline{\textit{Column sum}}. \\
Moreover, for the column sum of $A_{\ell}$, we directly have:
\begin{equation}\label{app-eq:valid_A_2}
    \mathbf{1}^{\top} A_{\ell} = \mathbf{1}^{\top} T^{-1}_{\ell} Q_{\ell} = \mathbf{1}^{\top}T_{\ell} T^{-1}_{\ell}Q_{\ell} = \mathbf{1}^{\top}Q_{\ell} = \mathbf{1}^{\top}. 
\end{equation}
Combining \cref{app-eq:valid_A_1,app-eq:valid_A_2}, this proves that \(A_{\ell}\) is a valid transition matrix. 

Therefore, for any $\mathbf D \in \mathcal D_{\epsilon}$, we can always construct $\mathbf W$ with jointly rate-distortion optimal backward test channel. This completes the proof. 
\end{proof}

\begin{theorem}[JC optimality of GFSC]
Consider a $K$-order Markov source $\mathbf X$ with positive distribution. 
There exist a non-empty distortion region $\mathcal D_{\epsilon} \triangleq \{ \mathbf D: 0 < D_{\ell} \le \epsilon_{\ell}, \; \ell = 1, \cdots, L \ \}$, such that for all $\mathbf D \in \mathcal D_{\epsilon}$, the minimum information sum-rate $R^{\mathrm{GFSC}}_{K,\mathrm{sum}}(\mathbf{D})$ achieves the JC rate-distortion function:
\begin{equation}
    R^{\mathrm{GFSC}}_{K,\mathrm{sum}}(\mathbf D) = R^{\mathrm{JC}}(\mathbf D), \quad \forall \mathbf D \in \mathcal D_{\epsilon}.
\end{equation}
\end{theorem}
\begin{proof}
\textbf{Step 1. Achievability.} \\
Fix an admissible $\delta$ in the clique-wise auxiliary variable construction such that $0 < \delta < \delta_0$. 
Choose the positive threshold $\epsilon_{\ell}(\delta)$ supplied by the Lemma~\ref{app-lem:rd_optimal_refine}, and fix any $\mathbf D \in \mathcal D_{\epsilon}$. 
Then Lemma~\ref{app-lem:rd_optimal_refine} provides jointly distributed variables:
\begin{equation}
    \left( \mathbf X, (U^{(\delta)})^{L-K}, \mathbf W \right)
\end{equation}
with the following properties:
\begin{enumerate}
\item The source marginal is still the prescribed $p_{\mathbf X}$. 
\item The Markov conditions hold:
\begin{equation}
    X_{\ell} \to W_{\ell} \to U_{\mathcal I_{\ell}}, \quad \ell = 1, \cdots, L.
\end{equation}
\item The refinement auxiliary variables $\mathbf W$ achieve the joint-coding rate-distortion function $R^{\mathrm{JC}}(\mathbf D)$ for all $\mathbf D \in \mathcal D_{\epsilon}$. 
\end{enumerate}
We now construct augmented auxiliary variables satisfying the GFSC architecture. 
Specifically, we define:
\begin{equation}\label{app-eq:aug_general_aux_var}
    Z_{\ell} \triangleq \begin{cases}
        (U^{(\delta)}_{\ell}, W_{\ell}), & 1 \le \ell \le L-K, \\
        W_{\ell}, & L- K < \ell \le L.
    \end{cases}
\end{equation}
Apparently, auxiliary variables $\{Z_{\ell}\}^L_{\ell = 1}$ are just regrouping of constructed $\{U^{(\delta)}_j\}^{L-K}_{j=1}$ and $\mathbf W$. 
By \eqref{app-eq:aug_general_aux_var}, the previously received messages $Z^{\ell - 1}$ contains $\left( (U^{(\delta)})^{ \ell - 1 }, W^{\ell - 1} \right)$. 
We first verify the required encoder Markov conditions in \eqref{eq:korder_mc_def}, i.e.,
\begin{equation}
    Z_{\ell} \to (X^{\min \{\ell +K, L\}}_{\ell}, Z^{\ell - 1}) \to \mathbf X.
\end{equation}
For $1 \le \ell \le L-K$, we have the following:
\begin{equation}
    \left.
    \begin{array}{r}
        U^{(\delta)}_{\ell} \to X^{\ell + K}_{\ell} \to (\mathbf X, U^{(\delta)}_{-\ell} )  \\
        W_{\ell} \to (X_{\ell}, U^{(\delta)}_{\mathcal I_{\ell}} ) \to X_{-\ell}
    \end{array}
    \right\} \Rightarrow Z_{\ell} \to (X^{\ell + K}_{\ell}, U^{(\delta)}_{\mathcal I_{\ell}}) \to \mathbf X.
\end{equation}
For $L-K < \ell \le L$, we have $Z_{\ell} = W_{\ell}$. Then:
\begin{equation}
    Z_{\ell} \to (X_{\ell}, U^{(\delta)}_{\mathcal I_{\ell}} ) \to X_{-\ell}
\end{equation}
Therefore, the encoder Markov conditions in \eqref{eq:korder_mc_def} hold for every stage. 
Next, define the deterministic reconstruction as 
\begin{equation}
    \hat X_{\ell} = W_{\ell}
\end{equation}
For $\ell \le L-K$, the decoder extracts the second component of $Z_{\ell}$. For $\ell > L-K$, the decoder directly keep $Z_{\ell}$ itself. 
Thus this is admissible for the defined stronger decoder $g_{\ell}(Z^{\ell})$. 
For $\ell$-th encoding stage and augmented auxiliary variables $Z_{\ell}$, by the conditional source coding theorem, the associate rate $\bar{R}_{\ell}$ is:
\begin{equation}
    \bar{R}_{\ell} = I(X^{ \min\{\ell + K, L\}}_{\ell}; Z_{\ell} | Z^{\ell - 1})
\end{equation}
which also belongs to the information RD region in Definition~\ref{def:gfsc_k_ird}. 
By the Markov condition in \cref{eq:korder_mc_def}, we have
\begin{equation}
    I(X^{ \min\{\ell + K, L\}}_{\ell}; Z_{\ell} | Z^{\ell - 1}) = I(\mathbf X; Z_{\ell} | Z^{\ell - 1})
\end{equation}
Consequently, we have the following sum-rate form by chain rule:
\begin{align}
    \sum^L_{\ell = 1} \bar{R}_{\ell} &= \sum^L_{\ell = 1} I(X^{ \min\{\ell + K, L\}}_{\ell}; Z_{\ell} | Z^{\ell - 1}) \\
    &= \sum^L_{\ell = 1} I(\mathbf X; Z_{\ell} | Z^{\ell - 1}) \\
    &= I(\mathbf X; \mathbf Z) \\
    &= I(\mathbf X; U^{L-K}, \mathbf W) \\
    &= I(\mathbf X; \mathbf W) = R^{\mathrm{JC}}(\mathbf D).
\end{align}
Since this is just one possible realization of GFSC architecture, then we have:
\begin{equation}\label{app-eq:main_thm_upper_bound}
    R^{\mathrm{GFSC}}_{K,\mathrm{sum}}(\mathbf D) \le R^{\mathrm{JC}}(\mathbf D), \quad \forall \mathbf D \in \mathcal D_{\epsilon}.
\end{equation}

\textbf{Step 2. Converse.} \\
By the Shannon lower bound, we have:
\begin{equation}
    R^{\mathrm{GFSC}}_{K,\mathrm{sum}}(\mathbf D) \ge R^{\mathrm{SLB}}(\mathbf{D}) = R^{\mathrm{JC}}(\mathbf D), \quad \forall \mathbf D \in \mathcal D_{\epsilon}.
\end{equation}
Combine with \cref{app-eq:main_thm_upper_bound}, we have $R^{\mathrm{GFSC}}_{K,\mathrm{sum}}(\mathbf D)  = R^{\mathrm{JC}}(\mathbf D)$ for all $\mathbf D \in \mathcal D_{\epsilon}$. 
This completes the proof. 
\end{proof}

\section{Gauss-Markov source models under MSE distortion}

First, we review joint coding rate-distortion function for Gaussian source. 
Consider Gaussian source $\mathbf{X} \sim \mathcal{N}(0, \Sigma)$ with $\Sigma \succ 0$. 
Let $\Sigma_e \triangleq \mathbb{E}[(\mathbf{X} - \hat{\mathbf{X}})(\mathbf{X} - \hat{\mathbf{X}})^{\top}]$ be the error covariance, the JC rate-distortion function in Theorem~\ref{thm:jc_rd_func} admits a closed-form characterization
\begin{equation}\label{eq:jc_rd_gauss}
\begin{aligned}
    R^{\mathrm{JCG}}(\mathbf{D}) = &\ \min_{\Sigma_e} \frac{1}{2} \log \frac{|\Sigma|}{|\Sigma_e|} \quad \mathrm{s.t.} \quad 0 \prec \Sigma_e \preceq \Sigma,\; (\Sigma_e)_{\ell\ell} \le D_{\ell}.
\end{aligned}
\end{equation}
Moreover, in the distortion region $\mathcal{D}^{\mathrm{JC}} \triangleq \left\{ \mathbf{D} : \mathrm{diag}(D_1, \cdots, D_L) \preceq \Sigma \right\}$, it simplifies to
\begin{equation}\label{eq:slb}
    R^{\mathrm{JCG}}(\mathbf{D}) = \frac{1}{2} \log \frac{|\Sigma|}{\prod_{\ell=1}^L D_{\ell}}.
\end{equation}
This coincides with the Shannon lower bound (SLB)~\citep{cover1999elements}, which is tight over $\mathcal{D}^{\mathrm{JC}}$.
Next, we briefly review Gaussian graphical models (GGMs).

\subsection{Gaussian graphical model}\label{app-sec:ggm}
For Gaussian sources, as previously mentioned in Section~\ref{subsec:graphical_models}, its graphical structure can be obtained directly from its inverse covariance matrix, i.e. information matrix or precision matrix. 
Let $J\triangleq\Sigma^{-1}$ denote the information matrix. 
The corresponding Gaussian graphical model is the undirected graph $\mathcal{G}(J) = (\mathcal{V},\mathcal{E}(J))$, 
with node set $\mathcal{V}\triangleq\{1,\ldots,L\}$ and edge set $\mathcal{E}(J) \triangleq \{\{j,k\}: J_{jk}\neq 0 \}$, 
where \(J_{jk}\) denotes the \((j,k)\)-th entry. 
For any \(\mathcal{A}\subseteq\mathcal{V}\), write $X_{\mathcal{A}} \triangleq (X_i)_{i \in \mathcal{A} } $. 
The neighborhood of source node \(\ell\) is
\begin{equation*}
\mathcal{N}(\ell) \triangleq \{i\in\mathcal{V}\setminus\{\ell\}: \{\ell,i\}\in\mathcal{E}(J) \}.
\end{equation*}
For Gaussian sources, a missing edge has an exact conditional-independence interpretation:
\begin{equation*}
J_{jk}=0 \quad \Longleftrightarrow \quad X_j \indep X_k | X_{\mathcal{V} \setminus \{j,k\}}.
\end{equation*}
Consequently, each variable is conditionally independent of all non-neighboring variables given its local neighborhood, i.e. $X_\ell \indep X_{\mathcal{V}\setminus (\mathcal{N}(\ell)\cup\{\ell\})} | X_{\mathcal{N}(\ell)}$.
For a \(K\)-order Gauss-Markov source, \(J\) is \(K\)-banded, i.e. $J_{jk}=0$ for $|j - k|>K$.
Therefore,
\begin{equation*}
\mathcal{N}(\ell) \subseteq \{\ell-K,\ldots,\ell-1,\ell+1,\ldots,\ell+K\} \cap \mathcal{V}.
\end{equation*}
Thus, conditioned on at most \(2K\) neighboring frames, \(X_\ell\) is independent of all remaining source variables. 
Figure~\ref{fig:gauss_markov_graph} illustrates this structure for a second-order Gauss-Markov source, whose edges connect only frames separated by at most two time steps.

We introduce several definitions that can be used to specify a subset of edges in GGMs, and will be used later to define a sequential coding order for the Gaussian version coding system. 
For a $K$-order model, we first define
\begin{align}
    \mathcal{E}^{(K)} \triangleq \{(j,k): 1 \le j < k \le L,\; k-j \le K\}. \label{eq:kdelay_all_pairs}
\end{align}
Here, \eqref{eq:kdelay_all_pairs} contains all frame pairs within the $K$ neighbors, which also represents the edge set of a $K$-order Gauss-Markov model. 
As illustrated in Figure~\ref{fig:gauss_markov_graph}, $\mathcal{E}^{(2)}$ includes all pairs with form $(j, j+1)$ and $(j,j+2)$, which are processed according to the ordering in \eqref{eq:pair_order}. 
Then, we process the pairs in \(\mathcal E^{(K)}\) lexicographically. Specifically, let
\begin{equation}
(i,\ell)\prec(j,k) \quad\Longleftrightarrow \quad i<j \; \text{or} \; (i=j \; \text{and} \; \ell<k).
\label{eq:pair_order}
\end{equation}
For any pair-indexed collection $Z^N \triangleq (Z_{j,k})_{(j,k)\in\mathcal E^{(K)}}$ with $N\triangleq|\mathcal E^{(K)}|$, 
define
\begin{align}
Z_{<(j,k)} &\triangleq (Z_{i,\ell})_{(i,\ell)\in\mathcal E^{(K)}, (i,\ell)\prec(j,k) }, \label{eq:pair_history} \\
Z_{\leq j} &\triangleq (Z_{i,\ell})_{ (i,\ell)\in\mathcal E^{(K)}, i\leq j }.
\label{eq:stage_history}
\end{align}
Here, \(Z_{<(j,k)}\) contains all variables generated before stage \((j,k)\), whereas \(Z_{\leq j}\) contains all variables available after completing stage \(j\). We apply this convention to messages and auxiliary variables, yielding \(M_{<(j,k)}\), \(M_{\leq j}\), \(U_{<(j,k)}\), and \(U_{\leq j}\).

\subsection{GFSC with pair-wise factorization}
\label{app-subsec:gfsc_pair_system}
As discussed in Section~\ref{subsec:graphical_models}, the clique factorization can be further decomposed as pair-wise factorization. 
In addition, as discussed in Section~\ref{subsec:gauss_rd_opt}, the pair-wise interaction can be formalized as linear interpolation between frame pairs. 
We rename the further factorized system as \emph{Interpolation based Sequential Coding} (ISC). 
Let $\mathbf X = (X_1, \cdots, X_L)$ be a sequentially observed source and $K$ the allowed delay relaxation ($K < L-1$). 
We define the encoding and decoding mappings of ISC-$K$ architecture as follows. 

\textbf{Encoding.} The encoding proceeds sequentially over the set $\mathcal{E}^{(K)}$. 
For each stage $(j,k)\in\mathcal{E}^{(K)}$, the encoder first constructs a ``virtual frame'' $\tilde{X}_{j,k}$ 
with an pair-wise operator $h_{j,k}: \mathcal{X}_j \times \mathcal{X}_k \to \tilde{\mathcal{X}}_{j,k}$.
$\tilde{X}_{j,k}$ is then encoded into message using a zero-delay sequential encoder 
$\phi_{j,k}: \tilde{\mathcal{X}}_{j,k} \times \mathcal{M}_{<(j,k)} \times \mathcal{W} \to \mathcal{M}_{j,k}$. 
The encoding mapping is given by $f_{j,k} = \phi_{j,k} \circ h_{j,k}$, i.e.,
\begin{equation}
f_{j,k}: \mathcal{X}_j \times \mathcal{X}_k \times \mathcal{M}_{<(j,k)} \times \mathcal{W} \to \mathcal{M}_{j,k}, \quad (j,k) \in \mathcal E^{(K)}.
\end{equation}
This formulation makes explicit that delay is handled through interpolation between $(X_j,X_k)$, while the coding process itself remains strictly sequential.
There are generally $N \triangleq |\mathcal E^{(K)}|$ encoding stages. $N = L-1$ for $1$-delay system and $N \ge L$ for $K > 1$ delay system.

\textbf{Decoding.} The $\ell$-th decoder reconstructs $X_{\ell}$ using all messages available up to $\ell$-stage decoding. 
The decoding mapping is given by:
\begin{equation}
g_{\ell}: \mathcal{M}_{\le\ell} \times \mathcal{W} \to \hat{\mathcal{X}}_{\ell}, \quad \ell = 1,\cdots,L.
\end{equation}
There are $L$ decoding stages for $L$ source frames respectively. So encoders and decoders are not necessarily matched one-by-one. 
Notice that we can combine and define a multi-stage encoder $\tilde{f}_{\ell} \triangleq f_{\ell, \ell + K} \circ \cdots \circ f_{\ell, \ell+1}$ with input $X^{\ell + K}_{\ell}$. This can be seen as a special case of GFSC architecture.

\subsection{Rate-distortion analysis}\label{app:rd_region}

For convenience, we only characterize the information RD region and sum-rate function of ISC-$K$ architecture.
The analysis imposes no assumptions on the source distribution and therefore applies to general sources. 
We first introduce the notation used throughout this section. 
For the length-$L$ source and distortion sequences, define:
\begin{equation*}
    \mathbf{X} \triangleq (X_1, \cdots, X_L), \qquad  \mathbf{D} \triangleq (D_1, \cdots, D_L),
\end{equation*}
For sequences indexed by the ordered set $\mathcal E^{(K)}$, define:
\begin{equation*}
    U^N \triangleq \{U_{j,k} : (j,k)\in \mathcal{E}^{(K)}\}, \quad
    R^N \triangleq \{R_{j,k} : (j,k)\in \mathcal{E}^{(K)}\},
\end{equation*}
where $N \triangleq |\mathcal E^{(K)}|$ and the entries follow the prescribed ordering of $\mathcal E^{(K)}$. 
With this notation in place, we begin with the definition of information rate-distortion region associated with the ISC-\(K\) system.

\begin{definition}[Information RD region of ISC-$K$]\label{app-def:vf_k_ird}
For a given source $\mathbf{X}$, the single-letter information RD region of ISC-$K$, denoted by $\mathcal{R}_K$, consists of all $(R^N, \mathbf{D})$ satisfying
\begin{align}
    R_{j,k} &\ge I(\tilde{X}_{j,k}; U_{j,k} | U_{<{j,k}}), 
    \quad (j,k) \in \mathcal{E}^{(K)}, \label{app-eq:korder_rate} \\
    D_{\ell} &\ge \mathbb{E} [\|X_{\ell} - \hat{X}_{\ell}\|^2], 
    \quad \ell = 1, \cdots, L, \label{app-eq:korder_dist}
\end{align}
for all auxiliary random variables $\{ \tilde{X}_{j,k}, U_{j,k} \}_{(j,k) \in \mathcal{E}^{(K)}}$ and $\hat{\mathbf{X}}$ satisfying $\tilde{X}_{j,k} = h_{j,k}(X_j, X_k)$, $\hat{X}_{\ell} = g_{\ell} (U_{\le{\ell}} )$ and the Markov constraints: 
\begin{equation}\label{app-eq:korder_mc_def}
\begin{gathered}
    U_{j,k} \to (\tilde{X}_{j,k}, U_{<{j,k}}) \to (X_j, X_k, U_{<{j,k}}) \to \mathbf{X},
\end{gathered}
\end{equation}
where $h_{j,k}(\cdot)$ and $g_{\ell}(\cdot)$ are deterministic functions. 
For a given $\mathbf{D}$, the information rate region $\mathcal{R}_{K}(\mathbf{D})$ is the closure of all $R^N$ such that $(R^N, \mathbf{D}) \in \mathcal{R}_K$.
\end{definition}

\begin{theorem}[Information minimum sum-rate]\label{app-thm:min_sum_rate}
For a given distortion tuple $\mathbf{D}$, the information minimum sum-rate function of ISC-$K$ is
\begin{equation}\label{app-eq:sum_rate}
    R^{\mathrm{ISC}}_{K,\mathrm{sum}}(\mathbf{D}) = \min I(\mathbf{X}; U^N ).
\end{equation}
The minimum is taken over all auxiliary variables $\{U_{j,k}\}_{(j,k)\in\mathcal{E}^{(K)}}$ satisfying Markov chains in \eqref{app-eq:korder_mc_def} and the distortion constraints
$D_{\ell} \ge \mathbb{E}[\|X_{\ell} - g_{\ell}(U_{\le{\ell}}) \|^2]$ for all $\ell$. 
\end{theorem}
\begin{proof}
From Definition~\ref{app-def:vf_k_ird}, for a given distortion tuple $\mathbf{D}$, the minimum sum-rate is given by
\begin{equation}
    R^{\mathrm{ISC}}_{K,\mathrm{sum}}(\mathbf{D}) = \min \sum_{(j,k)\in\mathcal{E}^{(K)}} I(\tilde{X}_{j,k}; U_{j,k} | U_{<{j,k}}),
\end{equation}
where the minimization is over all feasible auxiliary variables satisfying the constraints in \eqref{app-eq:korder_mc_def} and the distortion constraints.
Since $\tilde{X}_{j,k} = h_{j,k}(X_j, X_k)$ is a deterministic function of $(X_j, X_k)$, we have
\begin{equation*}
    I(\tilde{X}_{j,k}; U_{j,k} | U_{<{j,k}}) \le I(X_j, X_k; U_{j,k} | U_{<{j,k}}).
\end{equation*}
On the other hand, by the Markov chain $U_{j,k} \to (\tilde{X}_{j,k}, U_{<{j,k}} ) \to (X_j, X_k, U_{<{j,k}})$, the data processing inequality implies
\begin{equation*}
    I(\tilde{X}_{j,k}; U_{j,k} | U_{<{j,k}}) \ge  I(X_j, X_k; U_{j,k} | U_{<{j,k}}).
\end{equation*}
Combining the above two inequalities gives $I(\tilde{X}_{j,k}; U_{j,k} | U_{<{j,k}}) = I(X_j, X_k; U_{j,k} | U_{<{j,k}})$.
Substituting this into the sum-rate expression, we have
\begin{align}
    R^{\mathrm{ISC}}_{K,\mathrm{sum}}(\mathbf{D})
    &= \min \sum_{(j,k)\in\mathcal{E}^{(K)}} I(X_j, X_k; U_{j,k} | U_{<{j,k}}) \label{app-eq:sum_rate_0} \\
    &= \min \sum_{(j,k)\in\mathcal{E}^{(K)}} I(\mathbf{X}; U_{j,k} | U_{<{j,k}}) \label{app-eq:sum_rate_1} \\
    &= \min I(\mathbf{X}; U^N), \label{app-eq:sum_rate_2}
\end{align}
where \eqref{app-eq:sum_rate_1} follows from the Markov chain $U_{j,k} \to (X_j, X_k, U_{<{j,k}}) \to \mathbf{X}$ and \eqref{app-eq:sum_rate_2} follows from the chain rule of mutual information under the ordering induced by $\mathcal{E}^{(K)}$. 
This completes the proof.
\end{proof}

\subsection{Joint-coding optimality for gauss-markov sources}\label{app:jc_optimality}

In this section, we prove that the ISC architecture achieves JC rate-distortion optimality for Gauss-Markov sources within a nontrivial distortion region. 
Typically, the Gaussian graphical model (GGM) associated with a Gaussian source can be represented by $\mathcal G = (\mathcal V, \mathcal E)$, where $\mathcal E = \{ \{i, j\}: J_{ij} \ne 0 \}$ denotes the set of undirected edges. 
For notational convenience, in this section, we identify each undirected edge $\{i,j\}$ with the unique ordered pair $(i, j)$ satisfying $i < j$ and write $\mathcal E = \{ (i, j): J_{ij} \ne 0, i < j \}$.
This convention merely provides a unique indexing of the edges. 
It neither assigns directions to edges nor changes the underlying graph. 

\begin{lemma}[Existence of a nontrivial distortion region]
\label{app-lem:exist_eps}
Consider correlated Gaussian vector $\mathbf{X} \sim \mathcal{N}(0, \Sigma)$ with $\Sigma \succ 0$ and the GGM $\mathcal{G}(J)$ induced by its information matrix $J \triangleq \Sigma^{-1}$. Let $\{ U_{j,k}\}_{ (j, k) \in \mathcal{E}}$ be a set of Gaussian random variables jointly distributed with $\mathbf{X}$ satisfying:
\begin{equation}\label{app-eq:opt_mc_1}
    U_{j,k} = \omega_{j,k} X_j + ( 1- \omega_{j,k}) X_k + N_{j,k}, \quad (j,k) \in \mathcal{E},
\end{equation}
and the Markov chains:
\begin{gather}
    X_j \to \left( X_{\mathcal{N}(j) \setminus \{k\}}, U_{j,k} \right) \to X_k, \quad (j,k) \in \mathcal E, \label{app-eq:opt_mc_2} \\
    X_{\ell} \to U_{\ell, \mathcal{N}(\ell)} \to X_{-\ell}, \quad \ell = 1, \cdots, L, \label{app-eq:opt_mc_3}
\end{gather}
where $N_{j,k}\sim \mathcal{N}(0, \sigma^2_{N_{j,k}})$ are Gaussian random variables independent of $\mathbf{X}$. 
Then for any $\omega_{j,k} \in \mathbb{R}\backslash \{0,1\}$, the induced MSE distortion for each source variable $X_{\ell}$, denoted as $\epsilon_{\ell} \triangleq \mathbb{E}[\|X_{\ell} - \mathbb{E}[X_{\ell} | U_{\ell, \mathcal{N}(\ell)}]\|^2]$, is nontrivial, i.e.,
\begin{equation}\label{app-eq:distortion_bound}
    0 < \epsilon_{\ell} < \bar{D}_{\ell},
\end{equation}
where $\bar{D}_{\ell} \triangleq \mathbb{E}[ \| X_{\ell} - \mathbb{E}[X_{\ell} | X_{\mathcal{N}(\ell)} ] \|^2 ]$.
\end{lemma}
\begin{proof}
\underline{\textit{Construction of appropriate auxiliary variables}}: \\ 
First we show how to construct auxiliary random variables satisfying \cref{app-eq:opt_mc_1,app-eq:opt_mc_2,app-eq:opt_mc_3}. 
For Gaussian source $\mathbf{X}\sim \mathcal{N}(0, \Sigma)$ with information matrix $J \triangleq \Sigma^{-1}$, its probability density function can be written as:
\begin{align}
    p(\mathbf{x}) &= \frac{1}{ (2\pi)^{\frac{L}{2}} |\Sigma|^{\frac{1}{2}} } \exp \left( -\frac{1}{2} \mathbf{x}^{\top} J \mathbf{x}  \right) \\
    &= \frac{1}{ (2\pi)^{\frac{L}{2}} |\Sigma|^{\frac{1}{2}} } \exp \left( -\frac{1}{2} \sum^L_{\ell=1} J_{\ell \ell} x^2_{\ell} - \sum_{1 \le j < k \le L} J_{jk} x_j x_k  \right) \label{app-eq:gauss_density_J_1} \\
    &= \frac{1}{ (2\pi)^{\frac{L}{2}} |\Sigma|^{\frac{1}{2}} } \exp \left( -\frac{1}{2} \sum^L_{\ell=1} J_{\ell \ell} x^2_{\ell} - \sum_{(j,k) \in \mathcal{E}} J_{jk} x_j x_k \right), \label{app-eq:gauss_density_J_2}
\end{align}
where \eqref{app-eq:gauss_density_J_1} follows from that $J$ is a symmetric matrix, and \eqref{app-eq:gauss_density_J_2} follows from that $J_{jk} = 0$ for $(j,k) \notin \mathcal{E}$. 
Thus, the dependency structure of a Gaussian source $\mathbf{X}$ is completely characterized by the edge set $\mathcal{E}$ of its GGM. 
By \eqref{app-eq:opt_mc_1}, each auxiliary variable $U_{j,k}$ is associated with edge $(j,k)$ in the GGM and independent additive Gaussian noise $N_{j,k}$. 
Therefore, its conditional density function given $(X_j, X_k) = (x_j, x_k)$ can be written as:
\begin{equation}
    p(u_{j,k} | x_j, x_k) = \frac{1}{\sqrt{2\pi \sigma^2_{N_{j,k}}}} \exp\left( -\frac{ (u_{j,k} - \omega_{j,k}x_j - \bar{\omega}_{j,k}x_k)^2 }{2\sigma^2_{N_{j,k}}} \right),
\end{equation}
where $\bar{\omega}_{j,k} \triangleq 1 - \omega_{j,k}$. 
In addition, since those noise items $\{N_{j,k}\}_{(j,k)\in\mathcal{E}}$ are also mutually independent with each other, $\{U_{j,k}\}_{(j,k)\in\mathcal E}$ are conditionally independent given $\mathbf{X}$, i.e., $p(\mathbf{u}|\mathbf{x}) = \prod_{ (j,k) \in \mathcal{E}} p(u_{j,k} | x_j, x_k)$.
Applying Bayes' rule, we have $p(\mathbf{x} | \mathbf{u}) \propto p(\mathbf{x}) p(\mathbf{u} | \mathbf{x})$, which yields:
\begin{equation}
    p(\mathbf{x}|\mathbf{u}) \propto \exp\left( -\frac{1}{2} \sum^L_{\ell=1} J_{\ell \ell} x^2_{\ell} - \sum_{\{j,k\} \in \mathcal{E}} J_{jk} x_j x_k - \sum_{\{j,k\}\in\mathcal{E} } \frac{ (u_{j,k} - \omega_{j,k}x_j - \bar{\omega}_{j,k}x_k)^2 }{2\sigma^2_{N_{j,k}}} \right)
\end{equation}
Expanding the quadratic terms and collecting coefficients gives
\begin{equation}\label{app-eq:cond_density_expand}
    p(\mathbf{x}|\mathbf{u}) \propto \exp\left( -\frac{1}{2}\sum^L_{\ell=1}\lambda_{\ell}x^2_{\ell} + \sum^L_{\ell=1} \gamma_{\ell}(\mathbf{u}) x_{\ell} - \sum_{(j,k) \in \mathcal{E}} \left(J_{jk} + \frac{\omega_{j,k} \bar{\omega}_{j,k}}{\sigma^2_{N_{j,k}}} \right) x_jx_k \right),
\end{equation}
where
\begin{gather}
    \lambda_{\ell} \triangleq J_{\ell \ell} + \sum_{(\ell, k) \in \mathcal{E}} \frac{ \omega^2_{\ell,k} }{\sigma^2_{N_{\ell,k}}} + \sum_{(j, \ell) \in \mathcal E} \frac{ \bar{\omega}^2_{j,\ell} }{ \sigma^2_{N_{j,\ell}} }, \\
    \gamma_{\ell}(\mathbf{u}) \triangleq \sum_{(\ell, k) \in \mathcal{E}} \frac{\omega_{\ell,k}}{\sigma^2_{N_{\ell,k}}} u_{\ell,k} + \sum_{(j,\ell) \in \mathcal E} \frac{\bar{\omega}_{j,\ell}}{\sigma^2_{N_{j,\ell}}} u_{j,\ell}.
\end{gather}
The Markov constraints in \cref{app-eq:opt_mc_2,app-eq:opt_mc_3} imply that the source variables are conditionally independent given the adjacent auxiliary variables, i.e. $p(\mathbf{x}\mid\mathbf{u}) = \prod_{\ell=1}^{L} p(x_{\ell}|\mathbf{u})$.
Therefore, all pairwise interaction terms $x_jx_k$ in \eqref{app-eq:cond_density_expand} must vanish, i.e.,
\begin{equation}\label{app-eq:mc_noise_var}
    J_{jk} + \frac{\omega_{j,k} \bar{\omega}_{j,k}}{\sigma^2_{N_{j,k}}} = 0,
\end{equation}
which yields $\sigma^2_{N_{j,k}} = -\frac{\omega_{j,k} \bar{\omega}_{j,k}}{J_{jk}}$. 
As $\sigma^2_{N_{j,k}}$ denotes the noise variance, to ensure $\sigma^2_{N_{j,k}} > 0$, we need to set $\omega_{j,k} \in (0, 1)$ if $J_{jk} < 0$, and set $\omega_{j,k} > 1$ otherwise. 
With appropriate coefficients $\omega_{j,k}$, the constructed auxiliary variables apparently satisfy Markov constraints in \cref{app-eq:opt_mc_1,app-eq:opt_mc_2,app-eq:opt_mc_3}. 
This completes the construction of the auxiliary random variables.

\underline{\textit{Nontrivial MSE distortion analysis}}: \\
We now establish the distortion bounds for the auxiliary random variables $\{U_{j,k}\}_{(j,k) \in \mathcal E}$ constructed above. 
Without loss of generality, we consider the source node $\ell = 1$. 
Suppose $\mathbf{X}$ is a pairwise correlated Gaussian vector, the set of neighbors for $X_1$ is apparently $\mathcal{N}(1)=\{2,\ldots,L\}$.

First, for the upper bound in \eqref{app-eq:distortion_bound}, i.e.,
\begin{equation}
    \epsilon_1 = \mathbb{E}[\| X_1 - \mathbb{E}[X_1 | U_{1, \mathcal{N}(1) }] \| ^2] < \bar{D}_1,
\end{equation}
this follows directly from the Markov chain conditions in \cref{app-eq:opt_mc_1,app-eq:opt_mc_2,app-eq:opt_mc_3}, which guarantee that the auxiliary random variables $U_{1,\mathcal N(1)}$ capture all the dependence of $X_1$ on the remaining source variables. 
Next, we prove the strict positivity in~\eqref{app-eq:distortion_bound}. Since the MSE is always nonnegative, it suffices to show that $\epsilon_1 \neq 0$. 
Equivalently, it suffices to show that $X_1$ cannot be recovered exactly from $U_{1,\mathcal{N}(1)}$. 
Suppose, for contradiction, that
\begin{equation}\label{app-eq:conditional_expect}
    X_1 = \mathbb{E}[X_1\mid U_{1,\mathcal{N}(1)}] \quad \mathrm{a.s.}
\end{equation}
Since $(X_1,U_{1,\mathcal{N}(1)})$ are jointly Gaussian and zero mean, the conditional expectation is thereby a linear transformation. 
Therefore, \eqref{app-eq:conditional_expect} is equivalent to that there exist constants $a_2,\ldots,a_L$ such that
\begin{align}
    X_1 = \sum_{k=2}^{L} a_k U_{1,k} = \sum_{k=2}^{L} a_k \left( \omega_{1,k}X_1 + \bar{\omega}_{1,k}X_k + N_{1,k} \right) \quad \mathrm{a.s.}
\end{align}
Rearranging terms gives
\begin{equation}\label{app-eq:lb_condition}
    \left( 1 - \sum^L_{k=2} a_k \omega_{1,k} \right) X_1 - \sum^L_{k=2} a_k\bar{\omega}_{1,k} X_k = \sum^L_{k=2} a_k N_{1,k} \quad \mathrm{a.s.}
\end{equation}
The left-hand side of~\eqref{app-eq:lb_condition} is a linear function of $\mathbf{X}$, whereas the right-hand side is a linear function of the independent Gaussian noises $\{N_{1,k}\}_{k=2}^{L}$. 
Hence the two sides are two independent Gaussian random variables. 
As they are equal almost surely, both sides must be almost surely constant. 
Because all variables have zero mean, this constant must be zero. 
Thus, we need 
\begin{gather}
    \left( 1-\sum_{k=2}^{L}a_k\omega_{1,k} \right)X_1 - \sum_{k=2}^{L} a_k \bar{\omega}_{1,k}X_k = 0 \quad \mathrm{a.s.}, \label{app-eq:lb_condition_1}\\
    \sum_{k=2}^{L}a_kN_{1,k} = 0 \quad \mathrm{a.s.}, \label{app-eq:lb_condition_2}
\end{gather}
where \eqref{app-eq:lb_condition_1} implies that:
\begin{equation}
\begin{bmatrix}
    1 - \sum^L_{k=2} a_k\omega_{1,k} \\
    -a_2\bar{\omega}_{1,2} \\
    \vdots \\
    -a_L\bar{\omega}_{1,L}
\end{bmatrix} \in \mathrm{ker}(\Sigma),
\end{equation}
Since $\Sigma \succ 0$, we have $\ker(\Sigma)=\{\mathbf{0}\}$. 
Therefore, \eqref{app-eq:lb_condition_1} is equivalent to 
\begin{equation}
    1-\sum_{k=2}^{L}a_k\omega_{1,k}=0, \qquad a_k\bar{\omega}_{1,k}=0 \;\text{for } k=2,\ldots,L.
\end{equation}
Because appropriate auxiliary variable construction requires $\omega_{1,k} \in \mathbb{R}\setminus\{0,1\}$, which implies $\omega_{1,k}, \bar{\omega}_{1,k} \neq 0$.
Therefore, $a_k\bar{\omega}_{1,k}=0$ requires $a_k=0$ for all $k=2,\ldots,L$. 
This contradicts $1-\sum_{k=2}^{L}a_k\omega_{1,k}=0$. 
Therefore, \eqref{app-eq:lb_condition_1} cannot hold for appropriate constructed auxiliary variables. 
Now for \eqref{app-eq:lb_condition_2}, since the noises 
$\{N_{1,k}\}_{k=2}^{L}$ are mutually independent Gaussian random variables with zero mean, \eqref{app-eq:lb_condition_2} holds if and only if
\begin{equation}
    \mathrm{Var}\!\left(\sum_{k=2}^{L}a_kN_{1,k}\right) = \sum_{k=2}^{L}a_k^2\sigma^2_{N_{1,k}} = 0.
\end{equation}
Therefore, we must have either $a_k=0$ or $\sigma^2_{N_{1,k}}=0$ for $k=2,\ldots,L$. 
Recall that for appropriate constructed auxiliary variables, we have $\sigma^2_{N_{1,k}} = - \frac{ \omega_{1,k}\bar{\omega}_{1,k} }{ J_{1k} }$,
where $J_{1k} \neq 0$. 
Since $\omega_{1,k}\in\mathbb{R}\setminus\{0,1\}$, we have $\omega_{1,k}\bar{\omega}_{1,k}\neq0$ and thereby $\sigma^2_{N_{1,k}} > 0$. 
Therefore \eqref{app-eq:lb_condition_2} implies that $a_k = 0$, for $k=2,\ldots,L$. 
Substituting this into $X_1 = \sum_{k=2}^{L}a_kU_{1,k}$ yields $X_1 = 0$ almost surely, which contradicts $\Sigma\succ0$. Hence~\eqref{app-eq:lb_condition_2} also cannot hold for appropriate constructed auxiliary variables. 
Combining the above arguments, we conclude that $\epsilon_1$ is strictly positive and finite. 

This proof can be easily extended to all remaining source nodes $X_{\ell}$, $\ell = 2, \cdots, L$, with the same collection of auxiliary random variables $\{U_{j,k}\}_{(j,k)\in\mathcal{E}}$.
Therefore, let $\epsilon_{\ell} \triangleq \mathbb{E}[\| X_{\ell} - \mathbb{E}[X_{\ell} | U_{\ell, \mathcal{N}(\ell)}]\|^2 ]$ denotes the induced distortion for each source node $X_{\ell}$, we have $0 < \epsilon_{\ell} < \bar{D}_{\ell}$, for $\ell = 1, \cdots, L$. 
This completes the proof.
\end{proof}

The above Lemma~\ref{app-lem:exist_eps} mainly shows that the existence of appropriate edge-wise auxiliary variables $\{U_{j,k}\}_{(j,k) \in \mathcal E^{(K)}}$ for Gaussian source, which induce nontrivial MSE distortion $\epsilon_{\ell}$ (strictly positive and bounded) for each source variable $X_{\ell}$. 
A specific distortion region $\mathcal D_{\epsilon}$ can be thereby determined by those edge-wise auxiliary variables. 
The following Theorem~\ref{app-thm:main_thm} further provide rate-distortion optimality analysis based on the result of Lemma~\ref{app-lem:exist_eps}.

\paragraph{Berger-Tung scheme.}
Before proving Theorem~\ref{app-thm:main_thm}, we briefly review the Berger-Tung achievability scheme~\citep{berger2003rate} through a simple example. Consider three correlated sources
$\mathbf{X}=(X_1,X_2,X_3)$ and two encoders observing the overlapping source pairs $(X_1,X_2)$ and $(X_2,X_3)$, respectively. Choose auxiliary variables according to
$P_{\mathbf{X},U_{1,2},U_{2,3}} = P_{\mathbf{X}} P_{U_{1,2}\mid X_1,X_2} P_{U_{2,3}\mid X_2,X_3}$, i.e.,
\begin{equation}
    U_{1,2} \to (X_1, X_2) \to X_3, \quad U_{2,3} \to (X_2, X_3) \to X_1, 
\end{equation}
and let $\hat{X}_{\ell} = g_{\ell}(U_{1,2},U_{2,3})$ satisfy the desired distortion constraint for each $\ell = 1,2,3$. 
Each encoder quantizes its observation into an auxiliary codeword and transmits its bin index. 
The decoder then jointly recovers both auxiliary codewords and reconstructs $\mathbf{X}$. 
The Berger-Tung inner bound guarantees that any rate pair satisfying
\begin{align}
    R_{1,2} & \geq I(X_1,X_2;U_{1,2} | U_{2,3}),\\
    R_{2,3} & \geq I(X_2,X_3;U_{2,3} | U_{1,2}),\\
    R_{1,2} + R_{2,3} & \geq I(\mathbf{X};U_{1,2},U_{2,3}),
\end{align}
is achievable. Thus, for fixed auxiliary variables, the minimum achievable sum-rate is
\begin{equation}
    R_{\mathrm{sum}} = I(\mathbf{X};U_{1,2},U_{2,3}).
\end{equation}
The same construction extends to an arbitrary collection of overlapping source subsets. In the proof below, the constructed pairwise and node-wise auxiliary variables play the roles of the Berger-Tung auxiliaries, yielding the desired achievable sum-rate upper bound.

\begin{definition}
    Define a non-empty distortion region under the Gaussian case as $\mathcal D^{\mathrm{G}}_{\epsilon} \triangleq \{ (D_1, \cdots, D_L): 0 < D_{\ell} \le \epsilon_{\ell}, \ell = 1, \cdots, L \}$ for some strictly positive $0 < \epsilon_j < \infty$. 
\end{definition}

\begin{theorem}[Gaussian optimality]\label{app-thm:main_thm}
Consider a $K$-order Gauss-Markov source $\mathbf{X} \sim \mathcal{N}(0, \Sigma)$ with $\Sigma \succ 0$. 
There exists a non-empty distortion region $\mathcal{D}^{\mathrm{G}}_{\epsilon}$, such that for all $\mathbf{D} \in \mathcal{D}^{\mathrm{G}}_{\epsilon}$, the following sets of random variables can be constructed:
\begin{itemize}[leftmargin=19.5pt]
    \vspace{-0.5em}
    \item[i)] (Gaussian auxiliary variables)
    It suffices to consider jointly Gaussian auxiliary random variables $\{ \tilde{X}_{j,k},  U_{j,k}\}_{(j,k)\in\mathcal{E}^{(K)}}$, $\{U_{\ell}\}_{\ell=1}^L$, and $\hat{\mathbf{X}}$.
    Moreover, the optimal $U_{j,k}$ and $U_{\ell}$ satisfy the Markov constraints:
    \begin{equation}\label{app-eq:KoKd_mc_opt}
    \begin{gathered}
        X_j \to ( X_{\mathcal{N}(j)\setminus\{k\}}, \, U_{j,k} ) \to X_k, \quad (j,k) \in \mathcal E^{(K)}, \\
        U_{j,k} \to (X_j, X_k) \to \mathbf X , \quad (j,k) \in \mathcal E^{(K)} \\
        U_{\ell} \to X_{\ell} \to U_{\ell, \mathcal{N}(\ell)} \to X_{-\ell}, \quad \ell = 1, \cdots, L.
    \end{gathered}
    \end{equation}
    \vspace{-1em}
    \item[ii)] (Linear interpolation)  
    Given $(X_j, X_k)$ and $U_{j,k}$, there exists some coefficients $\omega_{j,k} \in \mathbb{R}\setminus\{0,1\}$ such that auxiliary variable $\tilde{X}_{j,k}$ admit linear deterministic functions for $(X_j, X_k)$:
    \begin{equation}
        \tilde{X}_{j,k} = \omega_{j,k} X_j + (1 - \omega_{j,k}) X_k,
    \end{equation}
    and satisfies the Markov chain $U_{j,k} \to \tilde X_{j,k} \to (X_j, X_k) \to \mathbf X$. 
    \item[iii)] (Linear reconstruction)  
    The optimal $\hat{\mathbf{X}}$ adopt linear decoding mappings.
\end{itemize}
Moreover, with those constructed auxiliary variables, the sum-rate distortion of such system achieves the Shannon lower bound in \eqref{eq:slb} and hence the joint coding rate-distortion optimality. 
\end{theorem}
\begin{proof}
\underline{\textit{Existence of $\mathcal{D}^{\mathrm{G}}_{\epsilon}$}}: \\
First, as established in Lemma~\ref{app-lem:exist_eps}, for the edge-wise auxiliary random variables $\{U_{j,k}\}_{(j,k)\in \mathcal{E}^{(K)}}$ constructed according to \cref{app-eq:opt_mc_1,app-eq:opt_mc_2,app-eq:opt_mc_3}, the induced distortion for each source variable $X_{\ell}$, defined as $\epsilon_{\ell} \triangleq \mathbb{E}[\|X_{\ell} - \mathbb{E}[X_{\ell} | U_{\ell, \mathcal{N}(\ell)}]\|^2]$, is strictly positive and bounded, i.e. $0 < \epsilon_{\ell} < \infty$.
This establishes the existence of a nontrivial distortion region:
\begin{equation}
    \mathcal{D}^{\mathrm{G}}_{\epsilon} \triangleq \left\{ (D_1, \cdots, D_L):\; 0 < D_{\ell} \le \epsilon_{\ell}, \; \ell = 1, \cdots, L \right\}.
\end{equation}
Moreover, for each source variable $X_{\ell}$ in a $K$-order Gauss-Markov source, it has the $K$-banded conditional-dependence structure, i.e. $\mathcal N(\ell) \subset \{ \ell - K, \cdots, \ell - 1, \ell + 1, \cdots, \ell + K \}$. 
Therefore, by the Markov constraints in \eqref{app-eq:KoKd_mc_opt}, $\E[X_{\ell} | U_{\ell, \mathcal N(\ell)}]$ is a function of $(U_{\ell-K, \ell}, \cdots, U_{\ell - 1, \ell}, U_{\ell, \ell+1}, \cdots, U_{\ell, \ell + K})$. 

\underline{\textit{Gaussian auxiliary variables}}:

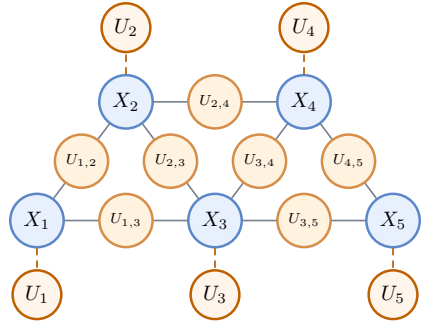
\begin{figure}[!t]
\vspace{-0.8em}
\centering
\resizebox{0.4\textwidth}{!}{
\begin{tikzpicture}[
    font=\small,
    >=Latex,
    line cap=round,
    line join=round,
    xnode/.style={
        circle,
        draw=xdraw,
        very thick,
        minimum size=0.9cm,
        fill=xfill,
        font=\normalsize,
        text=black
    },
    unode/.style={
        circle,
        draw=udraw,
        very thick,
        minimum size=0.78cm,
        fill=ufill,
        font=\scriptsize,
        text=black
    },
    vnode/.style={
        circle,
        draw=orange!75!black,
        very thick,
        minimum size=0.78cm,
        fill=orange!8,
        font=\normalsize,
        text=black
    },
    edge_style/.style={
        thick,
        draw=edgecol
    },
    dashed_arrow/.style={
        thick,
        draw=edgecol,
        dashed,
        ->
    },
    aux_edge/.style={
        thick,
        dashed,
        draw=orange!70!black
    }
]

\node[xnode] (x1) at (0,0) {$X_1$};
\node[xnode] (x3) at (3,0) {$X_3$};
\node[xnode] (x5) at (6,0) {$X_5$};

\node[xnode] (x2) at (1.5,2) {$X_2$};
\node[xnode] (x4) at (4.5,2) {$X_4$};

\draw[edge_style] (x1) -- (x2) node[midway, unode] (u12) {$U_{1,2}$};
\draw[edge_style] (x1) -- (x3) node[midway, unode] (u13) {$U_{1,3}$};
\draw[edge_style] (x2) -- (x3) node[midway, unode] (u23) {$U_{2,3}$};
\draw[edge_style] (x2) -- (x4) node[midway, unode] (u24) {$U_{2,4}$};
\draw[edge_style] (x3) -- (x4) node[midway, unode] (u34) {$U_{3,4}$};
\draw[edge_style] (x3) -- (x5) node[midway, unode] (u35) {$U_{3,5}$};
\draw[edge_style] (x4) -- (x5) node[midway, unode] (u45) {$U_{4,5}$};


\node[vnode] (ux1) at (0,-1.25) {$U_1$};
\node[vnode] (ux3) at (3,-1.25) {$U_3$};
\node[vnode] (ux5) at (6,-1.25) {$U_5$};

\node[vnode] (ux2) at (1.5,3.25) {$U_2$};
\node[vnode] (ux4) at (4.5,3.25) {$U_4$};

\draw[aux_edge] (x1) -- (ux1);
\draw[aux_edge] (x3) -- (ux3);
\draw[aux_edge] (x5) -- (ux5);

\draw[aux_edge] (x2) -- (ux2);
\draw[aux_edge] (x4) -- (ux4);

\end{tikzpicture}
}
\caption{Augmented graphical model with node-wise and edge-wise auxiliary variables.}
\vspace{-1em}
\label{app-fig:node_edge_aux_graph}
\end{figure}
Next, we show that arbitrary distortion tuple within the induced region, i.e. $\mathbf D \in \mathcal{D}^{\mathrm{G}}_{\epsilon}$, can be achieved by introducing a set of node-wise Gaussian auxiliary random variables $\{U_{\ell}\}^L_{\ell=1}$ satisfying the Markov chains:
\begin{equation}
    U_{\ell} \to X_{\ell} \to (X_{-\ell}, U_{\ell, \mathcal N(\ell)}),
\end{equation}
and then the JC optimality is also achieved. 
Suppose the edge-wise auxiliary random variables $\{U_{j,k}\}_{(j,k)\in\mathcal{E}^{(K)}}$ have been constructed according to \cref{app-eq:opt_mc_1,app-eq:opt_mc_2,app-eq:opt_mc_3}, which are all Gaussian variables. 
Then each source node $X_{\ell}$ can be decomposed as
\begin{equation}
    X_{\ell} = \mathbb{E}[X_{\ell} | U_{\ell, \mathcal{N}(\ell)}] + N_{\ell}, \quad \ell = 1, \cdots, L,
\end{equation}
where $N_{\ell} \sim \mathcal{N}(0, \sigma^2_{N_{\ell}})$ denotes Gaussian residual and is independent of $(U_{\ell, \mathcal{N}(\ell)}, X_{\mathcal{N}(\ell)})$. 
To further refine the reconstruction of $X_{\ell}$, we first introduce a Gaussian auxiliary random variable $\tilde U_{\ell}$ such that:
\begin{equation}
    U_{\ell} = X_{\ell} + W_{\ell}, \quad \text{with } W_{\ell} \sim \mathcal{N}(0, \sigma^2_{W_{\ell}}), \; W_{\ell} \indep X_{\ell}.
\end{equation}
Apparently, we have $U_{\ell} \to X_{\ell} \to (X_{-\ell}, U_{\ell, \mathcal N(\ell)})$. 
It can be verified that $X_{\ell}$ can be further decomposed as $X_{\ell} = \E[X_{\ell} | U_{\ell, \mathcal{N}(\ell)}, U_{\ell}] + Z_{\ell}$, where $Z_{\ell} \sim \mathcal N(0, \sigma^2_{Z_{\ell}})$ denote an independent Gaussian noise. 
With $\epsilon_{\ell} \triangleq \mathbb{E}[\|X_{\ell} - \mathbb{E}[X_{\ell} | U_{\ell, \mathcal{N}(\ell)}]\|^2]$ and introduced node-wise $U_{\ell}$, then we further have the corresponding distortion term as $D_{\ell} = (\epsilon_{\ell} + \frac{1}{\sigma^2_{W_{\ell}}})^{-1}$. 
Observe that by setting the variance $\sigma^2_{W_{\ell}}$ arbitrarily small, the induced distortion of $X_{\ell}$, i.e.,
\begin{equation}
    D_{\ell} = \mathbb{E}[\|X_{\ell} - \mathbb{E}[ X_{\ell} | U_{\ell, \mathcal{N}(\ell)}, U_{\ell}]\|^2 ],
\end{equation}
can be made arbitrarily close to zero. 
Therefore, by introducing such auxiliary variables $\{U_{\ell}\}_{\ell = 1}^L$ for each component, the achievable distortion can be continuously controlled over the range $(0, \epsilon_{\ell}]$. 
Consequently, any distortion tuple $\mathbf{D} \in \mathcal{D}^{\mathrm{G}}_{\epsilon}$ is achievable.
For example, Figure~\ref{app-fig:node_edge_aux_graph} illustrates the construction of $\{U_{j,k}\}_{(j,k)\in \mathcal{E}^{(2)}}$ and $\{U_{\ell}\}^5_{\ell = 1}$ for a second-order Gauss-Markov source $(X_1, \cdots, X_5)$. 

For the above designed coding system, by applying Lemma~1 in~\cite{chen2017generalized} together with the Berger-Tung inner/achievability bound, we obtain that, for all distortion tuples $\mathbf{D} \in \mathcal{D}^{\mathrm{G}}_{\epsilon}$,
\begin{equation}
    R^{\mathrm{ISC}}_{K,\mathrm{sum}}(\mathbf{D}) \le \frac{1}{2} \log \frac{|\Sigma|}{\prod_{\ell=1}^L D_{\ell}}.
\end{equation}
Conversely, by the Shannon lower bound, we have for any $\mathbf{D} \in \mathcal{D}^{\mathrm{G}}_{\epsilon} \subseteq \mathcal{D}_{\mathrm{JC}}$,
\begin{equation}
    R^{\mathrm{ISC}}_{K,\mathrm{sum}}(\mathbf{D}) \ge \frac{1}{2} \log \frac{|\Sigma|}{\prod_{\ell=1}^L D_{\ell}}.
\end{equation}
Combining the upper and lower bounds, for any $\mathbf{D} \in \mathcal{D}_{\epsilon}$, it follows that
\begin{equation}\label{app-eq:gauss_opt_rd_function}
    R^{\mathrm{ISC}}_{K,\mathrm{sum}}(\mathbf{D}) = \frac{1}{2} \log \frac{|\Sigma|}{\prod_{\ell=1}^L D_\ell}.
\end{equation}
This coincides with the joint coding rate-distortion function in \eqref{eq:slb}, and hence establishes joint coding optimality. This completes the proof.

\underline{\textit{Linear interpolation.}} \\
For jointly Gaussian $\mathbf{X}$ and $\{U_{j,k}\}_{(j,k)\in\mathcal{E}^{(K)}}$, the conditional expectation $\mathbb{E}[U_{j,k} | X_j, X_k]$ is linear on $(X_j, X_k)$. 
Hence, $U_{j,k}$ admits the orthogonal decomposition
\begin{align}
    U_{j,k} = \gamma_{j,k} X_j + \theta_{j,k} X_k + \tilde{N}_{j,k},
\end{align}
where $\tilde{N}_{j,k}$ is also Gaussian and independent of $(X_j, X_k)$. 
Since scaling of $U_{j,k}$ does not affect the rate-distortion tradeoff, we can normalize the coefficients by $\gamma_{j,k}+\theta_{j,k}$ to obtain
\begin{equation}
    U_{j,k} 
    = 
    \omega_{j,k} X_j + (1 - \omega_{j,k}) X_k + N_{j,k},
\end{equation}
where $\omega_{j,k} = \frac{\gamma_{j,k}}{\gamma_{j,k}+\theta_{j,k}}$ and $N_{j,k}$ is Gaussian and independent of $(X_j, X_k)$.
Furthermore, as shown in Lemma~\ref{app-lem:exist_eps}, the noise variance satisfies
\[
\sigma^2_{N_{j,k}} = -\frac{\omega_{j,k}(1-\omega_{j,k})}{J_{jk}}.
\]
In particular, $\omega_{j,k}\in(0,1)$ when $J_{jk}<0$, while $\omega_{j,k}>1$ when $J_{jk}>0$.

\underline{\textit{Linear reconstruction}}: \\
Given constructed edge-wise auxiliary variables $\{U_{j,k}\}_{(j,k) \in \mathcal E^{(K)}}$ and node-wise auxiliary variables $\{U_{\ell}\}^L_{\ell = 1}$, the final step is to reconstruct each source node $X_{\ell}$. 
Due to the Markov constraints in \eqref{app-eq:KoKd_mc_opt}, for each source variable $X_{\ell}$, we have the following reconstruction:
\begin{align*}
    \hat{X}_{\ell} &= \E[X_{\ell} | \{U_{j,k}\}_{(j,k) \in \mathcal E^{(K)}}, \{U_{\ell}\}^L_{\ell = 1}] = \E[X_{\ell} | U_{\ell, \mathcal N(\ell)}, U_{\ell}].
\end{align*}
Without loss of generality, we let $U_{\ell, \mathcal N(\ell)} = (U_{\ell-K, \ell}, \cdots, U_{\ell - 1, \ell}, U_{\ell, \ell+1}, \cdots, U_{\ell, \ell + K})$. 
Then for jointly Gaussian auxiliary variables, we have the optimal reconstruction $\hat{X}_{\ell}$ with linear decoding function: 
\begin{align*}
    \hat{X}_{\ell} &= {\mathbf{a}^{(\ell)}}^{\top} [U_{\ell-K,\ell}, \cdots, U_{\ell-1,\ell}, \cdots, U_{\ell,\ell+1}, \cdots, U_{\ell, \ell+K}, U_{\ell} ]^{\top} \\
    &= \sum^{K-1}_{k=0} a^{(\ell)}_{K-k} \cdot U_{\ell - 1 - k, \ell} + \sum^{2K-1}_{k=K} a^{(\ell)}_{k} \cdot U_{\ell, \ell - K + 1 + k} + a^{(\ell)}_{2K} \cdot U_{\ell} \\
    &= \sum^{K-1}_{k=0} a^{(\ell)}_{K-k} \cdot ( \omega_{\ell-1-k,\ell} X_{\ell-1-k} + \bar\omega_{\ell-1-k,\ell} X_{\ell} + N_{\ell-1-k,\ell} )  \\
    & \quad + \sum^{2K-1}_{k=K} a^{(\ell)}_{k} \cdot ( \omega_{\ell, \ell - K + 1 + k} X_{\ell} + \bar\omega_{\ell, \ell - K + 1 + k} X_{\ell - K + 1 + k} + N_{\ell, \ell - K + 1 + k} ) + a^{(\ell)}_{2K} \lambda_{\ell} \cdot X_{\ell} \\
    &= \sum^K_{k = -K} \beta_{K+1+k} X_{\ell + k} + Z_{\ell}, 
\end{align*}
where $Z_{\ell} = \sum^{K-1}_{k=0} a^{(\ell)}_{K-k} N_{\ell - 1- k,\ell} + \sum^{2K-1}_{k=K} a^{(\ell)}_k N_{\ell,\ell-K+1+k}$ is a Gaussian random variable independent with $\mathbf{X}$. This completes the proof. 
\end{proof}

\begin{theorem}[JC rate-distortion optimality]\label{app-thm:jc_opt}
Under the source model setting of Theorem~\ref{app-thm:main_thm}, there exists a non-empty distortion region $\mathcal D^{\mathrm{G}}_{\epsilon}$, such that for all $\mathbf{D} \in \mathcal{D}^{\mathrm{G}}_{\epsilon}$, $R^{\mathrm{ISC}}_{K, \mathrm{sum}}(\mathbf D)$ achieves the joint-coding rate-distortion function:
\begin{equation}
    R^{\mathrm{ISC}}_{K,\mathrm{sum}}(\mathbf{D}) = R^{\mathrm{JCG}}(\mathbf{D}), \quad \forall \mathbf D \in \mathcal D^{\mathrm{G}}_{\epsilon}.
\end{equation}
\end{theorem}
\begin{proof}
Theorem~\ref{app-thm:main_thm} already establishes the existence of two sets of edge-wise auxiliary random variables $\{U_{j,k}\}_{(j,k) \in \mathcal E^{(K)}}$ and node-wise auxiliary variables $\{U_j\}^L_{j=1}$, which satisfy the desired Markov constraints in \eqref{app-eq:KoKd_mc_opt} and achieve Shannon lower bound in the distortion region $\mathcal D_{\epsilon}$. 
In this proof, we show that these auxiliary variables are operationally compatible with the structure of the proposed ISC-$K$ system. 

\textbf{Step 1. Achievability scheme for ISC}: \\
In this part, we show those edge-wise and node-wise auxiliary variables can be incorporated into the proposed ISC-$K$ system without violating its coding structure:
\begin{enumerate}
    \item \textbf{Regrouped auxiliary variables.} As in the Definition~\ref{app-def:vf_k_ird}, there are only edge-wise auxiliary variable form $U_{j,k}$ involved in the ISC-$K$ system. 
    We first clarify how edge-wise auxiliary variables can be extended when node-wise auxiliary variables are present. 
    Basically, the node-wise variables do not require a separate class of messages.
    They can be incorporated into the pair-indexed ISC auxiliaries through a deterministic augmentation. 
    Given edge-wise and node-wise auxiliary variables as constructed in Theorem~\ref{app-thm:main_thm}, we consider a set of augmented auxiliary variables:
    \begin{equation}
        V_{j,k} \triangleq \begin{cases}
            U_{j,k}, & j < L-1, \; k < \min\{j+K,L\}, \\
            (U_{j,k},U_j), & j < L-1, \; k = \min\{j+K,L\}, \\
            (U_{L-1,L},U_{L-1},U_ L), & j=L-1, \; k=L.
        \end{cases}
    \end{equation}
    The idea is to simply augment the node-wise variable $U_j$ into the last edge-wise auxiliary variable associated with source node $X_j$. 
    The exception is the case when $j = L - 1$, where we have to append both $U_{L-1}$ and $U_L$ as there are no future frames after frame $X_L$. 
    Obviously, $\{V_{j,k}\}_{(j,k) \in \mathcal E^{(K)}}$ is a regrouping of components of $\{U_{j,k}\}_{(j,k)\in\mathcal E^{(K)}}$ and $\{U_{\ell}\}^L_{\ell = 1}$. 
    Therefore it represents the same collection of auxiliary variables. 
    \item \textbf{ISC Encoding.} As defined in Section~\ref{subsec:system_model}, ISC-$K$ encoding mapping $f_{j,k}$ can be decomposed as $f_{j,k} = \phi_{j,k} \circ h_{j,k}$, where $h_{j,k}$ denotes the deterministic interpolation function and $\phi_{j,k}$ denotes a sequential encoder. 
    For deterministic functions $h_{j,k}$, we adopt the following realizations:
    \begin{equation}
        h_{j,k}(X_j, X_k) = \tilde Y_{j,k} \triangleq \begin{cases}
            \tilde X_{j,k}, & j < L-1, \; k < \min\{j+K,L\}, \\
            (\tilde X_{j,k}, X_j), & j < L-1, \; k = \min\{j+K,L\}, \\
            (\tilde X_{j,k}, X_j, X_k), & j=L-1, \; k=L.
        \end{cases}
    \end{equation}
    Here, $\tilde X_{j,k}$ is attained by performing a linear interpolation between $(X_j, X_k)$ as stated in Theorem~\ref{app-thm:main_thm}. 
    Apparently we have the following Markov chains:
    \begin{equation}\label{app-eq:augaux_mc_for_interp}
        \left. 
        \begin{array}{r}
            U_{j,k} \to \tilde X_{j,k} \to (X_j, X_k) \to \mathbf X \\
            U_{\ell} \to X_{\ell} \to \mathbf X_{-\ell} 
        \end{array}
        \right\} \Rightarrow V_{j,k} \to \tilde Y_{j,k} \to (X_j, X_k) \to \mathbf X.
    \end{equation}
    Since the RHS Markov condition in \eqref{app-eq:augaux_mc_for_interp} is stronger than the encoder-side Markov condition of ISC-$K$ system in \eqref{eq:korder_mc_def}. 
    Therefore, the constructed auxiliary variables $V_{j,k}$ and $\tilde Y_{j,k}$ are a valid instantiation of encoder test channels permitted of ISC-$K$ system. 
    Then we perform the ISC-$K$ encoder on source pair $(X_j, X_k)$ to generate $V_{j,k}$. 
    \item \textbf{ISC Decoding.} As in \eqref{eq:korder_mc_def}, given received $V_{\le{\ell}}$ at $\ell$-stage ISC-$K$ decoder, we adopt the deterministic decoding function $\hat X_{\ell} = g_{\ell} ( V_{\le {\ell}} )$. 
\end{enumerate}
For augmented auxiliary variables $\{V_{j,k}\}_{(j,k) \in \mathcal E^{(K)}}$ and coding scheme constructed above, we have the following key observations:
\begin{enumerate}
    \item Since each $V_{j,k}$ contains all information of $U_{j,k}$, it also follow the Markov constraints analogous to $U_{j,k}$ stated in \cref{app-eq:opt_mc_2,app-eq:opt_mc_3}.
    \item Based on the definition of ISC-$K$ encoder, for each $(j,k)$-encoding stage and auxiliary variable $V_{j,k}$, the associate rate $\bar R_{j,k}$ follows from the conditional source coding theorem.
    \begin{align}
        \bar R_{j,k} = I(\tilde Y_{j,k}; V_{j,k} | V_{<{j,k}}).
    \end{align}
    Note that this has the same mutual information form as in Definition~\ref{app-def:vf_k_ird}, while ISC-$K$ coding is optimized over all admissible auxiliary variables. 
\end{enumerate}
Now for those constructed auxiliary variables, we can characterize the associate sum-rate and per-letter distortion. 
Specifically, for $j < L-1$ and $k < \min \{j+K, L\} $, the associate rate $\bar R_{j,k}$ follows by chain rule:
\begin{align}
    \bar R_{j,k} &= I(\tilde Y_{j,k}; V_{j,k} | V_{<{j,k}}) \\
    &= I(\tilde X_{j,k}; V_{j,k} | V_{<{j,k}}) \\
    &= I(X_j, X_k; V_{j,k} | V_{<{j,k}}).
\end{align}
For $j < L-1$ and $k = \min \{j+K, L\}$, the associate rate $\bar R_{j,k}$ follows by chain rule:
\begin{align}
    \bar R_{j,k} &= I(\tilde Y_{j,k}; V_{j,k} | V_{<{j,k}}) \\
    &= I(\tilde X_{j,k}, X_j; V_{j,k} | V_{<{j,k}}) \\
    &= I(X_j, X_k; V_{j,k} | V_{<{j,k}}).
\end{align}
For $j = L-1$ and $k = L$, the associate rate $R_{L-1,L}$:
\begin{align}
    \bar R_{L-1,L} &= I(\tilde Y_{L-1,L}; V_{L-1,L} | V_{<{L-1,L}}) \\
    &= I(\tilde X_{L-1,L}, X_{L-1}, X_L; V_{L-1,L} | V_{<{L-1,L}}) \\
    &= I(X_{L-1}, X_L; V_{L-1,L} | V_{<{L-1,L}}).
\end{align}
Now we can directly apply the same steps in \cref{app-eq:sum_rate_0,app-eq:sum_rate_1,app-eq:sum_rate_2} in the proof of Theorem~\ref{app-thm:min_sum_rate} to augmented auxiliary variables $\{V_{j,k}\}_{(j,k) \in \mathcal E^{(K)}}$, i.e.,
\begin{align}
\sum_{(j,k) \in \mathcal E^{(K)}} \bar R_{j,k} & = \sum_{(j,k) \in \mathcal E^{(K)}} I(X_j, X_k; V_{j,k} | V_{<{j,k}} ) \\
&= \sum_{(j,k) \in \mathcal E^{(K)}} I(\mathbf X; V_{j,k} | V_{<{j,k}} ) \\
&= I(\mathbf X; \{V_{j,k}\}_ {(j,k) \in \mathcal E^{(K)}} ) \\
&= I(\mathbf X; \{U_{j,k}\}_{(j,k) \in \mathcal E^{(K)}}, U_1, \cdots, U_L )
\end{align}
Now for decoding distortion of each source node $X_{\ell}$, since $\mathbf X$ and $\{V_{j,k}\}_{(j,k) \in \mathcal E^{(K)}}$ are jointly Gaussian, we can adopt the decoding introduced in the proof of Theorem~\ref{app-thm:main_thm}, i.e. $\E[X_{\ell} | U_{\ell, \mathcal N(\ell)}, U_{\ell}]$, which is valid for the ISC architecture based on the definition of $\ell$-th decoding function, i.e. $\hat X_{\ell} = g_{\ell}(V_{\le {\ell}})$. 
Therefore we have the induced distortion $\mathbf D = (D_1, \cdots, D_L) \in \mathcal D^{\mathrm{G}}_{\epsilon}$ and the corresponding rate-distortion function in \eqref{app-eq:gauss_opt_rd_function}. 
Notice that given the induced distortion $\mathbf D$, this is only one possible realization of ISC-$K$ coding system. Therefore we have the following upper bound:
\begin{equation}
    R^{\mathrm{ISC}}_{K,\mathrm{sum}}(\mathbf{D}) \le \bar R_{\mathrm{sum}}(\mathbf D) = \frac{1}{2} \log \frac{|\Sigma|}{\prod^L_{\ell = 1} D_{\ell}}.
\end{equation}

\textbf{Step 2. Converse part} \\
For the Gauss-Markov source $\mathbf X$ and arbitrary given distortion tuple $\mathbf D \in \mathcal D^{\mathrm{G}}_{\epsilon} \subseteq \mathcal D^{\mathrm{JC}}$, we can again use the Shannon lower bound:
\begin{equation}
    R^{\mathrm{ISC}}_{K,\mathrm{sum}}(\mathbf D) \ge R^{\mathrm{JCG}}(\mathbf D) = \frac{1}{2} \log \frac{|\Sigma|}{\prod^L_{\ell = 1} D_{\ell}}.
\end{equation}
Then apparently, for $K$-order Gauss-Markov source and arbitrary distortion $\mathbf D\in \mathcal D^{\mathrm{G}}_{\epsilon}$, we can adopt the ISC-$K$ coding scheme alongside those auxiliary random variables constructed in Theorem~\ref{app-thm:main_thm} and achieve the Shannon lower bound, thereby the joint-coding rate-distortion function. This completes the proof. 
\end{proof}

\subsection{Gaussian example}\label{app-sec:gauss_example}
In this section, we provide the explicit numerical calculation for the two-frame Gaussian source $(X_1, X_2) \sim \mathcal{N}(0, \Sigma)$. 
We assume the symmetric setting with unit variances $\sigma^2_1 = \sigma^2_2 = 1$ and correlation coefficient $\rho > 0$, such that:
\begin{equation}
    \Sigma = \begin{bmatrix}
        1 & \rho \\
        \rho & 1
    \end{bmatrix}.
\end{equation}

\subsubsection{Symmetric distortion case}
We first consider the symmetric distortion constraint:
\begin{equation}
    \mathbb{E}[\|X_j - \hat{X}_j\|^2] \le D, \quad j=1,2
\end{equation} 
\paragraph{Joint coding.} Recall the Gaussian joint coding rate-distortion function in \eqref{eq:jc_rd_gauss}. 
Let $\Sigma_e$ denote the reconstruction error covariance matrix. 
Under the symmetric distortion constraint, we parametrize:
\begin{equation}
    \Sigma_e = \begin{bmatrix}
        d & \theta \\
        \theta & d
    \end{bmatrix}.
\end{equation}
where $d \triangleq(\Sigma_e)_{11} = (\Sigma_e)_{22}$. 
Then the joint coding rate-distortion optimization problem becomes:
\begin{equation}
\begin{aligned}
    R^G_{\mathrm{JC}}(\mathbf{D}) = \min_{\Sigma_e}\; &\frac{1}{2} \log \frac{1-\rho^2}{|\Sigma_e|} \\
    \mathrm{s.t.}\; & 0 \prec \Sigma_e \preceq \Sigma, \\
    & (\Sigma_e)_{jj}\le D, \qquad j=1,2.
\end{aligned}
\end{equation}
Under the above parameterizations, this further reduces to
\begin{equation}
\begin{aligned}
    R^G_{\mathrm{JC}}(D) = \min_{\theta}\; & \frac{1}{2} \log \frac{1-\rho^2}{D^2-\theta^2} \\
    \mathrm{s.t.}\; & -D < \theta < D, \\
    & \rho + D - 1 \le \theta \le \rho + 1 - D.
\end{aligned}
\end{equation}
Solving the above optimization yields:
\begin{equation}
\label{app-eq:jc_gauss_example_rd}
    R^G_{\mathrm{JC}}(D) = \left\{\begin{array}{ll}
        \frac{1}{2} \log \frac{1-\rho^2}{D^2}, & 0 < D \le 1-\rho, \\[0.8ex]
        \frac{1}{2} \log \frac{1+\rho}{2D+\rho-1}, & 1-\rho < D \le 1, \\[0.8ex]
        0, & D \ge 1.
    \end{array}
    \right.
\end{equation}
The regime $0 < D \le 1-\rho$ corresponds to the high-resolution region where the Shannon lower bound is tight.

\paragraph{Sequential coding.} 
We next consider the strictly causal sequential coding setting. 
According to Corollary~1.2 in~\citet{ma2007delayed}, the sequential coding sum-rate function for the two-frame Gaussian source $(X_1,X_2)$ under symmetric distortion constraint is given by
\begin{equation}
    R_{\mathrm{seq,sum}}(D) = \left\{ \begin{array}{ll}
        \frac{1}{2} \log \frac{1}{D} + \frac{1}{2} \log \frac{ \rho^2 D + (1-\rho^2) }{ D }, & 0 < D \le 1, \\[1ex]
        0, & D \ge 1.
    \end{array}
    \right.
\end{equation}
The above expression can be interpreted as the sum of two terms corresponding to the sequential coding process:
\begin{equation}
    R_1(D) = \frac{1}{2}\log\frac{1}{D}, \quad R_2(D) = \frac{1}{2} \log \frac{ \rho^2 D + (1-\rho^2) }{ D }.
\end{equation}
The first term corresponds to encoding the first source $X_1$ without future information. 
The second term represents the additional rate required for encoding $X_2$ conditioned on the reconstruction of $X_1$. 
Compared with the joint-coding rate-distortion function in~\eqref{app-eq:jc_gauss_example_rd}, the sequential coding system incurs a nonzero performance gap due to the causal constraint, which prevents the encoder from jointly exploiting the correlation between $(X_1,X_2)$.

\paragraph{ISC.} We now consider the proposed ISC-$1$ framework. 
For the symmetric distortion setting, we fix the interpolation coefficient to $\omega_{1,2} = \frac{1}{2}$, and construct the interpolation-based auxiliary variable:
\begin{equation}
    \tilde{X}_{1,2} = \frac{1}{2} X_1 + \frac{1}{2} X_2, \quad  U_{1,2} = \tilde{X}_{1,2} + N_{1,2},
\end{equation}
where $N_{1,2}$ is Gaussian noise independent of $(X_1, X_2)$. 
The interpolated source has $\sigma^2_{\tilde{X}_{1,2}} = \frac{1 + \rho}{2}$. 
In the high resolution regime corresponding to the Markov chain $X_1 \to U_{1,2} \to X_2$, the noise variance is given by $\sigma^2_{N_{1,2}} = -\frac{1}{4J_{12}} = \frac{1 - \rho^2}{4\rho}$. 
This induces the distortion region:
\begin{equation}
    \mathcal{D}_{\epsilon} = \{ (D_1,D_2): D_1,D_2 \le 1-\rho \}
\end{equation}
where the JC optimality is achieved through the introduction of additional auxiliary random variables $U_1$ and $U_2$.

Next, consider the low-resolution regime $D \in (1 - \rho, 1]$. 
Under the symmetric distortion constraint, we have
\begin{equation}
\begin{aligned}
    D & = D_1 = \mathbb{E}[\|X_1 - \hat{X}_1 \|^2] \\
    &= \sigma^2_1 - \frac{\mathbb{E}[X_1 U_{1,2}]^2}{\sigma^2_{U_{1,2}}} \\
    &= 1 - \frac{ \frac{1}{4}(1 + \rho)^2 }{ \frac{1+\rho}{2} + \sigma^2_{N_{1,2}} }.
\end{aligned}
\end{equation}
Solving for the noise variance yields:
\begin{equation}
    \sigma^2_{N_{1,2}} = \frac{ (1 + \rho)(2D + \rho -1) }{4(1 - D)}.
\end{equation}
The corresponding rate is
\begin{equation}
\begin{aligned}
    R &= I(\tilde{X}_{1,2}; U_{1,2}) \\
    &= \frac{1}{2} \log\left( \frac{\sigma^2_{\tilde{X}_{1,2}}}{\sigma^2_{N_{1,2}}} + 1 \right) \\
    &= \frac{1}{2} \log \frac{1 + \rho}{2D + \rho - 1 }
\end{aligned}
\end{equation}
This coincides exactly with the joint-coding rate-distortion function in~\eqref{app-eq:jc_gauss_example_rd} for the low-resolution regime. 
Combined with the high-resolution result above, ISC-$1$ therefore achieves the Gaussian joint-coding rate-distortion performance for the two-frame Gaussian source.

\subsubsection{Asymmetric distortion}
We next consider the asymmetric distortion constraints:
\begin{equation}
    \mathbb{E}[\|X_1-\hat{X}_1\|^2] \le D_1, \quad
    \mathbb{E}[\|X_2-\hat{X}_2\|^2] \le D_2,
\end{equation}
where $D_1 \in (0, 1]$ is fixed. 
For convenience to compare within the high resolution regime, we further assume $D_1 \in (0, 1 - \rho^2]$.

\paragraph{Joint coding.} 
Under the asymmetric distortion setting, the reconstruction error covariance matrix takes the form
\begin{equation}
    \Sigma_e = \begin{bmatrix}
        D_1 & \theta \\
        \theta & d_2
    \end{bmatrix}
\end{equation}
where $d_2 \le D_2$. Then joint-coding rate-distortion optimization becomes:
\begin{equation}
\begin{aligned}
    R^G_{\mathrm{JC}}(\mathbf{D}) = \min_{\theta,d_2}\; &
    \frac12 \log \frac{ 1-\rho^2 }{ D_1d_2-\theta^2 } \\
    \mathrm{s.t.}\; & D_1d_2-\theta^2 > 0, \\
    & (1-D_1)(1-d_2) - (\rho-\theta)^2 \ge 0, \\
    & d_2 \le D_2.
\end{aligned}
\end{equation}
Solving the above optimization yields the following rate-distortion function:
\begin{equation}
\label{app-eq:jc_gauss_rd_asym}
    R^G_{\mathrm{JC}}(\mathbf{D}) =
    \left\{ \begin{array}{ll}
        \frac12 \log \frac{ 1-\rho^2 }{ D_1D_2 }, & 0 < D_2 \le
        \dfrac{1-\rho^2-D_1 }{ 1-D_1 }, \\[2ex]
        \frac12 \log
        \frac{ 1-\rho^2 }{ D_1D_2 - \left( \rho - \sqrt{ (1-D_1)(1-D_2) } \right)^2 }, &
        \dfrac{ 1-\rho^2-D_1 }{ 1-D_1 } < D_2 < 1-(1-D_1)\rho^2, \\[3ex]
        \frac12 \log \frac{1}{D_1}, & D_2 \ge 1-(1-D_1)\rho^2.
    \end{array}
    \right.
\end{equation}
The three regions correspond to different operating regimes of the Gaussian joint-coding solution. 
In particular, when $D_2$ is sufficiently small, the Shannon lower bound is tight and the optimal rate-distortion function reduces to the product form in the first case. 
As $D_2$ increases, the optimal reconstruction error covariance becomes correlated, leading to the middle regime. 
Finally, when $D_2$ is sufficiently large, only the distortion constraint on $X_1$ remains active, and the sum-rate reduces to the scalar Gaussian rate-distortion function of $X_1$ alone.

\paragraph{Sequential coding.} 
For the strictly causal sequential coding setting, the sum-rate function under asymmetric distortion constraints is given by:
\begin{equation}
    R_{\mathrm{seq,sum}}(\mathbf{D}) = \left\{ \begin{array}{ll}
        \frac12 \log \frac{1}{D_1} +
        \frac12 \log \frac{ \rho^2D_1+(1-\rho^2) }{ D_2 }, &
        D_2 < \rho^2D_1+(1-\rho^2), \\[2ex]
        \frac12 \log \frac{1}{D_1}, & \rho^2D_1+(1-\rho^2) \le D_2 \le 1, \\[2ex]
        0, & D_1 \ge 1,\; D_2 \ge 1.
    \end{array}
    \right.
\end{equation}
The above characterization still reflects the sequential nature of the coding process. When $D_2 \ge \rho^2D_1+(1-\rho^2)$, the distortion requirement on $X_2$ can already be satisfied using only the reconstruction of $X_1$, and therefore no additional rate is required for the second stage. 

\paragraph{ISC.} We now analyze the proposed ISC-$1$ under asymmetric distortion constraints.
Unlike the symmetric setting, the interpolation coefficient $\omega_{1,2}$ is optimized jointly with the distortion constraint. 
From the proof of Lemma~\ref{app-lem:exist_eps}, for $\rho > 0$, the optimal interpolation coefficient satisfies $\omega_{1,2} > 0$. 
We first characterize the high resolution regime where JC optimality is achieved through additional auxiliary random variables $U_1$ and $U_2$. 
Setting $\epsilon_1 \triangleq D_1$, the corresponding distortion threshold for the second source is 
\begin{equation}
    \epsilon_2 = \frac{1 - \rho^2 - \epsilon_1}{1 - \epsilon_1}.
\end{equation}
Therefore, for arbitrary $D_1 \in (0, 1 - \rho^2]$ and $D_2 \le \frac{1 - \rho^2 - D_1}{1 - D_1}$, ISC achieves JC rate-distortion function. 
Notably, the region coincides with the high-resolution regimes of the Gaussian joint-coding solution in~\eqref{app-eq:jc_gauss_rd_asym}. 
Outside this region, the sum-rate optimization reduces to:
\begin{equation}
\begin{aligned}
    R_{\mathrm{1,sum}}(\mathbf{D}) = \min_{\omega_{1,2}}\; &
    \frac12 \log \frac{ \sigma^2_{U_{1,2}} }{ \sigma^2_{N_{1,2}} } \\
    \mathrm{s.t.}\; &
    \mathbb{E}[ \|X_1-\hat{X}_1\|^2] = D_1, \\
    & \mathbb{E}[ \|X_2-\hat{X}_2\|^2 ] \le D_2.
\end{aligned}
\end{equation}
The variance term of auxiliary variables are:
\begin{align}
    & \sigma^2_{U_{1,2}} = \sigma^2_{\tilde{X}_{1,2}} + \sigma^2_{N_{1,2}} \\
    & \sigma^2_{\tilde{X}_{1,2}} = \omega^2_{1,2} + 2\omega_{1,2}(1 - \omega_{1,2})\rho + (1 - \omega_{1,2})^2.
\end{align}
We first consider the equality distortion constraint on the first source:
\begin{equation}
\begin{aligned}
    D_1 &= \mathbb{E}[\|X_1 - \hat{X}_1\|^2] \\
    &= 1 - \frac{ \left(\omega_{1,2} + (1 - \omega_{1,2})\rho \right)^2}{\sigma^2_{U_{1,2}}}.
\end{aligned}
\end{equation}
Equivalently, we have the noise variance:
\begin{equation}
    \sigma^2_{U_{1,2}} = \frac{ \left(\omega_{1,2} + (1 - \omega_{1,2})\rho \right)^2 }{ 1 - D_1}.
\end{equation}
For the second source, we obtain: 
\begin{equation}\label{app-eq:t_constaint}
\begin{aligned}
    D_2 & \ge \mathbb{E}[\|X_2-\hat{X}_2\|^2] \\
    &= 1 - \left(
    \frac{
    \omega_{1,2}\rho + (1-\omega_{1,2})
    }{
    \omega_{1,2} + (1-\omega_{1,2})\rho }
    \right)^2 (1-D_1).
\end{aligned}
\end{equation}
Define:
\begin{equation}
    t
    \triangleq
    \frac{
    \omega_{1,2}\rho
    +
    (1-\omega_{1,2})
    }{
    \omega_{1,2}
    +
    (1-\omega_{1,2})\rho
    }.
\end{equation}
Since $\omega_{1,2} \in (0, 1)$, we have $t > 0$. 
The distortion constraint in~\eqref{app-eq:t_constaint} becomes 
\begin{equation}
    t^2 \ge \frac{ 1-D_2 }{ 1-D_1 }.
\end{equation}
which is equivalent to $t \ge \sqrt{ \frac{1 - D_2}{ 1 - D_1} }$. 
Substituting this relation to the rate expression yields 
\begin{equation}
\begin{aligned}
    \frac{1}{2} \log \frac{ \sigma^2_{U_{1,2}} }{ \sigma^2_{N_{1,2}} } &= \frac{1}{2} \log \frac{1}{1 - \sigma^2_{\tilde{X}_{1,2}} / \sigma^2_{U_{1,2}} } \\
    &= \frac{1}{2} \log \frac{1}{ 1 - \frac{1 - D_1}{ \left(\omega_{1,2} + (1 - \omega_{1,2})\rho \right)^2 } \sigma^2_{\tilde{X}_{1,2}} } \\
    &= \frac{1}{2} \log \frac{1}{ 1 - \frac{1 - D_1 }{ 1 - \rho^2 } (1 - 2\rho t + t^2) }
\end{aligned}
\end{equation}
Therefore, the optimization reduces to minimizing $ 1 - 2\rho t + t^2$, whose unconstrained minimizer is $t^* = \rho$. 
Combining this with the distortion constraint in \eqref{app-eq:t_constaint} gives:
\begin{equation}
    t^* = \max \left\{\rho, \sqrt{\frac{1 - D_2}{1 - D_1} } \right\}.
\end{equation}
When $t^* = \rho$, the coding degenerates to $\omega_{1,2} = 1$, corresponding to coding only the first source. 
Otherwise, we have the optimal interpolation coefficient:
\begin{equation}
    \omega^*_{1,2} = \frac{1 - \rho \sqrt{ \frac{1 - D_2}{1 - D_1} }}{ 1 - \rho + (1 - \rho) \sqrt{ \frac{1 - D_2}{1 - D_1} } },
\end{equation}
for $D_2 < 1 - \rho^2(1 - D_1)$. 
Finally, the ISC sum-rate-distortion function becomes:
\begin{equation}
    R_{\mathrm{1,\mathrm{sum}}}(\mathbf{D}) =
    \left\{ \begin{array}{ll}
        \frac12 \log \frac{ 1-\rho^2 }{ D_1D_2 }, & 0 < D_2 \le
        \dfrac{1-\rho^2-D_1 }{ 1-D_1 }, \\[2ex]
        \frac12 \log
        \frac{ 1-\rho^2 }{ D_1D_2 - \left( \rho - \sqrt{ (1-D_1)(1-D_2) } \right)^2 }, &
        \dfrac{ 1-\rho^2-D_1 }{ 1-D_1 } < D_2 < 1-(1-D_1)\rho^2, \\[3ex]
        \frac12 \log \frac{1}{D_1}, & D_2 \ge 1-(1-D_1)\rho^2.
    \end{array}
    \right.
\end{equation}
This coincides exactly with the Gaussian joint-coding rate-distortion function in~\eqref{app-eq:jc_gauss_rd_asym}, thereby establishing the JC optimality of ISC-$1$ for the asymmetric distortion setting.

\begin{figure}[t]
\centering

\begin{tikzpicture}
\begin{axis}[
    hide axis,
    xmin=0, xmax=1,
    ymin=0, ymax=1,
    legend columns=3,
    legend to name=commonlegend,
    legend style={
        draw=none,
        font=\small,
        column sep=1.5em
    }
]
\addlegendimage{blue, solid, line width=1.5pt}
\addlegendentry{Joint}

\addlegendimage{green!70!black, dashed, line width=1.5pt}
\addlegendentry{Causal}

\addlegendimage{orange!90!white, dashdotted, line width=1.5pt}
\addlegendentry{ISC}

\end{axis}
\end{tikzpicture}

\pgfplotslegendfromname{commonlegend}


\begin{subfigure}[b]{0.32\textwidth}
\centering
\resizebox{\linewidth}{!}{
\begin{tikzpicture}
\begin{axis}[
    axis line style={line width=0.3pt},
    width=\linewidth,
    height=0.75\linewidth,
    xmin=0, xmax=1,
    ymin=0, ymax=3,
    xlabel={Distortion $D$},
    ylabel={Rate (Bits)},
    xlabel style={font=\tiny},
    ylabel style={font=\tiny},
    grid=both,
    major grid style={gray!30},
    minor grid style={gray!15},
    minor tick num=1,
    tick align=outside,
    tick pos=left,
    tick label style={
        font=\tiny,
        inner sep=-2pt
    },
    major tick length=2pt,
    minor tick length=2pt,
    legend style={
        at={(0.97,0.97)},
        anchor=north east,
        fill=white,
        draw=gray!30,
        font=\scriptsize
    },
    legend cell align={left},
    legend image post style={xscale=0.5},
    line width=1.5pt,
    mark size=2.2pt,
]

\pgfmathsetmacro{\rho}{0.7}
\pgfmathsetmacro{\dth}{1-\rho}
\pgfmathsetmacro{\Rth}{0.5*ln((1+\rho)/(1-\rho))/ln(2)}


\addplot[
    color=blue,
    mark=none,
    domain=0.01:1,
    samples=300,
]
{
    (x <= 1-\rho)
    *
    (0.5*ln((1-\rho^2)/(x^2))/ln(2))
    +
    (x > 1-\rho)
    *
    (0.5*ln((1+\rho)/(2*x+\rho-1))/ln(2))
};

\addplot[
    color=green!70!black,
    mark=none,
    dashed,
    domain=0.01:0.99,
    samples=250,
]
{0.5*ln((\rho^2*x + 1 - \rho^2)/(x^2))/ln(2)};

\addplot[
    color=orange!90!white,
    mark=none,
    dashdotted,
    domain=0.01:0.99,
    samples=300,
]
{
(x <= 1-\rho) * (0.5*ln((1-\rho^2)/(x^2))/ln(2))
+
(x > 1-\rho) *
(
0.5*ln(
(
(
0.5^2 + 2*0.5*(1-0.5)*\rho + (1-0.5)^2
)
/
(
((0.5 + (1-0.5)*\rho)^2)/(1-x)
-
(0.5^2 + 2*0.5*(1-0.5)*\rho + (1-0.5)^2)
)
+ 1
)
)/ln(2)
)
};

\addplot[
    only marks,
    mark=*,
    mark size=1.5pt,
    color=blue!70!black,
]
coordinates {(\dth,\Rth)};

\addplot[
    densely dashed,
    color=gray!60,
    line width=0.9pt
]
coordinates {(\dth,0) (\dth,4)};

\end{axis}
\end{tikzpicture}
}
\vspace{-1.0em}
\end{subfigure}
\hfill
\begin{subfigure}[b]{0.32\textwidth}
\centering
\resizebox{\linewidth}{!}{
\begin{tikzpicture}
\begin{axis}[
    axis line style={line width=0.3pt},
    width=\linewidth,
    height=0.77\linewidth,
    xmin=0, xmax=1,
    ymin=0, ymax=0.15,
    xlabel={Distortion $D$},
    ylabel={Rate Gap (Bits)},
    xlabel style={font=\tiny},
    ylabel style={font=\tiny},
    grid=both,
    major grid style={gray!30},
    minor grid style={gray!15},
    minor tick num=1,
    tick align=outside,
    tick pos=left,
    major tick length=2pt,
    minor tick length=2pt,
    tick label style={font=\tiny, inner sep=-2pt},
    ytick={0, 0.1},
    line width=1.1pt,
    mark size=2.4pt,
]

\pgfmathsetmacro{\rho}{0.5}

\addplot[
    color=purple,
    domain=0.01:0.99,
    samples=300,
]
{
    0.5*ln((\rho^2*x + 1 - \rho^2)/(x^2))/ln(2)
    -
    (
        (x <= 1-\rho)
        *
        (0.5*ln((1-\rho^2)/(x^2))/ln(2))
        +
        (x > 1-\rho)
        *
        (0.5*ln((1+\rho)/(2*x+\rho-1))/ln(2))
    )
};

\end{axis}
\end{tikzpicture}
}
\vspace{-1.0em}
\end{subfigure}
\hfill
\begin{subfigure}[b]{0.32\textwidth}
\centering
\resizebox{\linewidth}{!}{
\begin{tikzpicture}
\begin{axis}[
    axis line style={line width=0.3pt},
    width=\linewidth,
    height=0.75\linewidth,
    xmin=0, xmax=0.7,
    ymin=0.5, ymax=3.5,
    xlabel={Distortion $D_2$ (with $D_1 = d_1$)},
    ylabel={Rate (Bits)},
    xlabel style={font=\tiny},
    ylabel style={font=\tiny},
    grid=both,
    major grid style={gray!30},
    minor grid style={gray!15},
    minor tick num=1,
    tick align=outside,
    tick pos=left,
    tick label style={
        font=\tiny,
        inner sep=-2pt
    },
    major tick length=2pt,
    minor tick length=2pt,
    legend style={
        at={(0.97,0.97)},
        anchor=north east,
        fill=white,
        draw=gray!30,
        font=\scriptsize
    },
    legend cell align={left},
    legend image post style={xscale=0.5},
    line width=1.5pt,
    mark size=2.4pt,
]

\addplot[color=blue, opacity=0.35, forget plot] table {./imgs/dat/jc_D1_0.1.dat};
\addplot[color=green!70!black, dashed, opacity=0.35, forget plot] table {./imgs/dat/seq_D1_0.1.dat};
\addplot[color=orange!90!white, dashdotted, opacity=0.35, forget plot] table {./imgs/dat/lisc_D1_0.1.dat};


\addplot[color=blue, opacity=0.65, forget plot] table {./imgs/dat/jc_D1_0.2.dat};
\addplot[color=green!70!black, dashed, opacity=0.65, forget plot] table {./imgs/dat/seq_D1_0.2.dat};
\addplot[color=orange!90!white, dashdotted, opacity=0.65, forget plot] table {./imgs/dat/lisc_D1_0.2.dat};


\addplot[color=blue] table {./imgs/dat/jc_D1_0.4.dat};
\addplot[color=green!70!black, dashed] table {./imgs/dat/seq_D1_0.4.dat};
\addplot[color=orange!90!white, dashdotted] table {./imgs/dat/lisc_D1_0.4.dat};



\end{axis}
\end{tikzpicture}
}
\vspace{-1.0em}
\end{subfigure}

\vspace{-0.5em}
\caption{
Gaussian source $(X_1, X_2)$ with $\sigma^2_1 = \sigma^2_2 = 1$ and $\rho = 0.7$.
\textit{Left}: rate-distortion $R(D)$ comparisons for symmetric distortion constraint $D$. Interpolation coefficient $\omega_{1,2} = 0.5$ for ISC. 
\textit{Middle}: Rate gap between joint coding (ISC) and causal coding for symmetric distortion case. 
\textit{Right}: rate-distortion $R(D_2)$ comparisons for asymmetric distortion constraint with fixed $d_1 = 0.1, 0.2$, and $0.4$ upside down. 
}
\vspace{-1em}
\label{fig:jc_causal_lisc}
\end{figure}
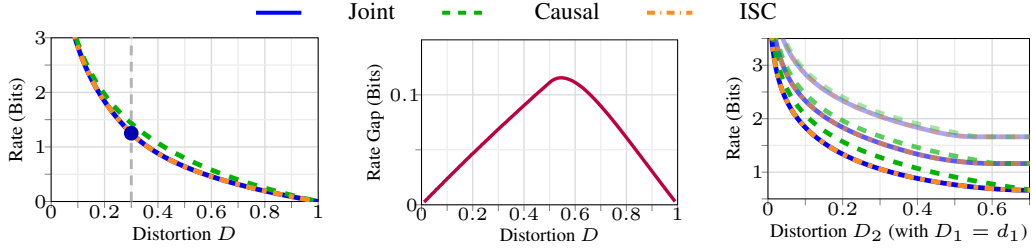

\subsubsection{Summary}\label{app-sec:gauss_example_summary}

We provide a simple two-frame Gaussian example to illustrate the behavior of the proposed ISC framework.
Consider jointly Gaussian sources $(X_1, X_2)$ with a quadratic distortion measure. 

\paragraph{Symmetric distortion.}
We first consider the symmetric setting $D_1 = D_2 = D$. 
Figure~\ref{fig:jc_causal_lisc} (left) shows the rate-distortion curves for joint coding, causal sequential coding, and ISC-$1$. 
In this case, the sum-rate is denoted by $R_{\mathrm{sum}}(D^2)$. 
As shown, ISC-$1$ achieves the same rate-distortion performance as joint coding in the high-resolution regime by introducing auxiliary variables through interpolation. 
Moreover, it strictly outperforms causal coding across the entire rate region.

\paragraph{Asymmetric distortion.}
We next consider the case where $D_1$ is fixed and vary $D_2$. 
Figure~\ref{fig:jc_causal_lisc} (right) shows the corresponding rate-distortion curves, denoted by $R_{D_1,\mathrm{sum}}(D_2)$. 
In this setting, the interpolation coefficients can be optimized according to the distortion constraint, and ISC achieves joint coding optimality across the entire feasible rate region.

\paragraph{Discussion.}
This example illustrates how interpolation-based auxiliary variables enable ISC to capture delayed dependencies efficiently. 
While the exact optimality result is specific to Gaussian sources, it provides a clear design principle for practical video coding: temporal dependencies can be exploited through local interpolation within a limited delay window without sacrificing coding efficiency.

\section{Experiment details}\label{app-sec:exp_details}
The proposed GFSC architecture is model-agnostic and can be integrated into different neural video compression (NVC) models.
In this section, we provide more implementation details about model-training, evaluation, and baseline setups. 
We also explicitly show how GFSC can be incorporated into DCVC~\citep{li2021deep} architecture, which serves as our primary baseline.
The same integration principle can be applied to other NVC architectures, such as VCT and DHVC, with only minor modifications.


\subsection{Training setup and overview}\label{app-subsec:train_details}

\textbf{Training and evaluation.}
All models are trained on Vimeo-90K~\citep{xue2019video} using 7-frame sequences ($\mathrm{GOP}=7$), with the remaining training hyperparameters following the corresponding original implementations. 
In our implementation, we focus on evaluating the effectiveness of the proposed GFSC mechanism rather than optimizing for peak performance. 
Therefore, for VCT and DHVC, we adopt a simplified training procedure corresponding to the final stage of VCT, where all components (encoder, decoder, and entropy model) are trained jointly from scratch. 
Specifically, we train the models for 50 epochs using their original rate-distortion objectives and optimization settings.
For DCVC, we adopt the full multi-stage training procedure of the original model as in~\cite{zhang2025opendcvcs} and introduce the delayed-frame modules only in the final training stage.
For evaluation, we report results on UVG~\citep{mercat2020uvg} and MCL-JCV~\citep{wang2016mcl}.
Following standard evaluation protocols, we encode the first 96 frames of each sequence with $\mathrm{GOP}=32$.
We evaluate rate-distortion performance using bitrate (bits-per-pixel, bpp) and PSNR computed in the RGB space. 
For conventional video codecs, reconstructed \verb|.YUV| videos are first converted to RGB before computing MSE and PSNR to ensure a consistent comparison.


\paragraph{Rate-distortion training.}
We follow the standard rate-distortion optimization used in DCVC~\cite{li2021deep}. 
First, the rate-distortion loss function is defined as
\begin{equation*}
    \mathcal{L} = R + \lambda D,
\end{equation*}
where $R$ denotes the bitrate and $D$ denotes the distortion. 
Both are averaged across the video sequence batch during training. 
The trade-off hyperparameter $\lambda$ is selected from  $\lambda \in 0.01 \times \{2, 1, 2^{-1}, 2^{-2}, 2^{-3}, 2^{-4}\}$, covering both low- and high-rate regimes.
For other settings, we use the Adam optimizer with a batch size of $8$, an initial learning rate of $1\times 10^{-5}$ and aux learning rate of $1 \times 10^{-3}$. 
Training is performed for $30$ epochs with a linear learning rate schedule: the learning rate is warmed up for the first $2$k steps and then linearly decayed to $1\times 10^{-6}$. 
All models are trained on a single NVIDIA RTX 5090 GPU using $2$~days.

\paragraph{Delayed coding baselines.} 
As discussed in Section~\ref{sec:exp}, we additionally compare ISC with practical delayed coding frameworks, including Random Access (RA) coding (i.e., bidirectional coding), under matched delay constraints ($K \le 3$). 
We first clarify the connection between sequential coding delay and RA coding configurations.

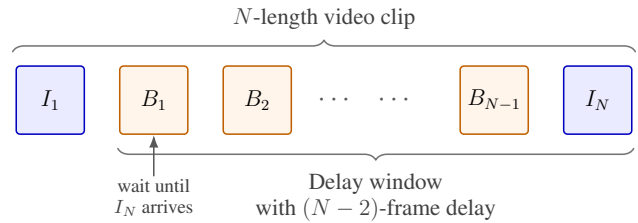
\begin{figure}[!ht]
\centering
\resizebox{0.6\linewidth}{!}{
\begin{tikzpicture}[
    font=\small,
    >=Latex,
    line cap=round,
    line join=round,
    iframe/.style={
        draw=blue!75!black,
        fill=blue!8,
        thick,
        rounded corners=2pt,
        minimum width=1.25cm,
        minimum height=1.25cm,
        font=\large
    },
    bframe/.style={
        draw=orange!75!black,
        fill=orange!8,
        thick,
        rounded corners=2pt,
        minimum width=1.25cm,
        minimum height=1.25cm,
        font=\large
    },
    brace/.style={
        decorate,
        decoration={brace, amplitude=6pt},
        thick,
        draw=black!55
    },
    mainarrow/.style={
        ->,
        thick,
        draw=black!65
    },
    lab/.style={
        inner sep=1pt,
        outer sep=0pt,
        text=black!75
    }
]

\node[iframe] (i1) at (0,0) {$I_1$};
\node[bframe] (b1) at (1.9,0) {$B_1$};
\node[bframe] (b2) at (3.8,0) {$B_2$};

\node[font=\Large, text=black!65] (dots1) at (5.25,0) {$\cdots$};
\node[font=\Large, text=black!65] (dots2) at (6.35,0) {$\cdots$};

\node[bframe] (bnm1) at (8.15,0) {$B_{N-1}$};
\node[iframe] (iN) at (10.05,0) {$I_N$};

\draw[brace] 
    ($(i1.north west)+(-0.05,0.25)$) --
    ($(iN.north east)+(0.05,0.25)$)
    node[midway, above=8pt, font=\large, text=black!75]
    {$N$-length video clip};

\draw[
    brace,
    decoration={brace, mirror, amplitude=6pt}
]
    ($(b1.south west)+(-0.02,-0.25)$) --
    ($(iN.south east)+(0.02,-0.25)$)
    node[midway, below=9pt, font=\large, align=center, text=black!75]
    {Delay window\\with $(N-2)$-frame delay};

\draw[mainarrow] ($(b1.south)+(0,-0.75)$) -- (b1.south);

\node[
    align=center,
    font=\normalsize,
    text=black!75
] at ($(b1.south)+(0,-1.15)$)
{wait until\\$I_N$ arrives};

\end{tikzpicture}
}
\caption{Illustration of the delay induced by B-frame coding over an $N$-length video clip.}
\label{app-fig:b_frame_delay_window}
\end{figure}

Consider an $L$-frame video sequence. 
In RA coding, the encoder first buffers a short frame group and performs bidirectional prediction within the group. 
For example, in an \verb|I-B-B-B-I| structure, intermediate B-frames depend on both past and future anchor frames. 
As a result, the encoder must wait for the arrival of the future I-frame before encoding the intermediate frames, leading to an effective practical delay determined by the distance between the first B-frame and the future anchor frame. 
The decoder then reconstructs frames in coding order and reorders them back to display order. 
Figure~\ref{app-fig:b_frame_delay_window} illustrates the corresponding delay configurations.

We consider two representative delayed coding baselines: DCVC-B~\cite{sheng2025bi} and HM-RA. 
For DCVC-B, the neural architecture allows flexible extension to delay settings. 
Therefore, both $K=1$ and $K=2$ are implemented by customizing the reference frame lists for each frame. 
For HM-RA, the hierarchical GOP structure is designed primarily for even GOP sizes. 
Accordingly, we modify the standard \verb|encoder_randomaccess_main.cfg| to implement $1$-delay and $3$-delay settings using \verb|I-B-I| and \verb|I-B-B-B-I| structures, respectively. 
The intermediate \verb|I-B-B-I| configuration is not directly supported under the standard HM-RA hierarchical coding structure.


\subsection{Network architecture}\label{app-subsec:net_arch}
\begin{figure}[!t]
\centering
\begingroup
\definecolor{archInk}{HTML}{354552}
\definecolor{archMuted}{HTML}{687986}
\definecolor{archMotion}{HTML}{648AA7}
\definecolor{archMotionFill}{HTML}{EDF4FA}
\definecolor{archContext}{HTML}{65958A}
\definecolor{archContextFill}{HTML}{EDF7F3}
\definecolor{archTMF}{HTML}{BE762A}
\definecolor{archTMFFill}{HTML}{FFF0DE}
\definecolor{archSTSA}{HTML}{8460A8}
\definecolor{archSTSAFill}{HTML}{F1EBF8}

\resizebox{.8\linewidth}{!}{
\begin{tikzpicture}[
    font=\normalsize,
    line cap=round,
    line join=round,
    flow/.style={
        -{Latex[length=2.0mm,width=1.45mm]},
        line width=0.85pt,
        rounded corners=1.5pt
    },
    refpath/.style={
        line width=0.8pt,
        rounded corners=1.5pt
    },
    refarrow/.style={
        refpath, -{Latex[length=2.0mm,width=1.45mm]}
    },
    archvar/.style={font=\large, inner sep=1.5pt},
    branch/.style={
        font=\small, 
    },
    operation/.style={
        rectangle, 
        minimum width=2.15cm, minimum height=0.68cm,
        line width=0.8pt, inner sep=3pt,
        align=center, font=\normalsize
    },
    motion/.style={
        draw=archMotion, 
        fill=archMotionFill
    },
    context/.style={
        draw=archContext, 
        fill=archContextFill
    },
    encoder/.style={
        trapezium,
        trapezium left angle=70,
        trapezium right angle=70,
        shape border rotate=180,
        minimum width=1.65cm, minimum height=0.68cm,
        inner sep=2pt, line width=0.8pt,
        font=\large
    },
    decoder/.style={
        trapezium,
        trapezium left angle=70,
        trapezium right angle=70,
        minimum width=1.65cm, minimum height=0.68cm,
        inner sep=2pt, line width=0.8pt,
        font=\large
    },
    quantizer/.style={
        rectangle, 
        draw=black, 
        fill=white, 
        line width=0.8pt,
        minimum size=0.54cm, inner sep=2pt,
        font=\normalsize
    },
    enhanced/.style={
        rectangle, 
        minimum height=0.90cm,
        inner sep=4pt, line width=1.1pt,
        font=\normalsize\bfseries, align=center
    },
    outputlabel/.style={
        font=\small, 
        inner sep=0pt
    }
]

\node[font=\large\bfseries] at (3.15,8.35) {(a) DCVC};
\node[font=\large\bfseries] at (13.55,8.35) {(b) DCVC + GFSC};
\draw[draw=black!75, line width=0.6pt] (0.0,7.98) -- (6.35,7.98);
\draw[draw=black!75, line width=0.6pt] (8.0,7.98) -- (19.10,7.98);

\node[branch] at (1.65,7.70) {Motion branch};
\node[branch] at (4.90,7.70) {Context branch};
\node[branch] at (10.80,7.70) {Motion branch};
\node[branch] at (16.10,7.70) {Context branch};

\node[archvar] (bref) at (1.05,7.16) {$\hat{X}_{t-1}$};
\node[archvar] (bxt) at (2.25,7.16) {$X_t$};
\node[operation,motion] (bme) at (1.65,6.0) {ME};
\node[encoder,motion] (bfm) at (1.65,5.0) {$f_m$};
\node[quantizer] (bqm) at (1.65,3.50) {$\mathrm Q$};
\node[decoder,motion] (bgm) at (1.65,2.0) {$g_m$};
\node[operation,motion] (bmc) at (1.65,1.0) {MC};
\node[archvar] (bcontext) at (1.65,-0.22) {$\bar{X}_t$};
\node[outputlabel] at (1.65,-0.62) {Context};

\draw[flow] (bref.south) -- ($(bme.north)+(-0.60,0)$);
\draw[flow] (bxt.south) -- ($(bme.north)+(0.60,0)$);
\draw[flow] (bme.south) -- (bfm.north);
\draw[flow] (bfm.south) -- (bqm.north);
\draw[flow] (bqm.south) -- (bgm.north);
\draw[flow] (bgm.south) -- (bmc.north);
\draw[flow] (bmc.south) -- (bcontext.north);

\coordinate (brefTap) at ($(bref.south)+(0,-0.08)$);
\draw[refpath] (brefTap) -| (0.02,1.0);
\draw[refarrow] (0.02,2.0) -- (bgm.west);
\draw[refarrow] (0.02,1.0) -- (bmc.west);
\fill[archMotion] (brefTap) circle (1pt);
\fill[archMotion] (0.02,2.0) circle (1pt);

\node[archvar] (bcinput) at (4.30,7.16) {$\bar{X}_t$};
\node[archvar] (bxinput) at (5.50,7.16) {$X_t$};
\node[encoder,context] (bfc) at (4.90,5.35) {$f_c$};
\node[quantizer] (bqc) at (4.90,3.50) {$\mathrm Q$};
\node[decoder,context] (bgc) at (4.90,1.7) {$g_c$};
\node[archvar] (bxhat) at (4.90,-0.22) {$\hat{X}_t$};
\node[outputlabel] at (4.90,-0.62) {Reconstruction};

\draw[flow] (bcinput.south) -- ($(bfc.north)+(-0.60,0)$);
\draw[flow] (bxinput.south) -- ($(bfc.north)+(0.60,0)$);
\draw[flow] (bfc.south) -- (bqc.north);
\draw[flow] (bqc.south) -- (bgc.north);
\draw[flow] (bgc.south) -- (bxhat.north);

\draw[draw=black!75, line width=0.6pt] (7.15,7.95) -- (7.15,5.65);
\draw[draw=black!75, line width=0.6pt] (7.15,3.98) -- (7.15,-0.58);
\node[align=center,font=\small\bfseries]
    at (7.15,5.14) {GFSC\\extension};
\draw[-{Latex[length=3mm,width=2.2mm]},line width=1.6pt]
    (6.60,4.55) -- (7.70,4.55);

\node[archvar] (gref0) at (8.75,7.16) {$\hat{X}_{t-1}$};
\node[archvar] (gxt0) at (9.95,7.16) {$X_t$};
\node[archvar] (gref1) at (11.65,7.16) {$\hat{X}_{t-1}$};
\node[archvar] (gxt1) at (12.85,7.16) {$X_{t+1}$};
\node[operation,motion] (gme0) at (9.35,6.15) {ME};
\node[operation,motion] (gme1) at (12.25,6.15) {ME};
\node[encoder,motion] (gfm0) at (9.35,5.05) {$f_m$};
\node[encoder,motion] (gfm1) at (12.25,5.05) {$f_m$};

\node[enhanced,draw=archTMF,fill=archTMFFill,minimum width=3.85cm]
    (tmf) at (10.80,3.75) {$\mathrm{TJM}_m$};
\node[quantizer] (gqm) at (10.80,2.65) {$\mathrm Q$};
\node[decoder,motion] (ggm) at (10.80,1.65) {$g_m$};
\node[operation,motion] (gmc) at (10.80,0.65) {MC};
\node[archvar] (gcontext) at (10.80,-0.22) {$\bar{X}_t$};
\node[outputlabel] at (10.80,-0.62) {Context};

\draw[flow] (gref0.south) -- ($(gme0.north)+(-0.60,0)$);
\draw[flow] (gxt0.south) -- ($(gme0.north)+(0.60,0)$);
\draw[flow] (gref1.south) -- ($(gme1.north)+(-0.60,0)$);
\draw[flow] (gxt1.south) -- ($(gme1.north)+(0.60,0)$);
\draw[flow] (gme0.south) -- (gfm0.north);
\draw[flow] (gme1.south) -- (gfm1.north);
\draw[flow] (gfm0.south) -- (gfm0.south |- tmf.north);
\draw[flow] (gfm1.south) -- (gfm1.south |- tmf.north);
\draw[flow] (tmf.south) -- (gqm.north);
\draw[flow] (gqm.south) -- (ggm.north);
\draw[flow] (ggm.south) -- (gmc.north);
\draw[flow] (gmc.south) -- (gcontext.north);

\coordinate (grefTap) at ($(gref0.south)+(0,-0.08)$);
\draw[refpath] (grefTap) -| (7.96,0.65);
\draw[refarrow] (7.96,1.65) -- (ggm.west);
\draw[refarrow] (7.96,0.65) -- (gmc.west);
\fill[archMotion] (grefTap) circle (1pt);
\fill[archMotion] (7.96,1.65) circle (1pt);

\node[archvar] (gcinput0) at (14.00,7.16) {$\bar{X}_t$};
\node[archvar] (gxinput0) at (15.20,7.16) {$X_t$};
\node[archvar] (gcinput1) at (17.00,7.16) {$\bar{X}_t$};
\node[archvar] (gxinput1) at (18.20,7.16) {$X_{t+1}$};
\node[encoder,context] (gfc0) at (14.60,5.55) {$f_c$};
\node[encoder,context] (gfc1) at (17.60,5.55) {$f_c$};
\node[enhanced,draw=archSTSA,fill=archSTSAFill,minimum width=4.30cm]
    (stsa) at (16.10,3.75) {$\mathrm{TJM}_c$};
\node[quantizer] (gqc) at (16.10,2.65) {$\mathrm Q$};
\node[decoder,context] (ggc) at (16.10,1.30) {$g_c$};
\node[archvar] (gxhat) at (16.10,-0.22) {$\hat{X}_t$};
\node[outputlabel] at (16.10,-0.62) {Reconstruction};

\draw[flow] (gcinput0.south) -- ($(gfc0.north)+(-0.60,0)$);
\draw[flow] (gxinput0.south) -- ($(gfc0.north)+(0.60,0)$);
\draw[flow] (gcinput1.south) -- ($(gfc1.north)+(-0.60,0)$);
\draw[flow] (gxinput1.south) -- ($(gfc1.north)+(0.60,0)$);
\draw[flow] (gfc0.south) -- ($(stsa.north)+(-1.50,0)$);
\draw[flow] (gfc1.south) -- ($(stsa.north)+(1.50,0)$);
\draw[flow] (stsa.south) -- (gqc.north);
\draw[flow] (gqc.south) -- (ggc.north);
\draw[flow] (ggc.south) -- (gxhat.north);

\node[font=\small,inner sep=0pt] at (9.55,-1.15) {
    ME: motion estimation \qquad
    MC: motion compensation \qquad
    $\mathrm Q$: quantization
};
\end{tikzpicture}%
}
\endgroup
\caption{Simplified DCVC and GFSC architectures. 
GFSC introduces temporal joint modeling (TJM) modules to both motion and context branch, highlighted in \emph{orange} and \emph{purple}, respectively. 
One delayed frame case is shown. 
Additional delayed frames provide further inputs to these modules. 
The mappings $f_m, f_c$ and $g_m,g_c$ denote motion/context encoders and decoders, respectively.}
\label{fig:dcvc_gfsc_architecture}
\end{figure}
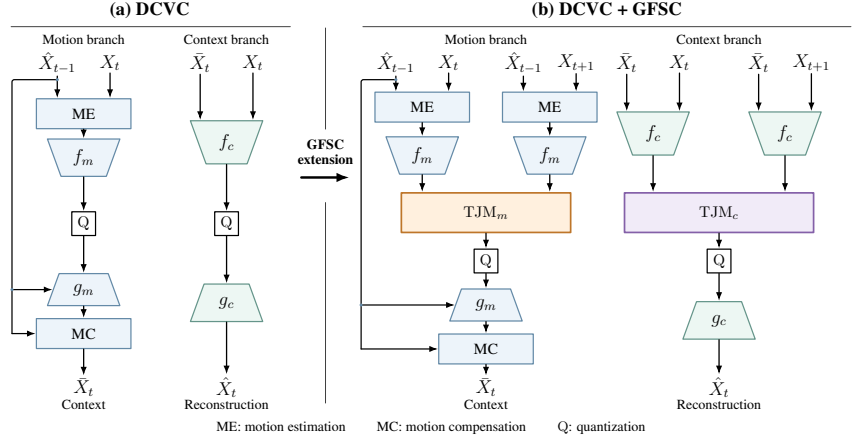

We describe how GFSC is integrated into the causal DCVC architecture.
The original DCVC codec contains separate I-frame and P-frame models. 
For P-frame coding, we distinguish two branches: a motion branch with analysis/synthesis transforms $(f_m, g_m)$ and a context branch with transforms $(f_c,g_c)$.
Each branch retains its original entropy-coding module based on the CompressAI implementation~\citep{begaint2020compressai}. 
Generally, GFSC introduces temporal joint modeling (TJM) modules between the analysis transforms and the subsequent coding modules, while leaving the remaining DCVC pipeline unchanged.
Figure~\ref{fig:dcvc_gfsc_architecture} compares the original DCVC and DCVC-GFSC architectures for the one-frame-delay case. 

\paragraph{One-frame delay ($K=1$).}
For every stage $\ell<L$, GFSC jointly models the current representation and the available one-frame lookahead. 
\begin{enumerate}
\item For an I-frame stage, let $f_I$ and $g_I$ denote the corresponding analysis and synthesis transforms. 
Given current and delayed frame pair $(\mathbf x_\ell,\mathbf x_{\ell+1})$, we first compute $\mathbf y_\ell = f_I(\mathbf x_\ell)$ and $\mathbf y_{\ell+1} = f_I(\mathbf x_{\ell+1})$. Then we have the following joint representation
\begin{equation*}
    \tilde{\mathbf y}_\ell = \mathrm{TJM}_{c} \left( \mathbf y_\ell,\mathbf y_{\ell+1} \right).
\end{equation*}
Here, $\mathrm{TJM}_{c}$ is implemented using the window-based spatial-temporal attention module described below. 
\item For P-frames, DCVC contains separate motion and context representations.
We therefore introduce two corresponding temporal modules, $\mathrm{TJM}_{m}$ and $\mathrm{TJM}_{c}$. 
For each branch $b\in\{m,c\}$,
\begin{equation*}
    \tilde{\mathbf y}^{b}_\ell = \mathrm{TJM}_{b} \left( \mathbf y^{b}_\ell, \mathbf y^{b}_{\ell+1} \right), \qquad \mathbf y^{b}_{\ell}=f_b(\cdot).
\end{equation*}
The context module $\mathrm{TJM}_{c}$ uses spatial-temporal self-attention to jointly learn delayed latent representations, whereas $\mathrm{TJM}_{m}$ uses a lightweight convolutional module to jointly model the motion features.
The resulting representations are then passed to the original entropy-coding and decoding pipeline.
All P-frame stages share the same network.
\end{enumerate}
Note that at the final stage $\ell = L$, no future frame is available.
GFSC therefore bypasses temporal fusion and directly forwards the current latent representation $\tilde{\mathbf y}^{b}_L = \mathbf y^{b}_L$ for $b\in\{m, c\}$.
Because the TJM outputs preserve the original latent tensor dimensions, the subsequent entropy models and synthesis transforms can be used without modification.

\paragraph{Extension to larger delays.}
The same construction extends directly to $K>1$. 
For the context branch, the spatial-temporal attention module jointly processes all available latent representations within the delay window:
\begin{equation*}
    \tilde{\mathbf y}^{c}_\ell = \mathrm{TJM}_{c}
    \left( \mathbf y^{c}_{\ell}, \cdots, \mathbf y^c_{\min\{\ell+K,L\}} \right).
\end{equation*}
The attention formulation naturally supports a variable number of available future frames near the sequence boundary. 
For the motion branch, $\mathrm{TJM}_{m}$ uses a fixed-size input corresponding to the selected delay budget.
For example, when $K=2$,
\begin{equation*}
    \tilde{\mathbf y}^{m}_\ell = \mathrm{TJM}_{m}
    \left(
        \mathbf y^{m}_\ell,
        \mathbf y^{m}_{\ell+1},
        \mathbf y^{m}_{\ell+2}
    \right).
\end{equation*}
If fewer than $K$ future representations are available, the missing inputs are replaced by zero tensors of the same shape. 
The same padding rule is used during both training and inference. 

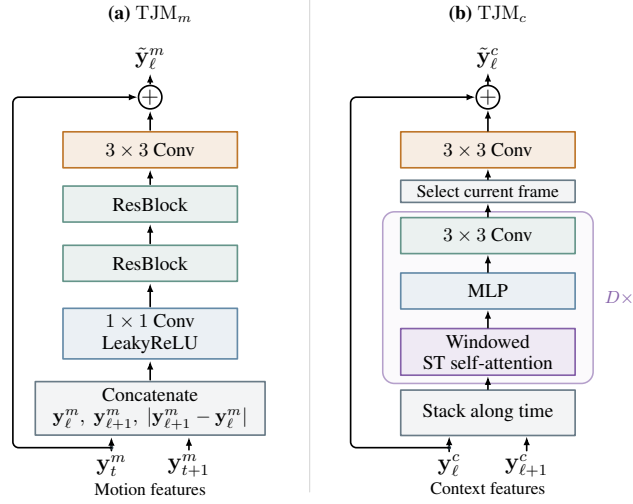
\begin{figure}[!t]
\centering
\begingroup
\definecolor{fusionInk}{HTML}{354552}
\definecolor{fusionMuted}{HTML}{697B87}
\definecolor{fusionBlue}{HTML}{648AA7}
\definecolor{fusionTeal}{HTML}{65958A}
\definecolor{fusionOrange}{HTML}{BE762A}
\definecolor{fusionPurple}{HTML}{8460A8}
\resizebox{.6\linewidth}{!}{
\begin{tikzpicture}[
    font=\normalsize, 
    line cap=round, 
    line join=round,
    flow/.style={-{Latex[length=1.6mm,width=1.2mm]},
        line width=0.85pt,
        rounded corners=3pt
    },
    motionflow/.style={flow,shorten <=0.5pt,shorten >=0.5pt},
    block/.style={
        rectangle,
        minimum width=3.15cm,minimum height=0.65cm,
        inner sep=3pt,line width=0.85pt,align=center},
    prep/.style={block,draw=fusionMuted,fill=fusionMuted!8},
    conv/.style={block,draw=fusionBlue,fill=fusionBlue!13},
    residual/.style={block,draw=fusionTeal,fill=fusionTeal!13},
    attention/.style={block,draw=fusionPurple,fill=fusionPurple!13},
    outputconv/.style={block,draw=fusionOrange,fill=fusionOrange!13},
    add/.style={
        circle,
        draw=black,
        fill=white,
        minimum size=0.40cm,inner sep=0pt,line width=0.85pt,font=\large},
    fusionvar/.style={font=\large,inner sep=1.5pt},
    note/.style={
        font=\small,
        align=center},
    heading/.style={font=\normalsize\bfseries,align=center}
]

\node[heading] at (0,8.15) {(a) $\mathrm{TJM}_m$};
\node[heading] at (6.10,8.15) {(b) $\mathrm{TJM}_c$};
\draw[black!15,line width=0.6pt] (2.88,-0.48) -- (2.88,8.48);

\node[fusionvar] (mt) at (-0.70,0) {$\mathbf y^m_t$};
\node[fusionvar] (md) at (0.70,0) {$\mathbf y^m_{t+1}$};
\node[note] at (0,-0.42) {Motion features};
\node[prep,minimum width=4.15cm,minimum height=0.82cm]
    (concat) at (0,1.03)
    {Concatenate\\$\mathbf y^m_{\ell},\; \mathbf y^m_{\ell+1},\;|\mathbf y^m_{\ell + 1}-\mathbf y^m_{\ell}|$};
\node[conv,minimum height=0.82cm] (reduce) at (0,2.38)
    {$1\times1$ Conv\\LeakyReLU};
\node[residual] (res1) at (0,3.65) {ResBlock};
\node[residual] (res2) at (0,4.72) {ResBlock};
\node[outputconv] (mvout) at (0,5.70) {$3\times3$ Conv};
\node[add] (mvadd) at (0,6.65) {$+$};
\node[fusionvar] (mvfused) at (0,7.35) {$\tilde{\mathbf y}^m_{\ell}$};

\draw[motionflow] (mt.north) -- (mt.north |- concat.south);
\draw[motionflow] (md.north) -- (md.north |- concat.south);
\draw[motionflow] (concat.north) -- (reduce.south);
\draw[motionflow] (reduce.north) -- (res1.south);
\draw[motionflow] (res1.north) -- (res2.south);
\draw[motionflow] (res2.north) -- (mvout.south);
\draw[motionflow] (mvout.north) -- (mvadd.south);
\draw[motionflow] (mvadd.north) -- (mvfused.south);

\coordinate (mtap) at (-0.70,0.32);
\fill[fusionInk] (mtap) circle (1pt);
\draw[flow] (mtap) -- (-2.48,0.32) -- (-2.48,6.65) -- (mvadd.west);

\draw[draw=fusionPurple!60,fill=fusionPurple!3,
      rounded corners=6pt,line width=0.8pt]
    (4.20,1.48) rectangle (8.00,4.58);
\node[note,text=fusionPurple,anchor=west] at (8.09,3.03) {$D\times$};

\node[fusionvar] (ft) at (5.40,0) {$\mathbf y^c_{\ell}$};
\node[fusionvar] (fd) at (6.80,0) {$\mathbf y^c_{\ell + 1}$};
\node[note] at (6.10,-0.42) {Context features};
\node[prep,minimum height=0.82cm] (stack) at (6.10,0.95)
    {Stack along time};
\node[attention,minimum height=0.82cm] (attn) at (6.10,2.05)
    {Windowed\\ST self-attention};
\node[conv] (mlp) at (6.10,3.15) {MLP};
\node[residual] (mix) at (6.10,4.15) {$3\times3$ Conv};
\node[prep,minimum height=0.40cm,inner sep=2pt,font=\small]
    (select) at (6.10,4.95) {Select current frame};
\node[outputconv] (ctxout) at (6.10,5.70) {$3\times3$ Conv};
\node[add] (ctxadd) at (6.10,6.65) {$+$};
\node[fusionvar] (ctxfused) at (6.10,7.35) {$\tilde{\mathbf y}^c_{\ell}$};

\draw[flow] (ft.north) -- (ft.north |- stack.south);
\draw[flow] (fd.north) -- (fd.north |- stack.south);
\draw[flow] (stack.north) -- (attn.south);
\draw[flow] (attn.north) -- (mlp.south);
\draw[flow] (mlp.north) -- (mix.south);
\draw[flow] (mix.north) -- (select.south);
\draw[flow] (select.north) -- (ctxout.south);
\draw[flow] (ctxout.north) -- (ctxadd.south);
\draw[flow] (ctxadd.north) -- (ctxfused.south);

\coordinate (ftap) at (5.40,0.32);
\fill[fusionInk] (ftap) circle (1pt);
\draw[flow] (ftap) -- (3.62,0.32) -- (3.62,6.65) -- (ctxadd.west);

\end{tikzpicture}%
}
\endgroup
\caption{Temporal joint modeling modules for motion and context branches in DCVC architecture. 
\textit{Left}: $\mathrm{TJM}_m$ concatenates motion features and their absolute difference as input. 
\textit{Right}: $\mathrm{TJM}_c$ jointly updates the two context features using a spatial-temporal self attention module.}
\label{fig:gfsc_fusion_modules}
\end{figure}

\textbf{Coding and quantization.}
After joint modeling within the $K$-delay window, the learned representation $\mathbf{z}_{\ell}$ is quantized and entropy-coded. 
Following the one-shot formulation in Section~\ref{subsec:system_model}, we adopt universal quantization with shared randomness~\cite{gray2002dithered, ziv2003universal, theis2021advantages} to implement one-shot channel simulation. 
Specifically, the encoder computes
\begin{equation*}
    \mathbf{z}_{\ell} = \Delta \left\lfloor \frac{ \tilde{\mathbf{y}}_{\ell} + \mathbf{w}}{\Delta} \right\rceil,
\end{equation*}
where $\mathbf{w} \sim \mathcal{U}(-\Delta/2, \Delta/2)$ denotes shared randomness. 
At the decoder side, the quantized latent is reconstructed as
\begin{equation}
    \mathbf{u}_{\ell} = \mathbf{z}_{\ell} - \mathbf{w},
\end{equation}
since the randomness $\mathbf{w}$ is shared between the encoder and decoder. 
Let $g_{\ell}$ denote the decoder network. 
The reconstructed frame is then obtained by:
\begin{equation}
    \hat{\mathbf{x}}_{\ell} = g_{\ell}(\mathbf{u}^{\ell}).
\end{equation}


\subsection{Additional experimental analysis}\label{app-subsec:add}
We further evaluate GFSC-enhanced DCVC on the 4K LIVE-HDR dataset~\citep{shang2022subjective}. Compared with causal DCVC, GFSC achieves BD-rate reductions of $3.78\%$, $4.50\%$, and $5.11\%$ for delays $K=1,2,3$, respectively. 
These results show that exploiting information from delayed frames also improves rate-distortion performance for 4K video.

\begin{figure}[!t]
\centering
\begingroup
\definecolor{uvgCausal}{HTML}{666E77}
\definecolor{uvgDelayOne}{HTML}{0072B2}
\definecolor{uvgDelayTwo}{HTML}{D58500}
\definecolor{uvgDelayThree}{HTML}{00876D}

\tikzset{
    uvgBase/.style={
        line width=1.05pt,
        mark size=1.75pt,
        mark options={solid,fill=white,line width=0.65pt},
    },
    uvgCausalLine/.style={
        color=uvgCausal,solid,mark=diamond*,
    },
    uvgOneLine/.style={
        color=uvgDelayOne,densely dashed,mark=o,
    },
    uvgTwoLine/.style={
        color=uvgDelayTwo,
        dash pattern=on 4pt off 1.5pt on 0.7pt off 1.5pt,
        mark=square*,
    },
    uvgThreeLine/.style={
        color=uvgDelayThree,solid,mark=triangle*,mark size=2.05pt,
    },
}

\newcommand{\uvgKey}[1]{
    \tikz[baseline=-0.55ex]{
        \draw[uvgBase,#1,mark=none] (0,0) -- (0.36,0);
        \draw[uvgBase,#1] plot coordinates {(0.18,0)};
    }\hspace{3pt}
}

\begin{minipage}[c]{0.53\linewidth}
\centering
\begin{tikzpicture}
\begin{axis}[
    width=\linewidth,
    height=0.75\linewidth,
    xmin=0.035,xmax=0.09,
    ymin=38.5,ymax=41.3,
    xtick={0.03,0.05,0.07,0.09,0.11},
    ytick={38,39,40,41,42},
    scaled x ticks=false,
    scaled y ticks=false,
    xticklabel style={
        /pgf/number format/fixed,
        /pgf/number format/fixed zerofill,
        /pgf/number format/precision=2,
    },
    yticklabel style={
        /pgf/number format/fixed,
        /pgf/number format/precision=0,
    },
    xlabel={Rate (bpp)},
    ylabel={PSNR (dB)},
    xlabel style={font=\footnotesize,yshift=1pt},
    ylabel style={font=\footnotesize,yshift=-1pt},
    tick label style={font=\scriptsize,inner sep=1pt},
    axis line style={draw=black!60,line width=0.45pt},
    tick style={draw=black!55,line width=0.4pt},
    tick align=outside,
    tick pos=left,
    major tick length=2pt,
    grid=major,
    major grid style={draw=black!12,line width=0.3pt},
    axis background/.style={fill=white},
    every axis plot/.append style={uvgBase},
    clip=true,
]
\addplot[uvgCausalLine] coordinates {
    (0.0299,37.903)
    (0.0408,39.140)
    (0.0531,39.972)
    (0.0793,40.854)
};
\addplot[uvgOneLine] coordinates {
    (0.0377,38.912)
    (0.0559,40.212)
    (0.0700,40.712)
    (0.0903,41.253)
};
\addplot[uvgTwoLine] coordinates {
    (0.0447,39.492)
    (0.0593,40.382)
    (0.0769,40.939)
    (0.0999,41.529)
};
\addplot[uvgThreeLine] coordinates {
    (0.0497,39.802)
    (0.0603,40.432)
    (0.0819,41.099)
    (0.1049,41.659)
};
\end{axis}
\end{tikzpicture}
\end{minipage}
\begin{minipage}[c]{0.45\linewidth}
\centering
{\small BD-rate comparison.\par}
\vspace{5pt}
{\footnotesize
\setlength{\tabcolsep}{3pt}
\renewcommand{\arraystretch}{1.32}
\begin{tabular}{|l|c|}
\hline
\raisebox{0.45\baselineskip}{ Model } & \shortstack{ BD-rate (\%)  } \\
\hline
DCVC      & $\phantom{-}0.00$ \\
\hline
DCVC-GFSC-1      & $-3.78$ \\
DCVC-GFSC-2      & $-4.50$ \\
DCVC-GFSC-3      & $-5.11$ \\
\hline
\end{tabular}\par
}
\vspace{4pt}
\end{minipage}

\endgroup
\caption{
\textit{Left}: Rate-PSNR curves on LIVE-HDR 4K.
\textit{Right}: BD-rate relative to causal DCVC. 
Negative values indicate rate savings.
}
\label{fig:4k_rd_bdrate}
\end{figure}
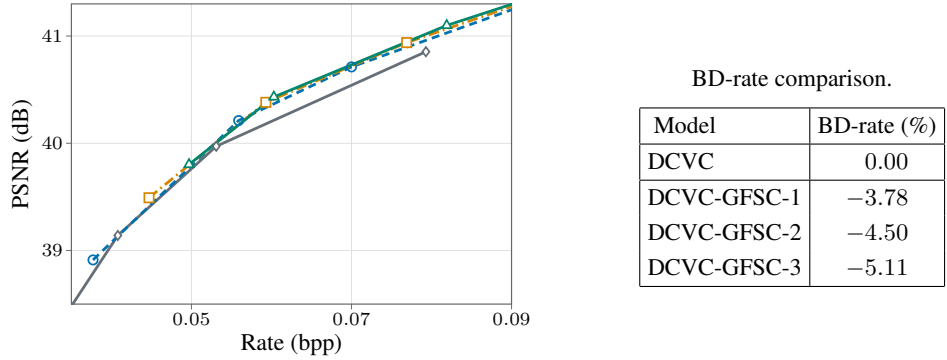





\section{Limitations}
\label{app-sec:limitations}

Despite the theoretical and empirical results established in this work, several questions remain open.
First, our joint-coding optimality result is characterized over a nontrivial distortion region.
Extending this characterization beyond the current region, and determining whether alternative auxiliary-variable constructions can enlarge the optimality regime, are important directions for future work.
This limitation concerns only the extent of the characterized region and does not affect the established optimality guarantee within it.

Second, while the general Markov analysis establishes the existence and graphical structure of JC-optimal clique-wise auxiliary variables, it does not yield a simple closed-form parameterization for arbitrary source distributions.
For Gauss-Markov sources, the additional Gaussian structure makes this construction explicit and reduces the clique-wise interactions to pairwise linear interpolation.
Developing similarly tractable constructions for broader non-Gaussian source families is an interesting theoretical direction.

Finally, the computational cost of GFSC grows approximately with the delay budget because additional future-frame interactions must be modeled.
This overhead is moderate in the limited-delay regime studied in this work, but scaling GFSC to substantially longer delays while controlling computation and memory remains an important engineering direction.

\end{document}